\documentclass{iosart2c}     

\usepackage[T1]{fontenc}
\usepackage{times}%
\usepackage{multirow}
\usepackage{subcaption}
\usepackage{graphicx}
\usepackage{epstopdf}
\usepackage{amssymb}
\usepackage{url}
\usepackage{tabularx}
\usepackage{rotating}
\usepackage{color}
\usepackage{enumitem}

\pubyear{0000}
\volume{0}
\firstpage{1}
\lastpage{1}

\begin{document}

\begin{frontmatter}    
\title{Bootstrapping and Multiple Imputation Ensemble Approaches for Classification Problems}
\runningtitle{Bootstrapping and Multiple Imputation Ensemble Approaches for Missing Data}

\author[A]{\fnms{Shehroz S.} \snm{Khan}},
\author[B]{\fnms{Amir} \snm{Ahmad}\thanks{Corresponding author. E-mail: amirahmad@uaeu.ac.ae}}
and
\author[C]{\fnms{Alex} \snm{Mihailidis}}
\runningauthor{Khan et al.}
\address[A]{Toronto Rehabilitation Institute, 550 University Avenue, Toronto, Canada, ON, M5G 2A2\\
E-mail: shehroz.khan@uhn.ca}
\address[B]{College of Information Technology, United Arab Emirates University, Al Ain, UAE\\
E-mail: amirahmad@uaeu.ac.ae}
\address[C]{Department of Occupational Sciences and Occupational Therapy, University of Toronto, Canada\\
E-mail: alex.mihailidis@utoronto.ca}

\begin{abstract}
Presence of missing values in a dataset can adversely affect the performance of a classifier. Single and Multiple Imputation are normally performed to fill in the missing values. In this paper, we present several variants of combining single and multiple imputation with bootstrapping to create ensembles that can model uncertainty and diversity in the data, and that are robust to high missingness in the data. We present three ensemble strategies: bootstrapping on incomplete data followed by (i) single imputation and (ii) multiple imputation, and (iii) multiple imputation ensemble without bootstrapping. We perform an extensive evaluation of the performance of the these ensemble strategies on eight datasets by varying the missingness ratio. Our results show that bootstrapping followed by multiple imputation using expectation maximization is the most robust method even at high missingness ratio (up to $30\%$). For small missingness ratio (up to $10\%$) most of the ensemble methods perform equivalently but better than single imputation. Kappa-error plots suggest that accurate classifiers with reasonable diversity is the reason for this behaviour. A consistent observation in all the datasets suggests that for small missingness (up to $10\%$), bootstrapping on incomplete data without any imputation produces equivalent results to other ensemble methods. 
\end{abstract}


\end{frontmatter}

\section{Introduction}

Predictive models assume that the data they use are complete, i.e., there are no missing values present in it. However, missingness in data is common and difficult to deal with \cite{HuangData2016, Donders2006Review,conroy2016dynamic}. Data with missing attribute values is called incomplete data. Many predictive algorithms cannot handle incomplete data, including support vector machines, neural networks, and logistic regression. Some classification algorithms can handle missingness in the data, such as decision trees (C4.5 \cite{quinlan2014c4}) and their variants. However, presence of a large amount of missingness in the data can deteriorate the performance of those classifiers.

There are several strategies to deal with incomplete data. A naive method is to remove any data object (or observation) with missing values. This strategy reduces the training data size; if the missingness ratio is high then generalizable models are difficult to learn. A better strategy is to replace a missing attribute value with some value - this is called imputation. Imputation can be single or multiple. In single imputation, a missing value is replace by one value, whereas in multiple imputation (MI), several values are imputed. MI performs better than single imputation in terms of modelling the uncertainty and variation due to the missing value \cite{rezvan2015rise}. Some common methods for imputation are fixed-value imputation, random imputation, nearest neighbour imputation, mean imputation \cite{gelman2006data,schmitt2015comparison} and expectation maximization imputation \cite{Lin2010} (see Section \ref{sec:imputation} for more details). It is to be noted that the present paper does not cover the situation in which the decision boundary is determined only by missing values.

MI methods generate multiple values corresponding to a missing value. To use multiple imputed data for training a classification algorithm, one option is to average multiple imputed values and replace the missing attribute values with a single value. The other option is to train different classifiers on different copies of imputed complete data and create an ensemble \cite{khan2012bayesian}. It has been shown that combining bootstrapping with MI can result in accurate classifiers \cite{wu2013new}. The reason is that MI accounts for the uncertainty due to the missing data, whereas bootstrapping accounts for the uncertainty due to sampling fluctuations \cite{wu2013new}. Combining both the ideas result in more diverse classifiers that aides the ensemble to perform better than a base classifier. In this paper, we discuss several ideas for creating ensembles to handle missing data. The ensemble techniques we test are: (i) bootstrapping with single imputation and average of MI, (ii) MI on bootstrap samples of incomplete data, and (iii) an ensemble of multiple imputed data. We use three popular data imputation techniques to validate the ensemble methods and discuss their relative performance. We systematically increase the amount of missingness in datasets (from $5\%$ to $30\%$) and evaluate the performance of each of these methods. Kappa-error plots \cite{Kappa} have been used to explain the performance of various ensemble methods at different level of missingness. We also show a comparison of kappa-error graphs to explain the diversity and accuracy of the different ensemble methods at different level of missingness. The results on eight UCI datasets \cite{Lichman:2013} show that the performance of MI after bootstrapping with expectation maximization imputation technique remains very robust despite increasing the missingness to a large value (up to $30\%$); however, it can be computationally extensive. Kappa error graphs show that bootstrapping with expectation maximization imputation technique creates accurate and diverse classifiers. MI after bootstrapping with mean imputation emerged as a robust and faster alternative when the missingness is low (up to $10\%$). We obtained a consistent observation on all the datasets that for low missingness (up to $10\%$), bagging ensembles on incomplete data performs equivalent to other imputation methods.

The major contributions of the paper are following;
\begin{enumerate}
\item  In this paper, a comprehensive comparison of different imputation methods and their ensembles, created by bagging and multiple imputations, is presented by varying the missingness ratio.
\item Kappa-error plots are used to explain the performance of the various ensemble methods.
\end{enumerate}

The rest of the paper is organized as follows. In Section \ref{sec:litrev}, we present the literature survey on data imputation using ensemble learning techniques. In Section \ref{sec:methods}, we present the different imputation methods used in the paper. Section \ref{sec:bmi} discusses the different bootstrapping and multiple imputation ensemble methods for handling incomplete data. Section \ref{sec:results} describes the experimental set up, datasets and results. We conclude the paper in Section \ref{sec:conclusion}.

\section{Literature Review}
\label{sec:litrev}
MI for missing data has been studied extensively in the literature (e.g., \cite{horton2001multiple, Donders2006Review, Harel2007Multiple}). In this literature review, we survey research papers that use ensemble learning with either multiple or single imputation to deal with missing data.

 Feelders \cite{feelders1999handling} compares surrogate splits in a decision tree with single and MI based on expectation maximization (EM) method. In the MI case, they compute the average over different imputations. Both the imputation methods perform better than surrogate split. They comment that averaging over MI and replacing with one value reduces the variance, in the same way as bagging, which improves the performance. 
Twala and Cartwright \cite{twala2010ensemble} propose an ensemble approach by creating sub-samples of incomplete data using bootstrap sampling. Each incomplete sample is fed to a decision tree classifier. The resulting ensemble is optimized in size by only choosing de-correlated decision trees and their output is combined to take a decision. In this method, direct imputation does not happen; however, they later incorporated additional MI techniques. The paper does not clearly state that at what stage MI was used in the ensemble. Although the proposed techniques are for classification, the results are shown on regression problems by discretizing the response attribute.
Wu and Jian \cite{wu2013new} present a procedure that performs MI on the incomplete dataset followed by non-parametric bootstrapping, which is much faster than performing bootstrapping followed by MI. Baneshi and Talei \cite{baneshi2012assessment} propose to perform MI using multiple imputation by chained equations (MICE) method on incomplete data followed by bootstrapping and the results are aggregated using statistical techniques. Tran et al. \cite{tran2017multiple} perform MI using MICE followed by bootstrapping with C4.5 decision tree as the base classifier. Their results show better performance in comparison to MI method to generate single imputed dataset and using three other single imputation methods to generate a complete dataset.
Valdiviezo and Van Aelst \cite{valdiviezo2015tree} combine missing data procedures with tree-based prediction methods after single and MI methods (MICE, MIST). They comment that if missingness is small, then single imputation is sufficient. However, if the missingness is moderate to large, then MI followed by tree-bagging is useful. 
Schomaker and Heumann \cite{schomaker2016bootstrap} comment that MI on bootstrapped samples and bootstrapped samples on multiple imputed datasets are the best options to calculate randomization valid confidence intervals when combining bootstrapping with MI. They further suggest that MI of bootstrap samples may be preferred for large imputation uncertainty (or low missingness) and bootstrapping of MI may be preferred for smaller imputation uncertainty (high missingness).

Other types of classifier fusion techniques are also explored by researchers to handle incomplete data. 
Su et al. \cite{su2009making} propose a classifier ensemble method to handle missing data. They started with an incomplete data, then further remove fixed percentage of attribute values to create different versions of the original incomplete data. They impute these datasets separately, present them to separate classifiers and combine their classification results. Their results suggest that ensemble learning with (bayesian) expectation maximization performs better than several single classifiers on many datasets. An issue with this approach is that, it removes more missing values from an incomplete data to create different datasets, which can compromise the accuracy of the method. 
Twala and Cartwright \cite{twala2005ensemble} present an ensemble method that imputes incomplete data using Bayesian MI and nearest neighbour imputation separately. These two imputations are fed to decision trees and their results combined. 
Nanni et al.\cite{nanni2012classifier} propose a MI approach that uses random subspaces method. Their general idea is to cluster incomplete data into a fixed number of clusters and then replace the missing values of missing data objects within a cluster with its center (or the mean of the cluster). This can reduce the information loss introduced by mean imputation if the full data is replaced by the mean vector. Several runs of random subspace is then performed on the imputed data to create an ensemble. Their method shows high performance on several health datasets and it does not drop when the missingness is increased to $30\%$.
Setz et al.\cite{setz2009using} present a classifier fusion of Linear and Quadratic classifiers with mean imputation and reduced feature modeling for emotion recognition task. Hassan et al. \cite{hassan2007regression} propose to perform MI several times to generate several samples of the original data and then feed them to classifiers and create an ensemble of several neural networks. They propose a univariate and multivariate version and show that they performed better than mean imputation and EM. 
Kumutha and Palaniammal \cite{kumutha2013enhanced} perform KNN imputation on gene expression data followed by bootstrapping. Khan et al. \cite{khan2012bayesian} propose a Bayesian MI ensemble method for one-class classification problems. They create two types of ensemble: one that averages the MI and trains a single classifier and the other that learns different classifiers on multiple imputed datasets. Their results show better performance of these methods in comparison to mean imputation as the missingness is increased.

The literature review shows that several ensemble methods exist to handle missing data while building generalizable classifiers. Bootstrapping the MI and MI of the bootstrap samples of the incomplete data are being used to learn better classifiers from incomplete data. Averaging MI and classifier fusion are other plausible techniques. Most of the research papers we reviewed did not compare different techniques of ensemble and study the effect on performance as the missingness in the data increases. These papers also did not provide insights into the diversity and accuracy of classifiers within an ensemble that might influence its performance.
In this paper, we consider three types of ensemble methods to handle incomplete data: (i) bootstrapping with single or average of MI, (ii) bootstrapping with MI, and (iii) ensemble of MI. These three approaches span different ways of creating diverse ensembles on incomplete data. Within each category, different types of imputation methods are used, such as mean imputation, Gaussian random imputation and expectation imputation (see Section \ref{sec:imputation} for details). 
Fusion of different types of classifiers is out of the scope of this paper. The different imputation methods used in this paper are described next.

\section{Imputation Methods}
\label{sec:methods}
Missingness can occur due to several reasons and can be of different types, such as Missing Completely at Random (MCAR), Missing at Random (MAR), Missing Not At Random (MNAR) \cite{Rubin1976}. 
 Rubin \cite{Rubin1976} proposed a topology for different kinds of missingness distributions. 
MAR allows the probabilities of missingness to depend on observed data but not on missing data. 
An important special case of MAR, called MCAR, occurs when the distribution does not depend on any value of the observed and missing data. 
MNAR is a situation that is neither MAR nor MCAR and arises when the distribution of missingness depends on the missing values in the data.
Mathematically, let the full data ($Y_{full}$) comprises of observed data ($Y_{obs}$) and missing data ($Y_{mis}$), i.e. 
\[Y_{full} = (Y_{obs}, Y_{mis}),\]
then missing data will be MAR, if
\begin{equation}
 P(X|Y_{full})=P(X|Y_{obs})
\end{equation}
where $X$ is a missingness indicator variable. $X=1$ when $Y$ is observed and $X=0$ when $Y$ is missing. Here $Y$ represents a group of items that is either entirely observed or entirely  missing. $X$ can be integer indicating the highest $j$ for which $Y_j$ is observed. $X$ can also be a matrix of binary indicators of the same dimension as the data. The missing data will be MCAR, if
\begin{equation}
 P(X|Y_{full})=P(X)
\end{equation}

In this paper, we investigate the use of the following four base imputation methods:

\begin{enumerate}
    \item \textbf{Mean Imputation} (MEI) -- In the MEI method, a missing attribute value is replaced by its mean \cite{khan2012bayesian}. If there are multiple missing values in an attribute, they all will be replaced by the same value because MEI gives one imputed value.
    
    \item  \textbf{Gaussian Random Imputation} (GRandI) -- In this method, we find the mean ($\mu$) and standard deviation ($\sigma$) of an attribute with missing values. Then we generate a uniformly distribute random standard normal variate ($z$) between $-Z$ and $+Z$. We use the following formula to impute a missing value
    \[ x = \sigma * z +  \mu \]
    Thus, the imputed value follows a Gaussian distribution. If there are multiple missing values in an attribute, they will not be imputed with same value because every time a different randomly chosen $z$ is generated. Similarly, if a missing value is imputed multiple times, GRandI will give different imputed values.

    \item  \textbf{Expectation Maximization Imputation (EMI)} Dempster, Laird and Rubin \cite{RubinMissingEMI} propose the use of an iterative solution, EM algorithm, for imputation for data with MAR missingness. The estimation or E-step of the EM algorithm computes the expected value of the sum of the variables with missing data assuming that we have a value for the population mean, and variance-covariance matrix. The maximization, or M-step, uses the expected value of the sum of a variable to estimate the population mean and covariance. When the fraction of missing values is large with one or more parameters, the convergence of this method is slower. EMI could be the slowest among other studied imputation methods in the paper when missingness ratio is large. Different initialization to EM produce different imputations for a missing value; hence, EMI can produce MI. 

\end{enumerate}

The above three methods are used as the base imputation methods in this paper. They will be combined in different ways to create ensembles, which is described in detail in Sections \ref{sec:imputation} and \ref{sec:bmi}. 
No Imputation (No-Imp) is the simplest method to handle missingness in the data, i.e., the incomplete data is not imputed for a given missingness ratio. In this case, a classifier is trained on the incomplete data. This serves as the baseline method to compare against other imputation approaches. We choose a C4.5 decision tree as the base classifier because it can handle missing attribute values.

\subsection{Single Imputation}
\label{sec:imputation}
Single imputation refers to the approaches that impute one value for a given missing value in incomplete data. 
In these methods, either a single value or multiple values are generated. If multiple values are generated, then their average is used as a single value and imputed in place of the missing value. After average imputation, one classifier can be trained on the complete data set. 
We use the following imputation methods for single imputation:

\begin{enumerate}
    \item MEI
    \item Average of GRandI
    \item Average of EMI 
\end{enumerate}

\begin{figure}[htb]
\centering
    \begin{subfigure}{0.23\textwidth}
  \centering
  \includegraphics[scale=0.3]{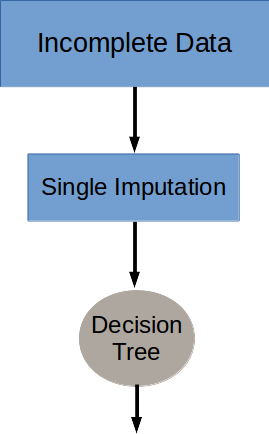} 
  \caption{Single Imputation} 
  \end{subfigure}
  \begin{subfigure}{0.23\textwidth}
  \centering
  \includegraphics[scale=0.3]{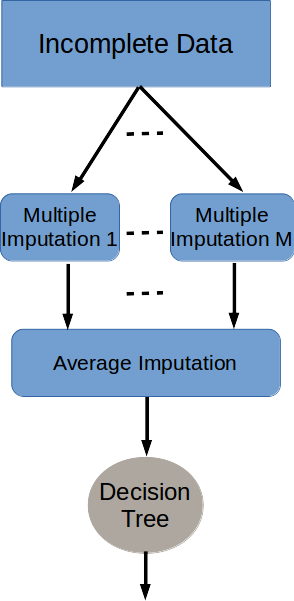} 
  \caption{Average of MI} 
  \end{subfigure}
  \caption{Two variations of Single Imputation}
  \label{fig:si}
\end{figure}

MEI imputes one value for a given missing value; therefore, there is no need to take an average. 
Whereas, GRandI and EMI generate multiple values for imputation. In these cases, the average of their MI is taken and a missing value is replaced by a single value. Both the approaches for single imputation and average of MI are shown in Figure \ref{fig:si}.

\section{Bagging and MI Ensemble}
\label{sec:bmi}

In a complete data set, ensemble approaches can improve the classification performance \cite{Kuncheva}. Bootstrapping or bagging is a popular ensemble learning approach where data is re-sampled with substitution several times \cite{Brieman, dahiya2017feature}. The reason for good performance of bagging is that it creates multiple datasets which lead to diverse and accurate classifiers.
In this paper, we consider bagging on incomplete data for ensemble learning. A C4.5 decision tree is used as a base classifier. Since C4.5 can handle missing values; it can be used to train a No-Imp equivalent of each ensemble methods for comparison purposes. Let us now define the following parameters that we will used to describe different ensemble approaches for missing data:
\begin{itemize}[label=\textbullet]
 \item $R$ -- missingness ratio,
  \item $M$ -- number of MI, and
  \item $B$ -- size of the ensemble.
\end{itemize}

We now discuss the three types of ensemble learning approaches to handle missing values.

\subsection{Bagging Single Imputation}

\begin{figure}[htb]
    \centering
    \includegraphics[scale=0.3]{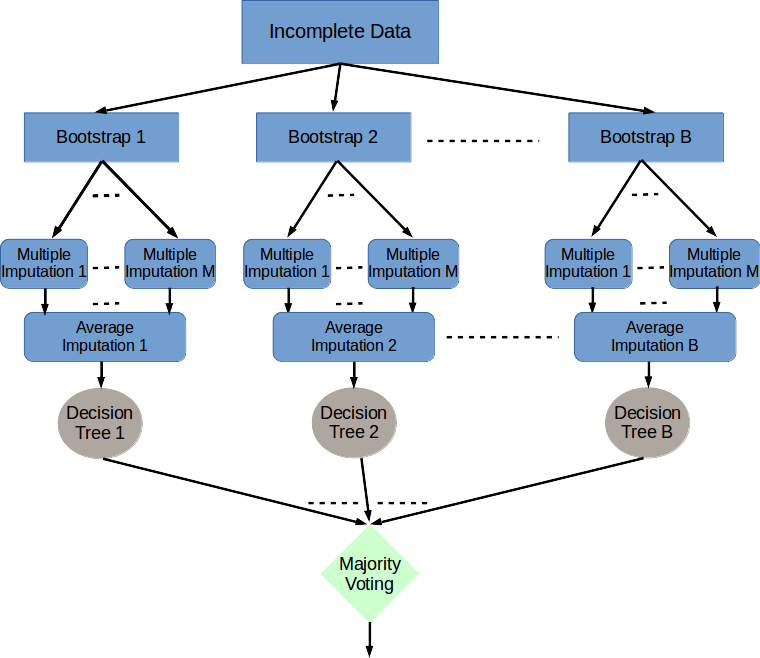}
    \caption{Bagging Single Imputation}
    \label{fig:BagSI}
\end{figure}

In this method, an incomplete data set is re-sampled $B$ times. This will result in $B$ sub-samples of the incomplete data set. Depending on the value of $R$, some sub-samples may be complete or incomplete. Then, we perform average MI (or equivalently, single imputation) on all the incomplete sub-samples and train a decision tree classifier on them. This leads to $B$ classifiers and a majority voting can be used to take a final decision. 
In summary, this method first performs bootstrapping on the incomplete data, followed by single imputation on each of the sub-samples.
Therefore, it retains the basic diversity aspect of bagging and replaces missing values with imputed values. Depending upon a particular imputation method, this can also lead to accurate classifiers. Combined with both the ideas of diversity and accuracy, we expect this method to perform better than the average imputation method (or equivalently, single imputation). The No-Imp equivalent method for this approach does not impute missing values in the $B$ sub-samples. Figure \ref{fig:BagSI} shows the different components of Bagging Single Imputation.

\subsection{Bagging MI}

\begin{figure}[htb]
    \centering
    \includegraphics[scale=0.3]{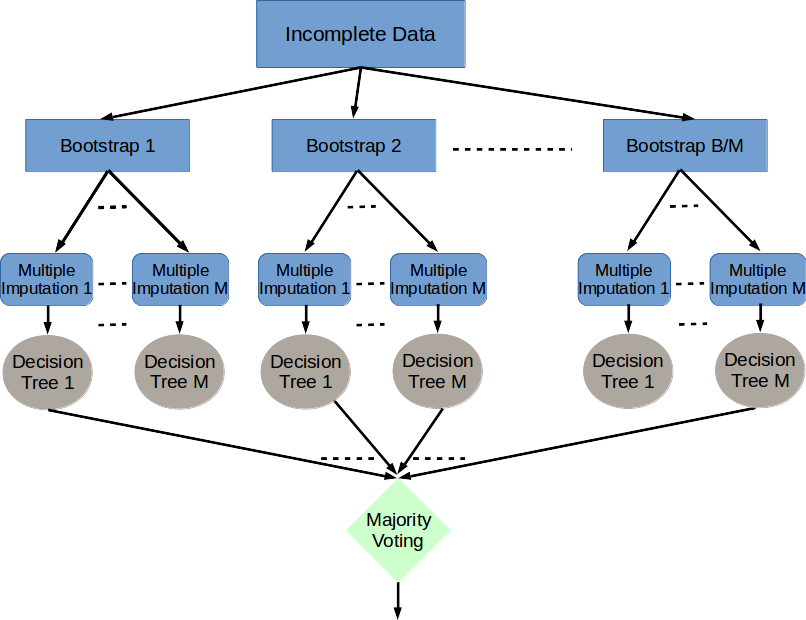}
    \caption{Bagging Multiple Imputation}
    \label{fig:BagMI}
\end{figure}

In this method, an incomplete data set is re-sampled $B/M$ times. Then, on each of these sub-samples, MI is performed $M$ times. This will result in $B$ imputed (complete) data sets. Thus, $B$ separate classifiers can be trained and their results combined with majority voting. As MEI does single imputation, only $B/M$ sub-samples will be different in this case and MI on these bootstrap samples will generate same imputed values. Therefore, MEI is excluded from this approach. There is no No-Imp equivalent of this method because, it will have only unique incomplete $B/M$ sub-samples and the rest of them will be duplicates. 
This method generates $B$ classifiers, which is the same number as Bagging Single Imputation method. Therefore, both the methods can be fairly compared when the base classifier is the same (C4.5 in our case).
It is to be noted that this method generates less diverse sub-samples than Bagging Single Imputation method; however, MI on these sub-samples can lead to more accurate classifier. To avoid numerical calculation problems, $B$ should be a multiple of $M$ in this method. Figure \ref{fig:BagMI} shows the steps involved in performing Bagging MI.

The other possibility is to perform MI on the incomplete data followed by bootstrapping each of those samples. In our case, we are using small value of MI, which means there will be higher uncertainty in the estimates. As commented by Schomaker and Heumann \cite{schomaker2016bootstrap}, bootstrapping of MI may be preferred for smaller imputation uncertainty (or moderate to large values of M). Therefore, we do not use this type of ensemble technique in this paper.

\subsection{MI Ensemble}

\begin{figure}[htb]
    \centering
    \includegraphics[scale=0.3]{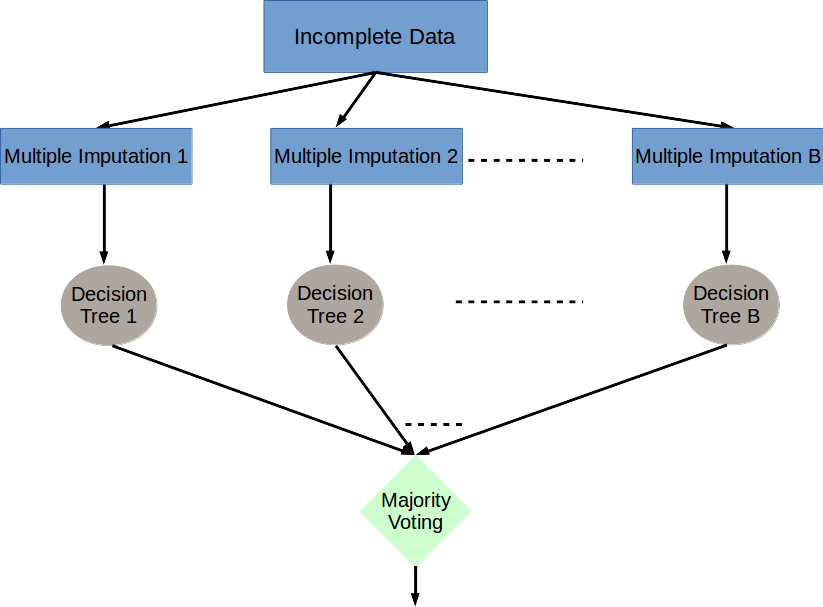}
    \caption{Multiple Imputation Ensemble}
    \label{fig:MI}
\end{figure}

In this method, MI is performed on the original incomplete data $M=B$ times. This is done to generate $B$ different copies of the incomplete dataset; hence $B$ classifiers can be trained and fair comparison can be done with the two above approaches. The results of these $B$ classifiers are combined using the majority voting method. This method will have the least diversity in comparison to the Bagging Single Imputation and Bagging MI because the same data is always used for imputing missing values. However, the individual classifiers may be more accurate if the underlying imputation method gives good estimates for missing values. The MEI does not impute multiple times and No-Imp method would result in $B$ duplicate copies of original incomplete data; therefore, both the methods cannot be applied while using MI ensemble. A graphical representation of MI Ensemble is shown in Figure \ref{fig:MI}

\subsection{Analysis}
The computational complexity of an ensemble method depends upon the number of datasets generated and imputation methods. In general, all three types of ensemble methods will be computationally expensive than running a single decision tree on an incomplete dataset. However, their performance is expected to be much higher due to the diversity and accuracy modeled by bootstrapping and MI ensemble. We keep the number of classifiers in all of these ensemble techniques to be same, so that no one technique may benefit from them and the comparisons are fair. That is, bootstrapping and MI, whether combined or not should always yield exactly $B$ imputed datasets for a given incomplete data.

Bagging Single Imputation generates $B$ bootstraps, perform $M$ imputations on each of these sub-samples and averages them to a single value. Therefore it generates $B+(B\times M)$ number of datasets.
Bagging MI generates $B/M$ bootstrap samples followed by $M$ number of MI on them; therefore, the number of datasets generated are $(B/M) + B$. Whereas, MI ensemble approach generates $B$ number of different copies of the original incomplete dataset by performing MI on them.
Performing MI takes the most time in creating an ensemble. 
We make sure that the number of decision tree classifiers among these ensemble methods remain the same; however, their computational complexity will be different due to the generation of different number of datasets.
Bagging Single Imputation generates the largest number of datasets, followed by Bagging MI and MI Ensemble. A comparison between these methods in terms of number of bootstrap samples generated, number of MI performed and number of datasets created by a given ensemble imputation method is shown in Table \ref{tab:bmiOP}. 

\begin{table*}
\centering
    \caption{Comparison of different ensemble imputation approaches.}
    \label{tab:bmiOP}
    \begin{tabular}{|l|l|l|l|}\hline
    \textbf{Method Name} & \textbf{\#Bootstrap Samples} & \textbf{\#MIs} & \textbf{\#Datasets Created}\\ \hline
    Bagging Single Imputation & $B$ & $B\times M$ & $B+(B\times M)$\\ \hline  
    Bagging MI & $B/M$ & $(B/M) \times M$ & $B+(B/M)$ \\ \hline
    MI Ensemble & $-$ & $B$ & $B$ \\ \hline
    \end{tabular}
\end{table*}

\section{Experimentation}
\label{sec:results}

We ran the programs on Intel i5-6200U, 2 cores CPU with 2.30GHz and 8GB RAM. We used Java SDK version 1.8 and Weka API Developer version 3.9.2 \cite{hall2009weka} to implement different imputation algorithms and the decision tree classifier. The C4.5 algorithm (\textit{J48} package in Weka) is used as a base decision tree classifier because it can handle missing values. Weka uses the \textit{EMImputation} package for EMI. The initial parameters in the original implementation of the \textit{EMImputation} package are fixed i.e. all means are zero, all variances are one, and all covariances are zero because the data is standardized. Therefore, this method always gives one fixed imputed value for a given missing value irrespective of multiple runs. This setting prevents variation in the MI for the EMI method. Therefore, we changed the code of \textit{EMImputation}, such that the elements of initial covariance matrix can randomly vary between $-1$ to $1$ (as the data is already standardized). This allows EMI to produce different values for every run of MI. The full source code along with the data sets used in this paper is available at \url{https://github.com/titubeta/EnsembleImputation}.

In this paper we discuss three base imputation methods (MEI, EMI, GRandI, see Section \ref{sec:methods}). The average imputations (or single imputation) of each of these base methods along with No-Imp give four methods (Methods $1-4$ in Table \ref{tab:datainformation}). Similarly, the Bagging Single Imputation approach gives three methods corresponding to each of the three base imputation imputation method and one corresponding to No-Imp method (Methods $5-8$ in Table \ref{tab:datainformation}). For both the Bagging MI and MI Ensemble, there are no No-Imp or MEI methods; therefore, they give two methods each corresponding to EMI and GRandI (Methods $9-12$ in Table \ref{tab:datainformation}). Therefore, in total, we compare $12$ different imputation methods, out of which 
\begin{itemize}[label=\textbullet]
 \item Eight are different ensemble imputation methods and four methods do not use ensemble imputation.
 \item Ten are imputation methods and two method does not involve any imputation (Methods $1$ and $5$ in Table \ref{tab:datainformation}).
\end{itemize}

\begin{table*}
\centering
\caption{{Acronyms for different imputation methods and their descriptions.}}
\begin{tabular}{|p{1.5cm}|p{2.5cm}|p{10cm}|}
\hline
\textbf{\#Method} & \textbf{Acronym} & \textbf{Description} \\ \hline  
1&No-Imp & No Imputation on incomplete data
\\2&MEI & Mean Imputation
\\3&GRandI & Average of Gaussian Random Imputation
\\4&EM & Average of Expectation Maximization imputation 
\\ \hline 5&BagNoImp & Ensemble by Bagging without imputation
\\6&BagMEI & Ensemble by Bagging with mean imputation 
\\7&BagGRandI&Ensemble by Bagging with Gaussian Random Imputation 
\\8&BagEM&Ensembles by Bagging with Expectation Maximization Imputation for each dataset
\\9&BagMIGRandI&Ensembles by MI over Bagging by Gaussian Random Imputation
\\10&BagMIEM&Ensembles by MI over Bagging by Expectation Maximization Imputation
\\11&MIGrandI&Ensembles by  MI by Gaussian Random Imputation
\\12&MIEM&Ensembles by  MI by Expectation Maximization Imputation
\\
\hline
\end{tabular}
\label{tab:datainformation}
\end{table*}

\subsection{Introducing Missing}
To introduce $R$ amount of missingness in the data, we adopt the following strategy. For every attribute of the data, following MCAR procedure, we randomly remove $R$ number of attribute values. The attributes values are removed such that the same attribute value is not removed more than once.  As the ratio of missingness increases, the probability for inducing missing values to an entire data object also increases. This situation, in particular, is problematic for EMI method because it will not impute such data objects and the same amount of training data may not be used for training the models. For MEI and GRandI, this situation is not a problem because they either replace all the missing values of the entire missing data object with the mean value of that attribute or with Gaussian distributed random value. To avoid this problem, we keep track of the last attribute while removing attribute values to check if such a case is happening. If a flag is set, then we do not remove that attribute value, rather set the index to the top of that feature and replace the first available non-missing attribute value. This will prevent removing all the attribute values of a data object.  
Therefore, if the number of features are $F$, then for a given missingness ratio $R$, a total of $F\times R$ attribute values will be removed. 

\subsection{Parameters}
The following values are set for the different parameters used in the experiments:
\begin{itemize}[label=\textbullet]
\item $R$ - missingness ratio is varied from $0\%, 5\%, 10\%,
\\15\%, 20\%, 25\%, 30\%$. It is to be noted that $0\%$ missingness means complete data with no missingness.
\item $M$ - number of imputations is set to $5$ \cite{khan2012bayesian}.
\item $B$ - size of ensemble is set to $25$ \cite{kuncheva2007classifier}.
\item $Z$ - for GRandI, it is set to $[-4, 4]$.
\item $CV$ - Number of cross validation folds is set to $2$.
\item $T$ - times to repeat the experiment to balance out random variations. It is set to $30$
\end{itemize}

A $2$-fold cross validation is performed for every imputation method (or corresponding No-Imp) and it is repeated $T$ times by randomizing the data. The average of performance across $T$ times $2$-folds is reported as the performance metric. 
Performance metrics used are accuracy and Kappa-error plots. Kappa-error  plots are used to study the accuracy and diversity of members of an ensemble.

\subsection{Datasets}
We use four datasets from the health domain and four from the general domain from the UCI data repository \cite{Lichman:2013} to evaluate different ensemble imputation methods. The description of these datasets is presented in Table \ref{tab:data}. Experiments are carried out with different types of datasets; from health and non-health domains to capture different types of classification problems, leading to generalizable results..

\begin{table*}
\centering
\caption{Description of the datasets. The feature values in all the datasets are numeric.}
\label{tab:data}
    \begin{tabular}{|l|l|l|l|l|} \hline
         \textbf{Domain} &\textbf{Dataset}    & \textbf{\# Data Objects} & \textbf{\# Features} & \textbf{\# Classes} \\ \hline
         \multirow{4}{*}{\rotatebox[origin=c]{90}{Health}} & Breast Tissues & $106$ & $10$ & $6$ \\ \cline{2-5}
                                 & New Thyroid & $215$ & $5$ & $3$\\ \cline{2-5}
                                 & Parkinsons & $197$ & $23$ & $2$\\ \cline{2-5}
                                 & Pima Indiana diabetes & $768$ & $8$ & $2$\\ \hline \hline         
         \multirow{4}{*}{\rotatebox[origin=c]{90}{Non-Health}} &  Column & $310$ &  $6$ & $3$\\ \cline{2-5}
                                 & Glass & $214$ & $10$ & $7$\\ \cline{2-5}
                                 & Seeds & $210$ & $7$ & $3 $\\ \cline{2-5}
                                 & Wine & $178$ & $13$ & $3$\\ \hline
    \end{tabular}
\end{table*}

\subsection{Results}

Tables \ref{tab:breast} - \ref{tab:wine} show the results for each of the $8$ datasets. In each table, the first column represents the imputation method. The subsequent columns show the accuracy of each imputation method as the missingness ratio is increased from complete data ($0\%$) to $30\%$. To compare various classification methods, we perform Friedman's rank sum test \cite{Demsar:2006:SCC:1248547.1248548}. Calvo and Rodrigo recommend the Bergmann and Hommel test as a post-hoc all pair-wise comparison method with high statistical power \cite{statisticaltest}. We used their \textit{R} package (\textit{scmamp}) is this paper \cite{statisticaltestpackage}. This test computes the p-value for each pair of algorithms corrected for multiple testing using Bergman and Hommel's correction. Results are presented in Tables \ref{0testing} and \ref{30testing}. The significant differences (p-value $<$ 0.05) are shown in bold. We summarize the results from these tables as follows:

\begin{enumerate}
    \item Presence of a large number of missing values deteriorate the performance of a decision tree classification algorithm with single imputation. For example, for Breast Tissue data, the accuracy of EM algorithm reduce to $0.524$ from $0.629$ when $30\%$ missingness was introduced in comparison to complete data. Similar performance degradation was observed for all the methods for all the datasets. 
    
    \item All the ensemble imputation methods performed better than their corresponding single imputation methods for $10\%$ or more missingness ratio. Ensemble methods showed that they are more robust as the missingness ratio is increased as compared to the corresponding single imputation methods. For example, for Breast Tissues dataset with $30\%$ missingness ratio, the accuracy of single EM method decreases by around $17\%$ (0.629 to 0.524), whereas the accuracy of Bagging Single Imputation with EM method decreases by around $5\%$ (0.648 to 0.614). 
    
    
     \item For smaller missingness ratio (up to $10\%$), MI over bootstrap and MI ensemble with MEI and GRandI showed no significant superiority over each other. 
     They also perform worse than or are equivalent to methods that use imputation on bootstrap samples of incomplete data. It is to be noted that MEI or GRandI are less computationally extensive than EM. 
     
     \item For smaller missingness ratio (up to $10\%$), bootstrapping of incomplete data without imputation generally show similar performance to other ensemble imputation techniques. However, ensemble imputation methods can have slight advantage for some datasets.
     
     \item Overall, the methods BagEM and  BagMIEM  that combine bagging with EM emerge as the best choice due to their robust performance on high missingness ratio (up to 30\%). For example, with 30\% missingness, for Breast Tissues dataset with no-imputation the accuracy degrades from 0.629 to 0.468 whereas the accuracy of BagEM degrades from 0.648 to 0.614 and the accuracy of BagMIEM degrades from 0.625 to 0.608. Similar behaviour is observed for other datasets. The statistical test for classifiers for 30\% missingness (Table \ref{30testing}) suggests that BagEM has advantage over BagMIEM as BagEM shows statistically better results against most of the classification methods. However, a large number of missingness ratio means that the EM method will take more time to impute missing values. 
     
\end{enumerate}
In the next section, we will study the performance of different ensemble methods by using kappa-error plots.

\begin{table*}[htb]
\centering
\caption{Breast Tissues data}
\label{tab:breast}
    \begin{tabular}{|p{25mm}|l|l|l|l|l|l|l|} \hline
     Imputation Methods    & \multicolumn{7}{c|}{Missingness Ratio ($\%$)} \\ \cline{2-8}
                           &  0 & 5 & 10 & 15 & 20 & 25 & 30 \\ \hline
No-Imp & 0.629 & 0.581 & 0.565 & 0.554 & 0.516 & 0.51 & 0.468 \\ \hline
MEI & 0.629 & 0.579 & 0.548 & 0.525 & 0.504 & 0.495 & 0.474 \\ \hline
GRandI & 0.629 & 0.576 & 0.544 & 0.513 & 0.468 & 0.436 & 0.418 \\ \hline
EM & 0.629 & 0.625 & 0.61 & 0.6 & 0.586 & 0.553 & 0.524 \\ \hline \hline
BagNoImp & 0.648 & 0.614 & 0.593 & 0.591 & 0.562 & 0.541 & 0.525 \\ \hline
BagMEI & 0.648 & 0.613 & 0.583 & 0.576 & 0.553 & 0.536 & 0.526 \\ \hline
BagGRandI & 0.648 & 0.633 & 0.614 & 0.622 & 0.595 & 0.561 & 0.535 \\ \hline
BagEM & 0.648 & 0.642 & 0.638 & 0.647 & 0.64 & 0.625 & 0.614 \\ \hline \hline
BagMIGRandI & 0.625 & 0.615 & 0.612 & 0.607 & 0.577 & 0.567 & 0.539 \\ \hline
BagMIEM & 0.625 & 0.624 & 0.626 & 0.631 & 0.629 & 0.626 & 0.608 \\ \hline \hline
MIGrandI & 0.629 & 0.607 & 0.609 & 0.606 & 0.584 & 0.56 & 0.55 \\ \hline
MIEM & 0.629 & 0.626 & 0.622 & 0.614 & 0.617 & 0.62 & 0.594 \\ \hline
\end{tabular}
\end{table*}


\begin{table*}[htb]
\centering
\caption{New Thyroid data}
\label{tab:thyroid}
    \begin{tabular}{|p{25mm}|l|l|l|l|l|l|l|} \hline
     Imputation Methods    & \multicolumn{7}{c|}{Missingness Ratio ($\%$)} \\ \cline{2-8}
                           &  0 & 5 & 10 & 15 & 20 & 25 & 30 \\ \hline
No-Imp & 0.915 & 0.906 & 0.902 & 0.892 & 0.881 & 0.867 & 0.857 \\ \hline
MEI & 0.915 & 0.907 & 0.899 & 0.885 & 0.888 & 0.877 & 0.862 \\ \hline
GRandI & 0.915 & 0.887 & 0.855 & 0.83 & 0.798 & 0.772 & 0.74 \\ \hline
EM & 0.915 & 0.906 & 0.904 & 0.898 & 0.886 & 0.866 & 0.855 \\ \hline \hline
BagNoImp & 0.929 & 0.921 & 0.919 & 0.902 & 0.896 & 0.884 & 0.867 \\ \hline
BagMEI & 0.929 & 0.918 & 0.913 & 0.902 & 0.901 & 0.893 & 0.875 \\ \hline
BagGRandI & 0.929 & 0.93 & 0.923 & 0.902 & 0.89 & 0.881 & 0.855 \\ \hline
BagEM & 0.929 & 0.922 & 0.927 & 0.922 & 0.919 & 0.915 & 0.907 \\ \hline \hline
BagMIGRandI & 0.924 & 0.928 & 0.917 & 0.902 & 0.892 & 0.874 & 0.859 \\ \hline
BagMIEM & 0.924 & 0.921 & 0.916 & 0.912 & 0.912 & 0.907 & 0.9 \\ \hline \hline
MIGrandI & 0.915 & 0.919 & 0.916 & 0.906 & 0.898 & 0.888 & 0.866 \\ \hline
MIEM & 0.915 & 0.907 & 0.908 & 0.899 & 0.893 & 0.883 & 0.878 \\ \hline
\end{tabular}
\end{table*}


\begin{table*}[htb]
\centering
\caption{Parkinsons data}
\label{tab:parkinson}
    \begin{tabular}{|p{25mm}|l|l|l|l|l|l|l|} \hline
     Imputation Methods    & \multicolumn{7}{c|}{Missingness Ratio ($\%$)} \\ \cline{2-8}
                           &  0 & 5 & 10 & 15 & 20 & 25 & 30 \\ \hline
No-Imp & 0.835 & 0.834 & 0.822 & 0.827 & 0.814 & 0.809 & 0.809 \\ \hline
MEI & 0.835 & 0.824 & 0.81 & 0.806 & 0.796 & 0.79 & 0.794 \\ \hline
GRandI & 0.831 & 0.813 & 0.787 & 0.781 & 0.754 & 0.732 & 0.724 \\ \hline
EM & 0.831 & 0.833 & 0.824 & 0.829 & 0.809 & 0.804 & 0.785 \\ \hline \hline
BagNoImp & 0.862 & 0.854 & 0.846 & 0.846 & 0.833 & 0.824 & 0.823 \\ \hline
BagMEI & 0.862 & 0.856 & 0.845 & 0.837 & 0.841 & 0.825 & 0.828 \\ \hline
BagGRandI & 0.862 & 0.863 & 0.852 & 0.843 & 0.838 & 0.828 & 0.818 \\ \hline
BagEM & 0.862 & 0.865 & 0.859 & 0.858 & 0.857 & 0.85 & 0.845 \\ \hline \hline
BagMIGRandI & 0.849 & 0.856 & 0.845 & 0.838 & 0.832 & 0.822 & 0.816 \\ \hline
BagMIEM & 0.849 & 0.852 & 0.841 & 0.845 & 0.852 & 0.843 & 0.841 \\ \hline \hline
MIGrandI & 0.835 & 0.861 & 0.85 & 0.848 & 0.836 & 0.832 & 0.827 \\ \hline
MIEM & 0.835 & 0.836 & 0.827 & 0.829 & 0.827 & 0.836 & 0.838 \\ \hline
\end{tabular}
\end{table*}

\begin{table*}[htb]
\centering
\caption{Pima Indiana Diabetes data}
\label{tab:pima}
    \begin{tabular}{|p{25mm}|l|l|l|l|l|l|l|} \hline
     Imputation Methods    & \multicolumn{7}{c|}{Missingness Ratio ($\%$)} \\ \cline{2-8}
                           &  0 & 5 & 10 & 15 & 20 & 25 & 30 \\ \hline
No-Imp & 0.731 & 0.734 & 0.723 & 0.723 & 0.719 & 0.713 & 0.706 \\ \hline
MEI & 0.731 & 0.727 & 0.713 & 0.711 & 0.7 & 0.695 & 0.691 \\ \hline
GRandI & 0.731 & 0.713 & 0.701 & 0.695 & 0.681 & 0.672 & 0.666 \\ \hline
EM & 0.731 & 0.721 & 0.719 & 0.713 & 0.71 & 0.698 & 0.69 \\ \hline \hline
BagNoImp & 0.754 & 0.75 & 0.746 & 0.74 & 0.737 & 0.724 & 0.721 \\ \hline
BagMEI & 0.754 & 0.749 & 0.741 & 0.735 & 0.731 & 0.721 & 0.714 \\ \hline
BagGRandI & 0.754 & 0.747 & 0.743 & 0.734 & 0.728 & 0.722 & 0.713 \\ \hline
BagEM & 0.754 & 0.743 & 0.744 & 0.742 & 0.736 & 0.725 & 0.721 \\ \hline \hline
BagMIGRandI & 0.733 & 0.739 & 0.735 & 0.731 & 0.727 & 0.713 & 0.709 \\ \hline
BagMIEM & 0.733 & 0.731 & 0.725 & 0.727 & 0.722 & 0.715 & 0.717 \\ \hline \hline
MIGrandI & 0.731 & 0.736 & 0.738 & 0.729 & 0.721 & 0.718 & 0.709 \\ \hline
MIEM & 0.731 & 0.72 & 0.716 & 0.717 & 0.71 & 0.703 & 0.705 \\ \hline
\end{tabular}
\end{table*}

\begin{table*}[htb]
\centering
\caption{Column data}
\label{tab:column}
    \begin{tabular}{|p{25mm}|l|l|l|l|l|l|l|} \hline
     Imputation Methods    & \multicolumn{7}{c|}{Missingness Ratio ($\%$)} \\ \cline{2-8}
                           &  0 & 5 & 10 & 15 & 20 & 25 & 30 \\ \hline
No-Imp & 0.804 & 0.785 & 0.767 & 0.751 & 0.735 & 0.715 & 0.699 \\ \hline
MEI & 0.804 & 0.771 & 0.762 & 0.732 & 0.725 & 0.71 & 0.698 \\ \hline
GRandI & 0.804 & 0.767 & 0.734 & 0.704 & 0.673 & 0.664 & 0.62 \\ \hline
EM & 0.804 & 0.791 & 0.781 & 0.756 & 0.74 & 0.72 & 0.703 \\ \hline \hline
BagNoImp & 0.83 & 0.81 & 0.79 & 0.769 & 0.761 & 0.737 & 0.721 \\ \hline
BagMEI & 0.83 & 0.803 & 0.789 & 0.773 & 0.766 & 0.754 & 0.731 \\ \hline
BagGRandI & 0.83 & 0.807 & 0.788 & 0.772 & 0.764 & 0.75 & 0.725 \\ \hline
BagEM & 0.83 & 0.817 & 0.81 & 0.792 & 0.788 & 0.775 & 0.758 \\ \hline \hline
BagMIGRandI & 0.819 & 0.8 & 0.787 & 0.77 & 0.764 & 0.744 & 0.716 \\ \hline
BagMIEM & 0.819 & 0.804 & 0.798 & 0.782 & 0.779 & 0.759 & 0.748 \\ \hline \hline
MIGrandI & 0.804 & 0.788 & 0.783 & 0.761 & 0.757 & 0.738 & 0.717 \\ \hline
MIEM & 0.804 & 0.79 & 0.782 & 0.76 & 0.754 & 0.729 & 0.717 \\ \hline
\end{tabular}
\end{table*}


\begin{table*}[htb]
\centering
\caption{Glass data}
\label{tab:glass}
    \begin{tabular}{|p{25mm}|l|l|l|l|l|l|l|} \hline
     Imputation Methods    & \multicolumn{7}{c|}{Missingness Ratio ($\%$)} \\ \cline{2-8}
                           &  0 & 5 & 10 & 15 & 20 & 25 & 30 \\ \hline
No-Imp & 0.648 & 0.623 & 0.607 & 0.59 & 0.575 & 0.546 & 0.532 \\ \hline
MEI & 0.648 & 0.626 & 0.594 & 0.578 & 0.557 & 0.531 & 0.519 \\ \hline
GRandI & 0.648 & 0.592 & 0.548 & 0.499 & 0.48 & 0.447 & 0.419 \\ \hline
EM & 0.648 & 0.637 & 0.62 & 0.61 & 0.594 & 0.557 & 0.511 \\ \hline \hline
BagNoImp & 0.699 & 0.683 & 0.66 & 0.64 & 0.617 & 0.592 & 0.587 \\ \hline
BagMEI & 0.699 & 0.682 & 0.652 & 0.636 & 0.618 & 0.601 & 0.587 \\ \hline
BagGRandI & 0.699 & 0.682 & 0.659 & 0.639 & 0.611 & 0.583 & 0.567 \\ \hline
BagEM & 0.699 & 0.694 & 0.682 & 0.68 & 0.668 & 0.654 & 0.633 \\ \hline \hline
BagMIGRandI & 0.667 & 0.67 & 0.646 & 0.629 & 0.601 & 0.579 & 0.557 \\ \hline
BagMIEM & 0.667 & 0.667 & 0.66 & 0.646 & 0.636 & 0.638 & 0.622 \\ \hline \hline
MIGrandI & 0.648 & 0.661 & 0.649 & 0.633 & 0.615 & 0.59 & 0.577 \\ \hline
MIEM & 0.648 & 0.637 & 0.618 & 0.612 & 0.595 & 0.593 & 0.592 \\ \hline
\end{tabular}
\end{table*}

\begin{table*}[htb]
\centering
\caption{Seeds data}
\label{tab:seeds}
    \begin{tabular}{|p{25mm}|l|l|l|l|l|l|l|} \hline
     Imputation Methods    & \multicolumn{7}{c|}{Missingness Ratio ($\%$)} \\ \cline{2-8}
                           &  0 & 5 & 10 & 15 & 20 & 25 & 30 \\ \hline
No-Imp & 0.891 & 0.864 & 0.854 & 0.834 & 0.83 & 0.818 & 0.806 \\ \hline
MEI & 0.891 & 0.858 & 0.848 & 0.842 & 0.828 & 0.817 & 0.808 \\ \hline
GRandI & 0.891 & 0.845 & 0.806 & 0.779 & 0.752 & 0.722 & 0.686 \\ \hline
EM & 0.891 & 0.893 & 0.889 & 0.873 & 0.874 & 0.853 & 0.836 \\ \hline \hline
BagNoImp & 0.903 & 0.885 & 0.883 & 0.872 & 0.867 & 0.853 & 0.851 \\ \hline
BagMEI & 0.903 & 0.872 & 0.864 & 0.865 & 0.854 & 0.846 & 0.842 \\ \hline
BagGRandI & 0.903 & 0.888 & 0.883 & 0.871 & 0.868 & 0.859 & 0.854 \\ \hline
BagEM & 0.903 & 0.901 & 0.898 & 0.896 & 0.89 & 0.889 & 0.888 \\ \hline \hline
BagMIGRandI & 0.892 & 0.877 & 0.874 & 0.865 & 0.861 & 0.859 & 0.843 \\ \hline
BagMIEM & 0.892 & 0.889 & 0.89 & 0.884 & 0.886 & 0.888 & 0.883 \\ \hline \hline
MIGrandI & 0.891 & 0.876 & 0.882 & 0.871 & 0.866 & 0.864 & 0.852 \\ \hline
MIEM & 0.891 & 0.893 & 0.891 & 0.881 & 0.881 & 0.875 & 0.874 \\ \hline
\end{tabular}
\end{table*}

\begin{table*}[htb]
\centering
\caption{Wine data}
\label{tab:wine}
    \begin{tabular}{|p{25mm}|l|l|l|l|l|l|l|} \hline
     Imputation Methods    & \multicolumn{7}{c|}{Missingness Ratio ($\%$)} \\ \cline{2-8}
                           &  0 & 5 & 10 & 15 & 20 & 25 & 30 \\ \hline
No-Imp & 0.889 & 0.883 & 0.861 & 0.851 & 0.84 & 0.827 & 0.809 \\ \hline
MEI & 0.889 & 0.875 & 0.859 & 0.85 & 0.838 & 0.825 & 0.81 \\ \hline
GRandI & 0.889 & 0.861 & 0.827 & 0.788 & 0.765 & 0.714 & 0.672 \\ \hline
EM & 0.889 & 0.89 & 0.884 & 0.88 & 0.864 & 0.834 & 0.805 \\ \hline \hline
BagNoImp & 0.919 & 0.917 & 0.913 & 0.902 & 0.896 & 0.88 & 0.868 \\ \hline
BagMEI & 0.919 & 0.911 & 0.903 & 0.899 & 0.886 & 0.88 & 0.868 \\ \hline
BagGRandI & 0.919 & 0.931 & 0.929 & 0.922 & 0.916 & 0.893 & 0.881 \\ \hline
BagEM & 0.919 & 0.92 & 0.913 & 0.92 & 0.928 & 0.924 & 0.921 \\ \hline \hline
BagMIGRandI & 0.901 & 0.917 & 0.921 & 0.911 & 0.907 & 0.885 & 0.881 \\ \hline
BagMIEM & 0.901 & 0.897 & 0.896 & 0.898 & 0.905 & 0.911 & 0.913 \\ \hline \hline
MIGrandI & 0.889 & 0.91 & 0.915 & 0.921 & 0.913 & 0.894 & 0.886 \\ \hline
MIEM & 0.889 & 0.892 & 0.884 & 0.884 & 0.885 & 0.862 & 0.89 \\ \hline
                         
\end{tabular}
\end{table*}

\begin{table*}[!ht]
\centering
\caption{Friedman's post-hoc test with Bergmann and Hommel's correction with complete data (or no missingness). The significant differences (p-value $<$ 0.05) are shown in bold.}
 \begin{tabular}{|l|llllllll|}
    \hline
     & BagNoImp & BagMEI & BagGRandI & BagEM & BagMIGRandI & BagMIEM & MIGrandI & MIEM \\
    \hline
    BagNoImp & n/a & 1.000 & 1.000 & 1.000 & 0.127 & 0.127 & {\bf 0.003} & {\bf 0.003} \\
    BagMEI & 1.000 & n/a & 1.000 & 1.000 & 0.127 & 0.127 & {\bf 0.003} & {\bf 0.003} \\
    BagGRandI & 1.000 & 1.000 & n/a & 1.000 & 0.127 & 0.127 & {\bf 0.003} & {\bf 0.003} \\
    BagEM & 1.000 & 1.000 & 1.000 & n/a & 0.127 & 0.127 & {\bf 0.003} & {\bf 0.003} \\
    BagMIGRandI & 0.127 & 0.127 & 0.127 & 0.127 & n/a & 1.000 & 1.000 & 1.000 \\
    BagMIEM & 0.127 & 0.127 & 0.127 & 0.127 & 1.000 & n/a & 1.000 & 1.000 \\
    MIGrandI & {\bf 0.003} & {\bf 0.003} & {\bf 0.003} & {\bf 0.003} & 1.000 & 1.000 & n/a & 1.000 \\
    MIEM & {\bf 0.003} & {\bf 0.003} & {\bf 0.003} & {\bf 0.003} & 1.000 & 1.000 & 1.000 & n/a \\
    \hline
    \end{tabular}
    \label{0testing}
\end{table*}

\begin{table*}[!ht]
\centering
\caption{Friedman's post-hoc test with Bergmann and Hommel's correction with 30\% missingness ratio. The significant differences (p-value $<$ 0.05) are shown in bold.}
    \begin{tabular}{|l|llllllll|}
    \hline
     & BagNoImp & BagMEI & BagGRandI & BagEM & BagMIGRandI & BagMIEM & MIGrandI & MIEM \\
    \hline
    BagNoImp & n/a & 1.000 & 1.000 & {\bf 0.006} & 1.000 & 0.075 & 1.000 & 1.000 \\
    BagMEI & 1.000 & n/a & 1.000 & {\bf 0.008} & 1.000 & 0.118 & 1.000 & 1.000 \\
    BagGRandI & 1.000 & 1.000 & n/a & {\bf 0.002} & 1.000 & {\bf 0.039} & 1.000 & 1.000 \\
    BagEM & {\bf 0.006} & {\bf 0.008} & {\bf 0.002} & n/a & {\bf 0.000} & 1.000 & {\bf 0.006} & 0.186 \\
    BagMIGRandI & 1.000 & 1.000 & 1.000 & {\bf 0.000} & n/a & {\bf 0.002} & 1.000 & 0.346 \\
    BagMIEM & 0.075 & 0.118 & {\bf 0.039} & 1.000 & {\bf 0.002} & n/a & 0.088 & 1.000 \\
    MIGrandI & 1.000 & 1.000 & 1.000 & {\bf 0.006} & 1.000 & 0.088 & n/a & 1.000 \\
    MIEM & 1.000 & 1.000 & 1.000 & 0.186 & 0.346 & 1.000 & 1.000 & n/a \\
    \hline
    \end{tabular}
    \label{30testing}
\end{table*}

\begin{sidewaystable}[htb]
\caption{Kappa-error plots for 10\% missingness for different ensemble methods. The  most  desirable  pairs  of classifiers will lie at the bottom left corner (high diversity and low average error).}
\scalebox{0.9}{
\begin{tabular}{p{2cm}cccccccc}
Dataset & BagNoImp & BagMEI & BagGRandI & BagEM & BagMIGRandI & BagMIEMI & MIGRandI & MIEMI \\ \hline
Breast Tissue &
\begin{subfigure}{0.1\textwidth}
    \includegraphics[height=10.5mm]{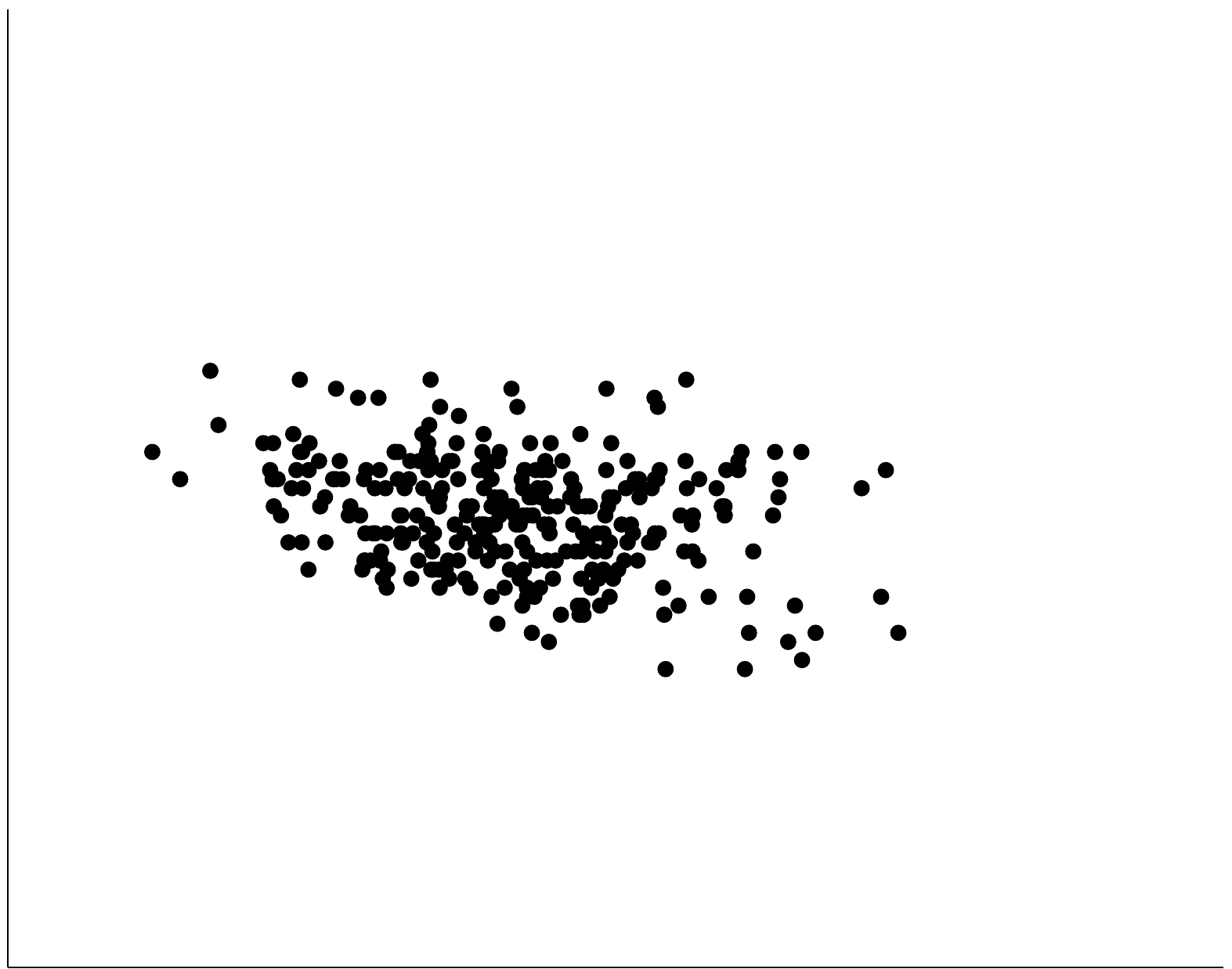}
\end{subfigure}&
\begin{subfigure}{0.1\textwidth}
    \includegraphics[height=10.5mm]{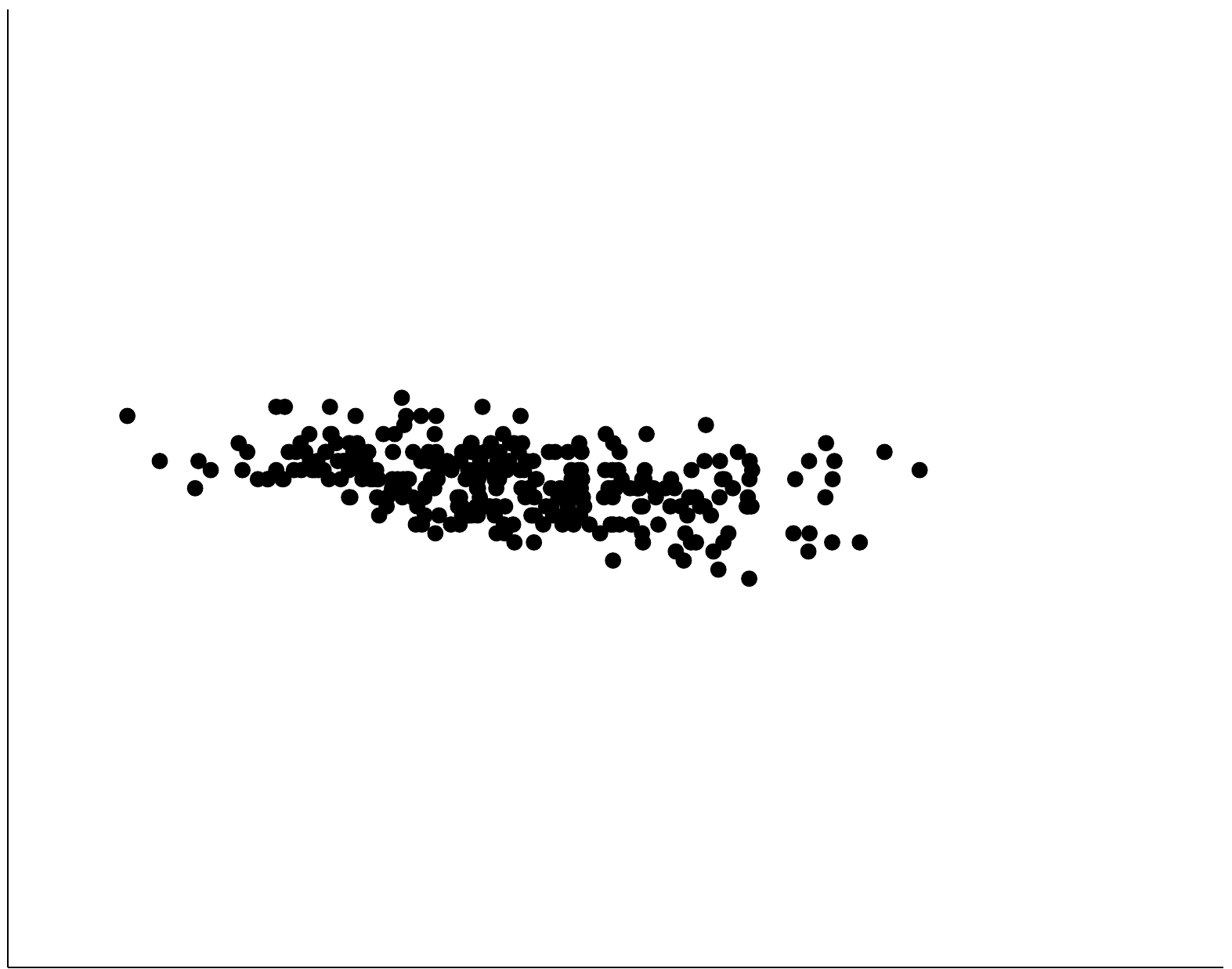}
\end{subfigure}&
\begin{subfigure}{0.1\textwidth}
    \includegraphics[height=10.5mm]{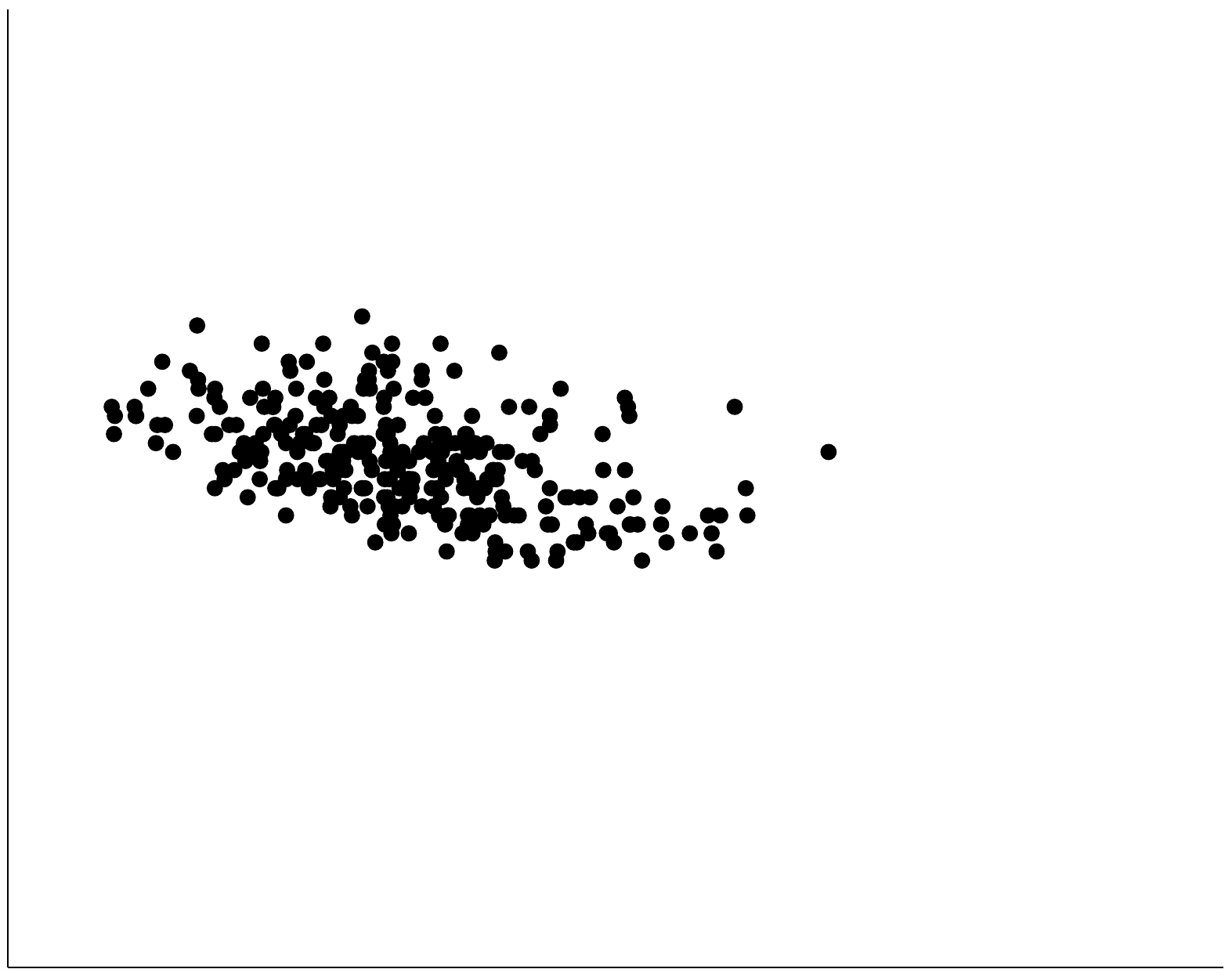}
\end{subfigure}&
\begin{subfigure}{0.1\textwidth}
    \includegraphics[height=10.5mm]{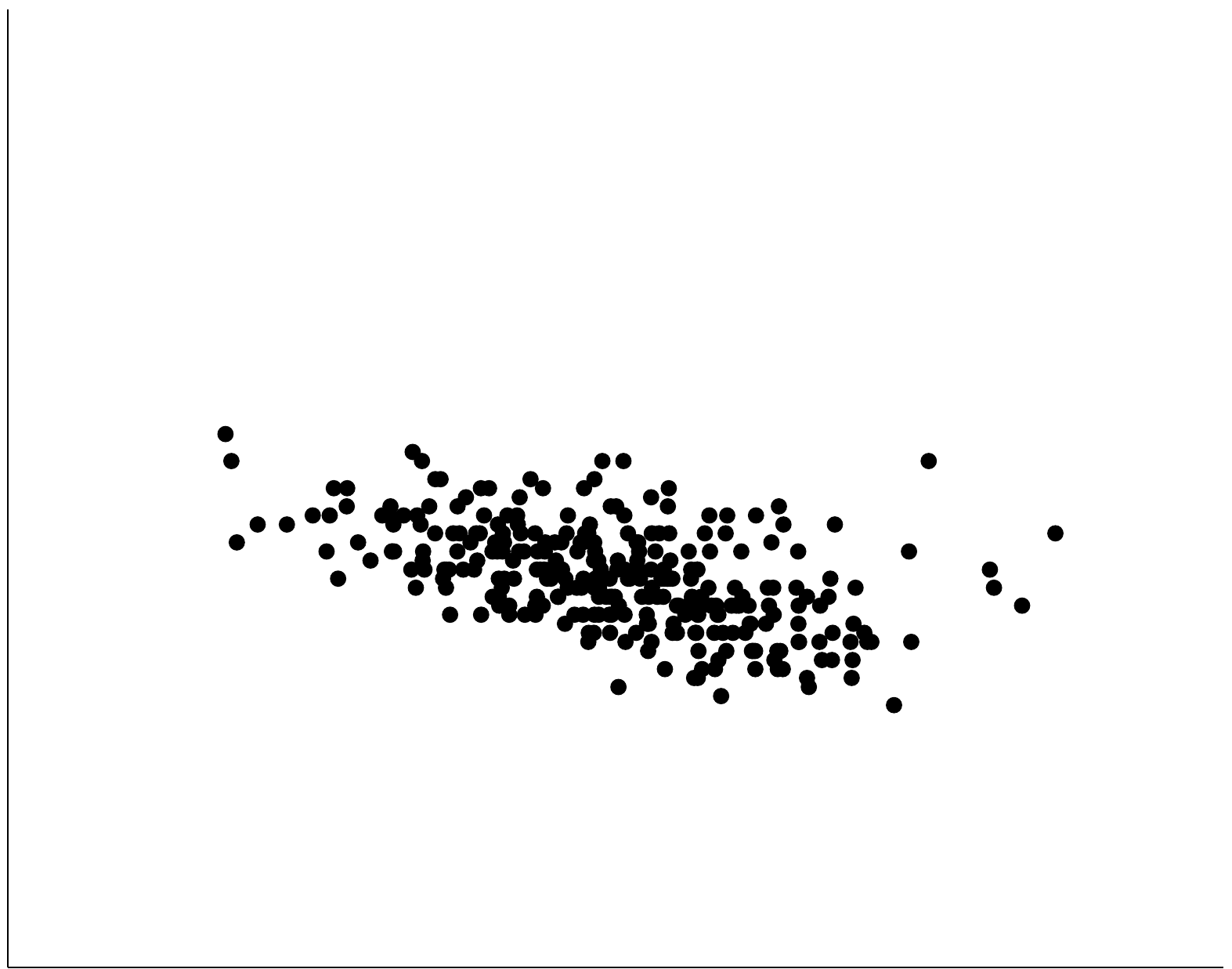}
\end{subfigure}&
\begin{subfigure}{0.1\textwidth}
    \includegraphics[height=10.5mm]{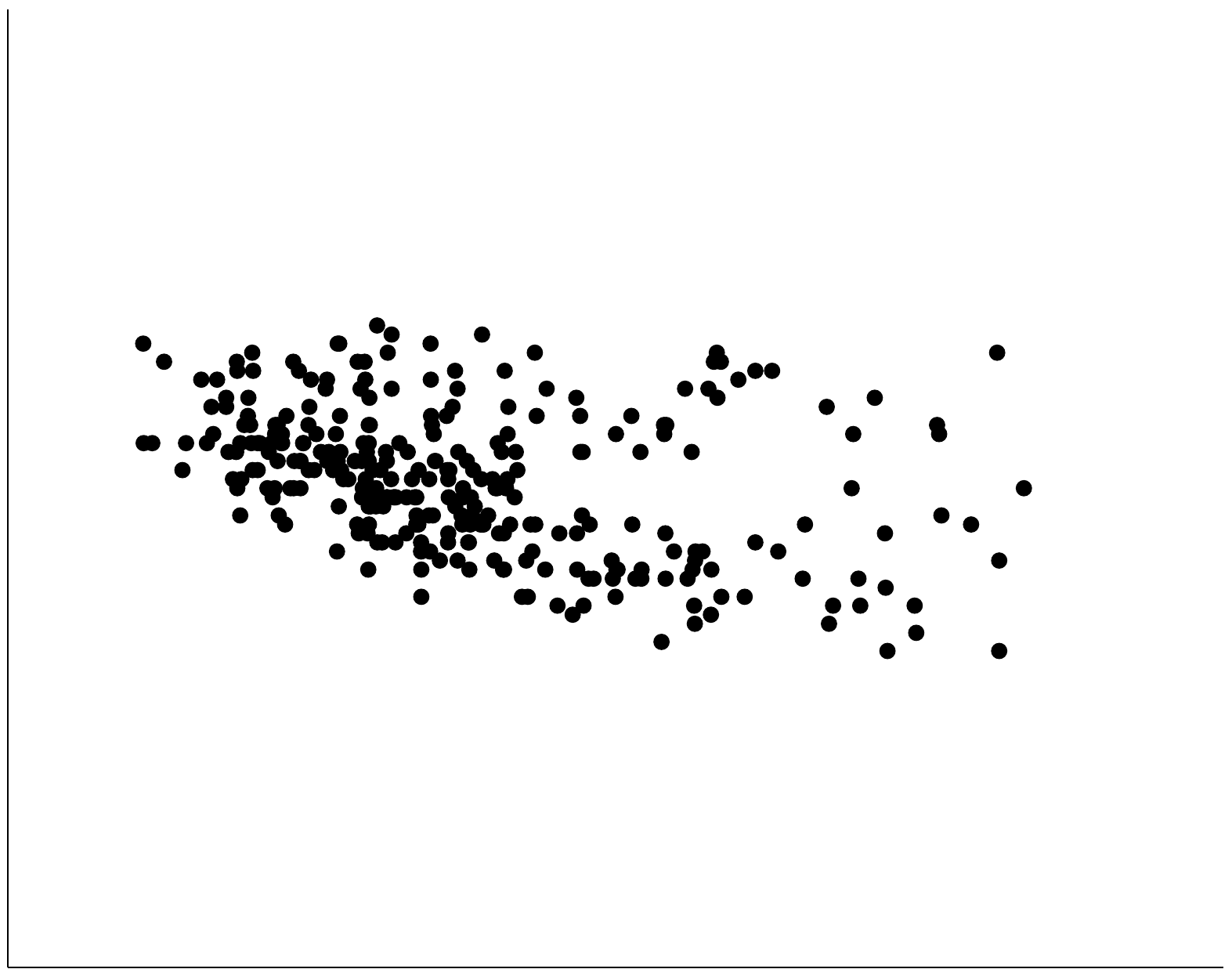}
\end{subfigure}&
\begin{subfigure}{0.1\textwidth}
    \includegraphics[height=10.5mm]{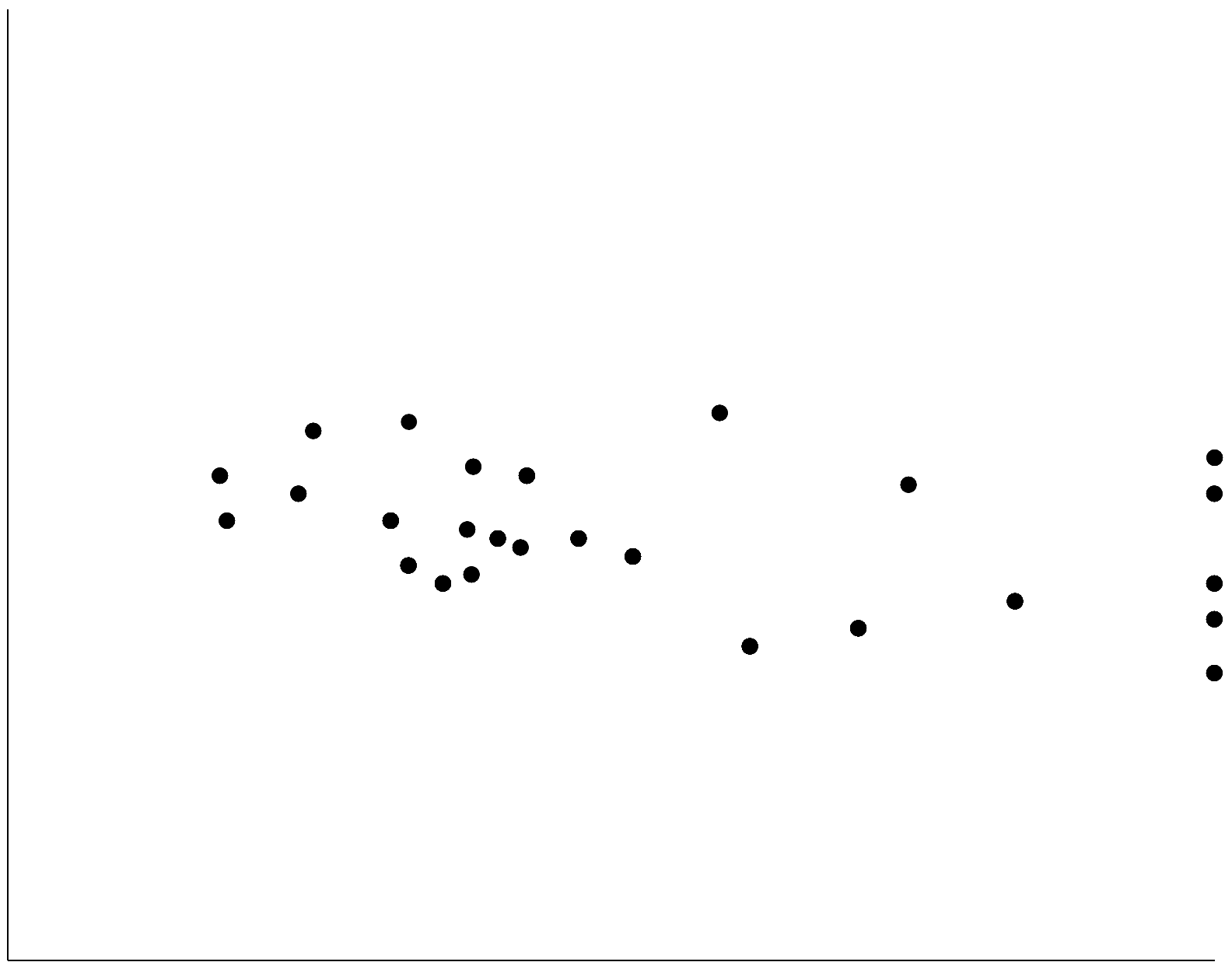}
\end{subfigure}&
\begin{subfigure}{0.1\textwidth}
    \includegraphics[height=10.5mm]{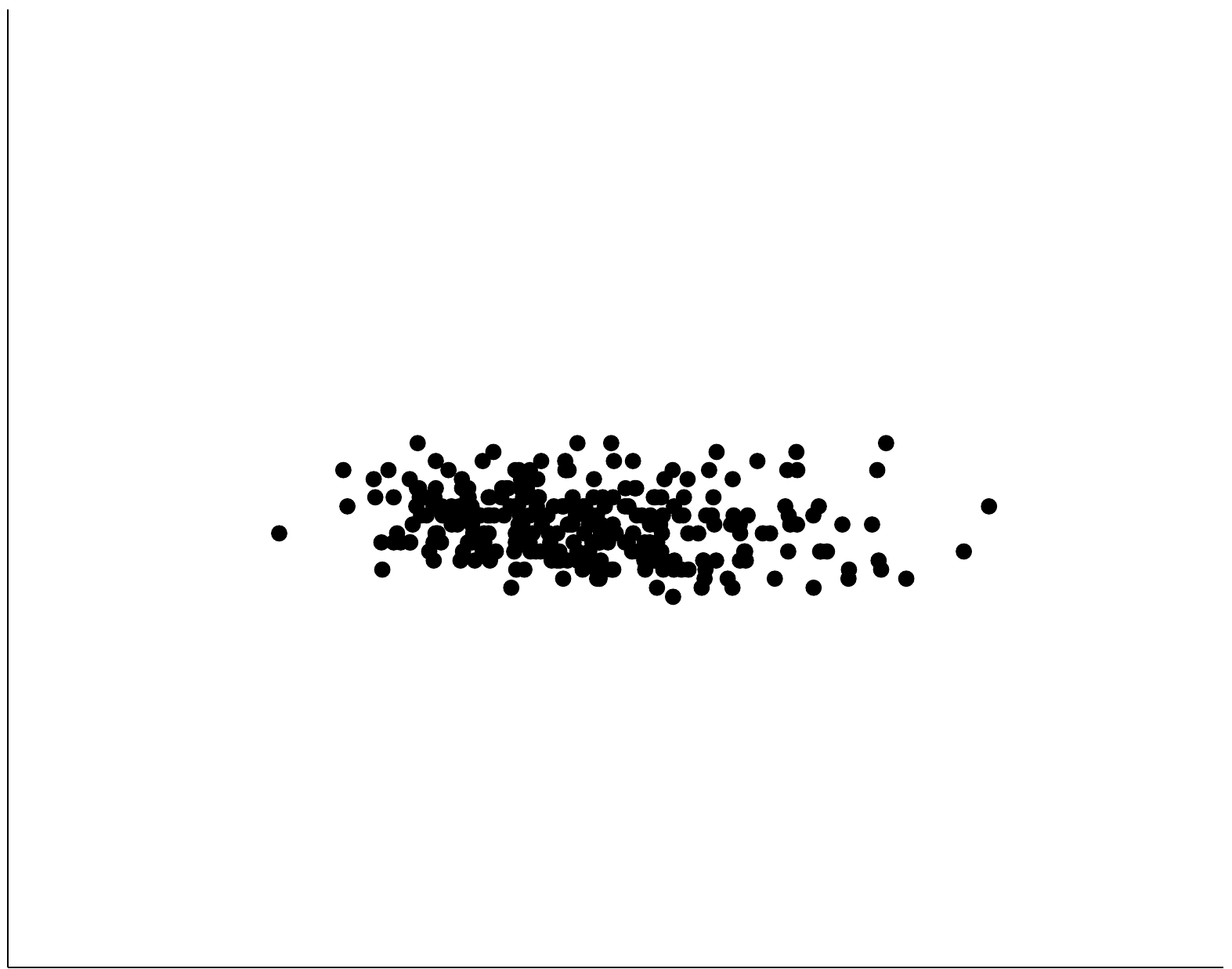}
\end{subfigure}&
\begin{subfigure}{0.1\textwidth}
    \includegraphics[height=10.5mm]{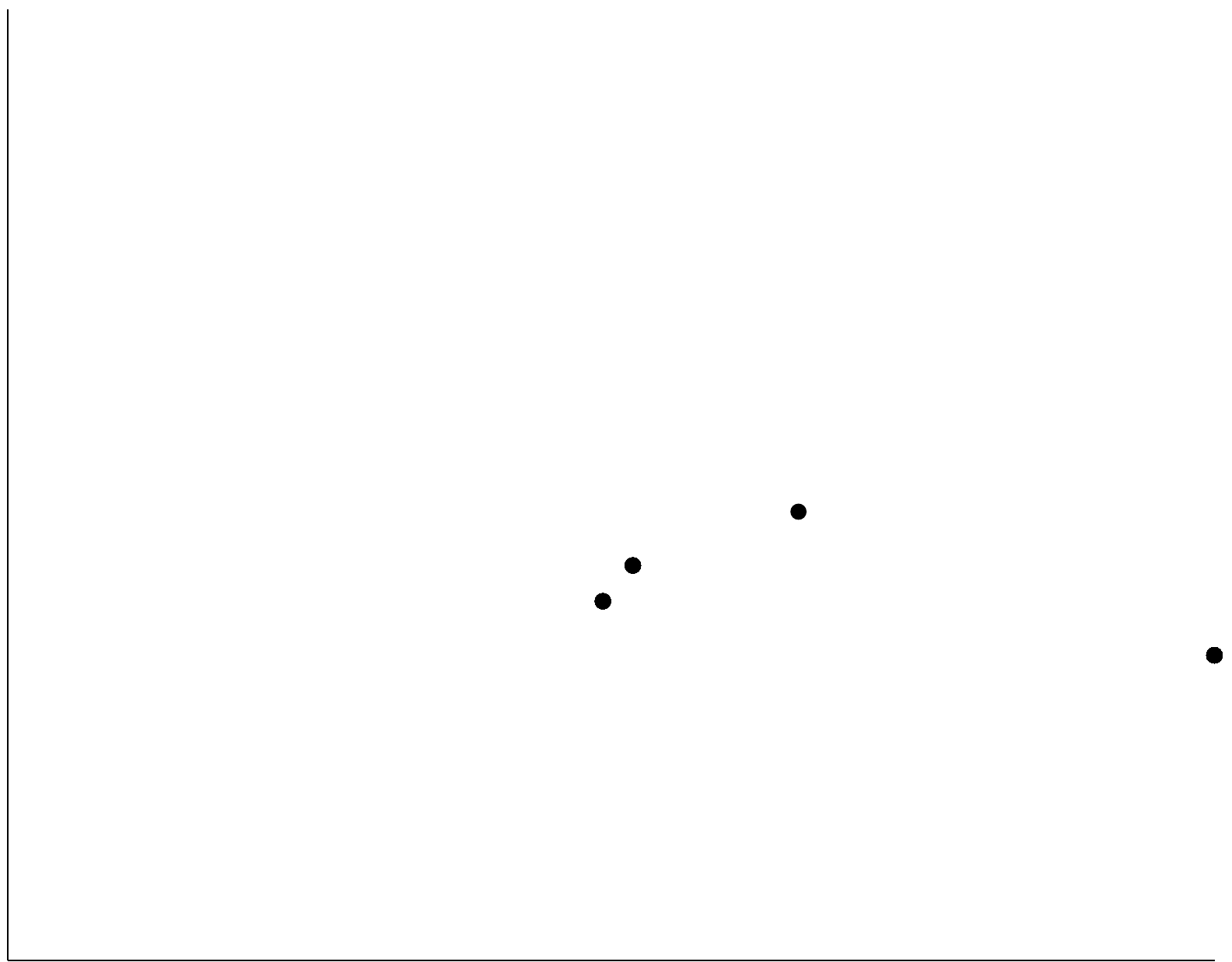}
\end{subfigure}\\ \\
New-Thyroid &
\begin{subfigure}{0.1\textwidth}
    \includegraphics[height=10.5mm]{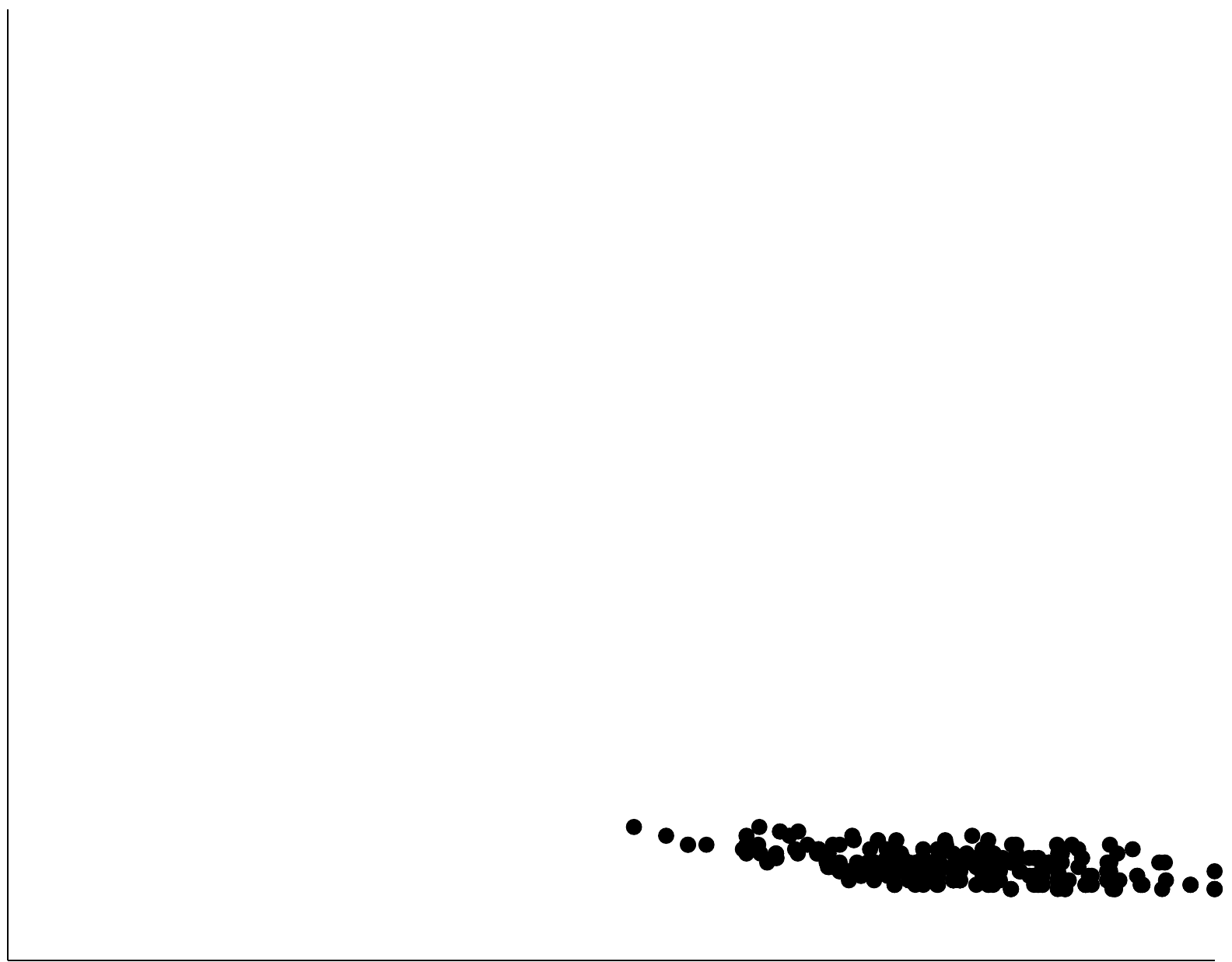}
\end{subfigure}&
\begin{subfigure}{0.1\textwidth}
    \includegraphics[height=10.5mm]{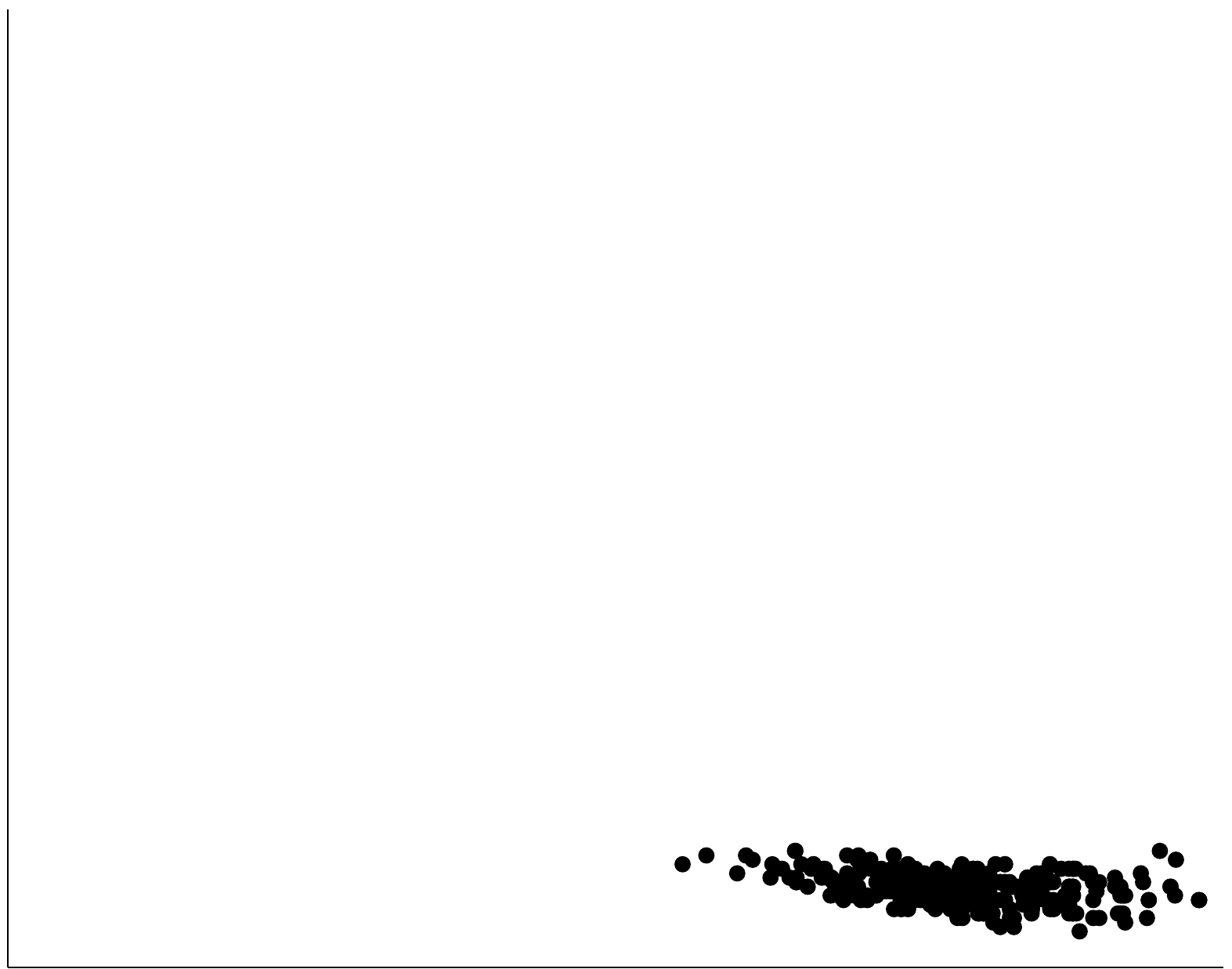}
\end{subfigure}&
\begin{subfigure}{0.1\textwidth}
    \includegraphics[height=10.5mm]{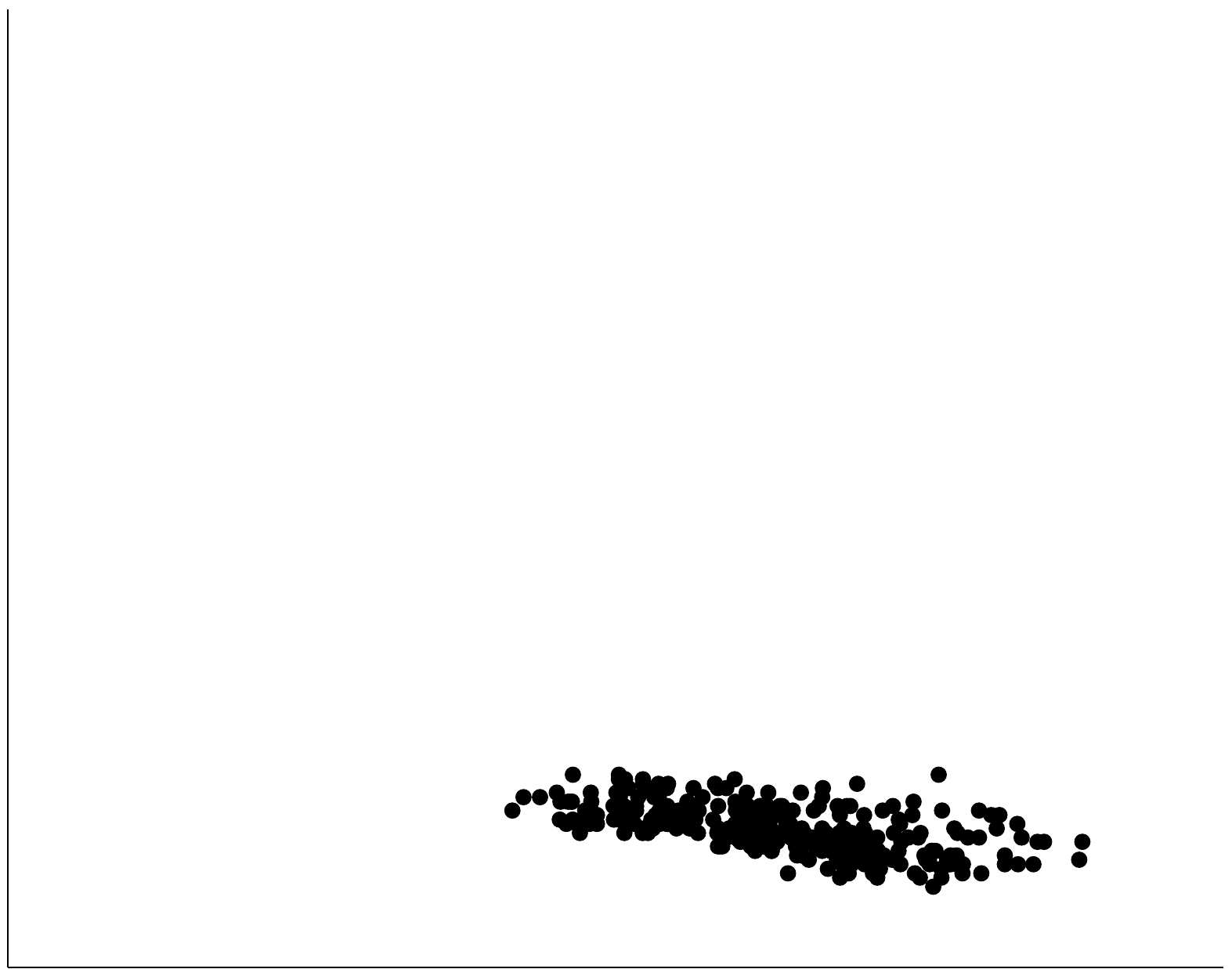}
\end{subfigure}&
\begin{subfigure}{0.1\textwidth}
    \includegraphics[height=10.5mm]{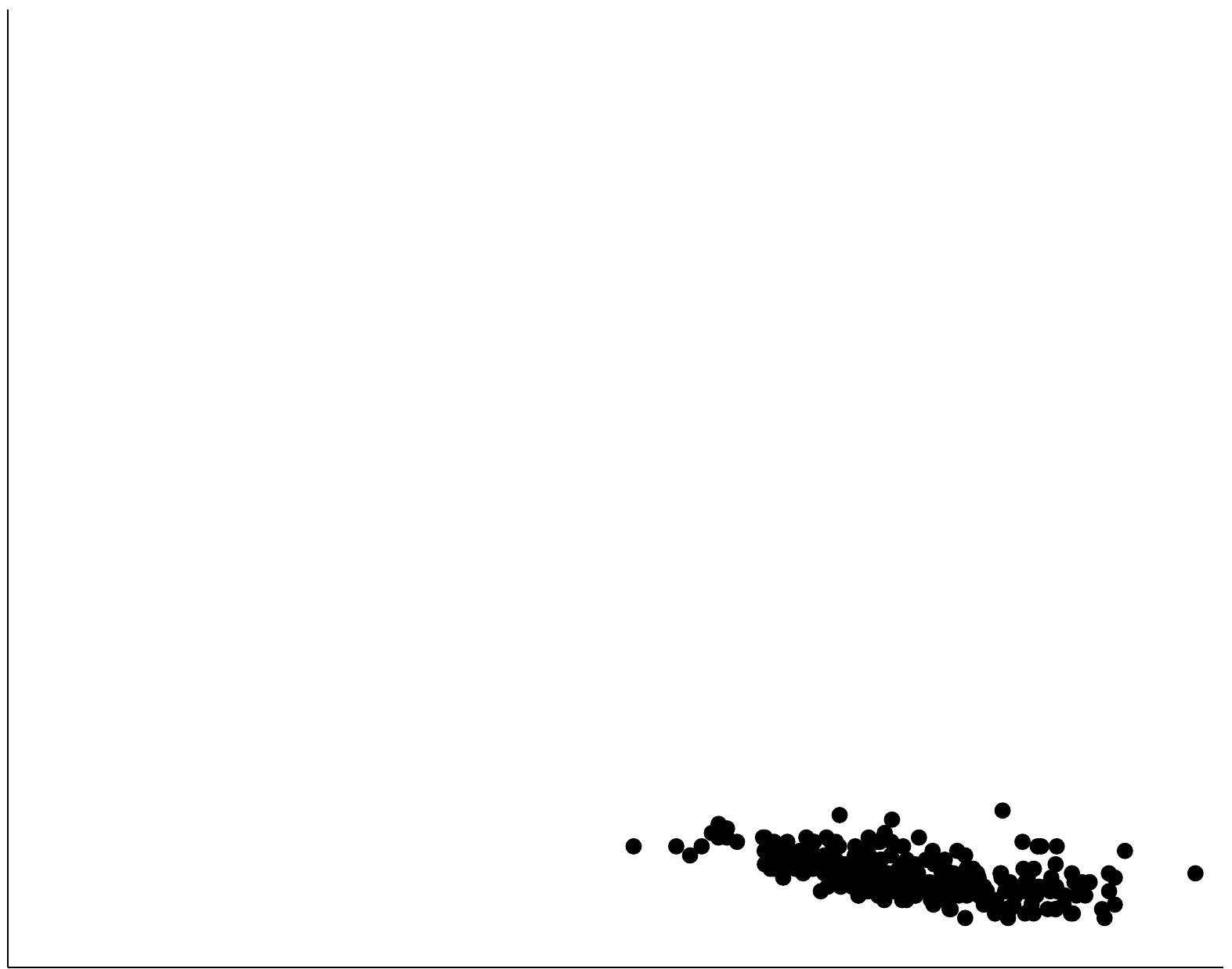}
\end{subfigure}&
\begin{subfigure}{0.1\textwidth}
    \includegraphics[height=10.5mm]{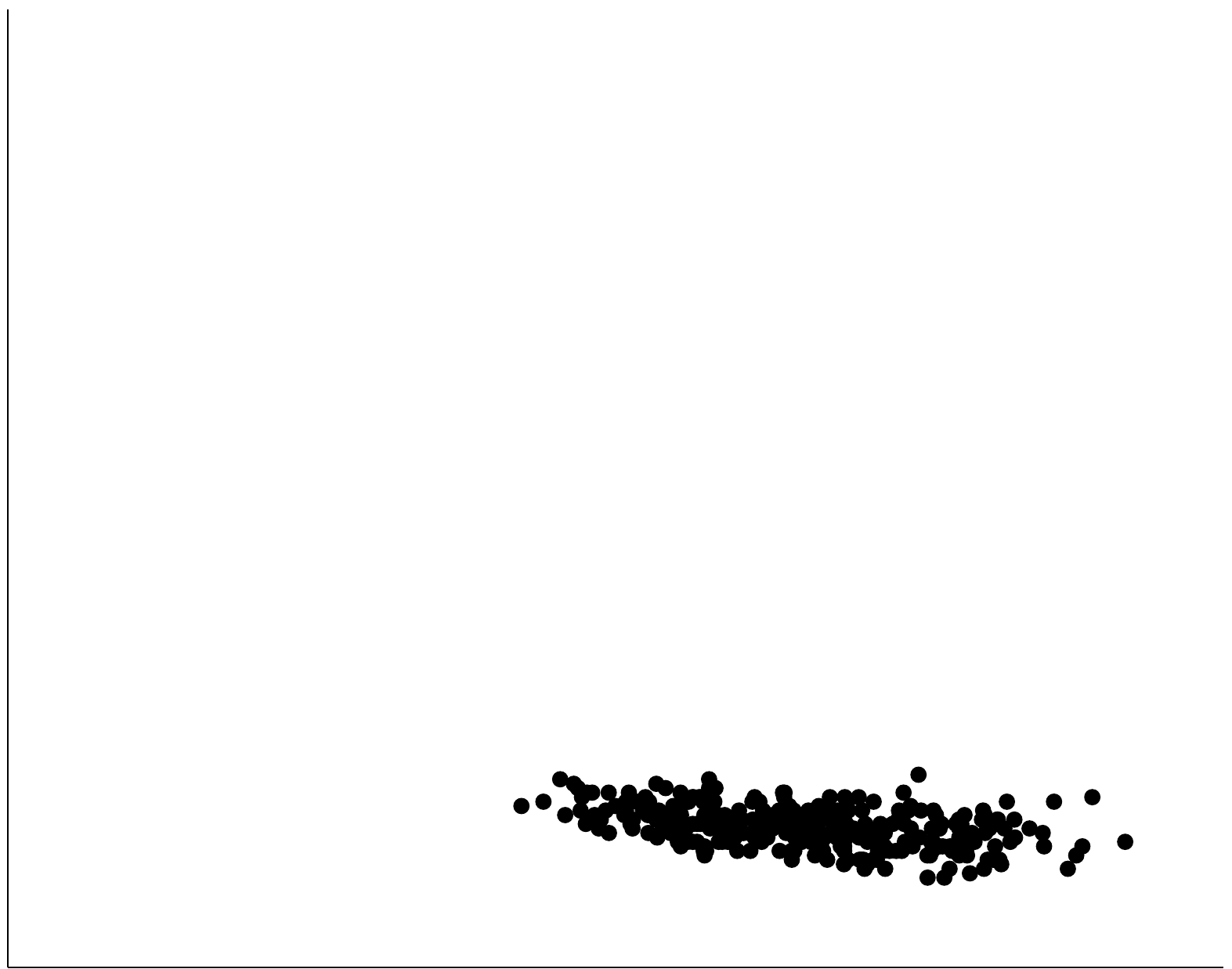}
\end{subfigure}&
\begin{subfigure}{0.1\textwidth}
    \includegraphics[height=10.5mm]{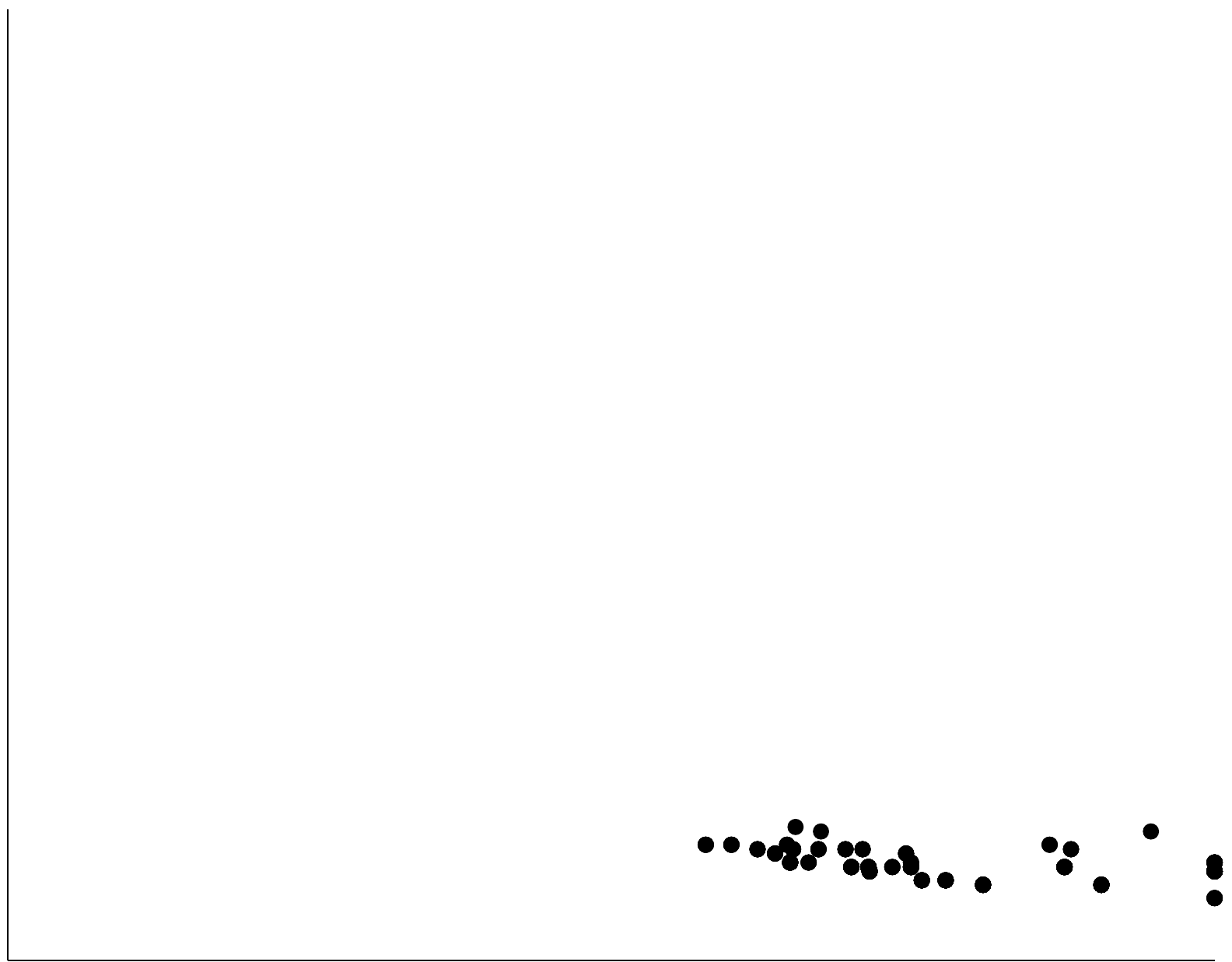}
\end{subfigure}&
\begin{subfigure}{0.1\textwidth}
    \includegraphics[height=10.5mm]{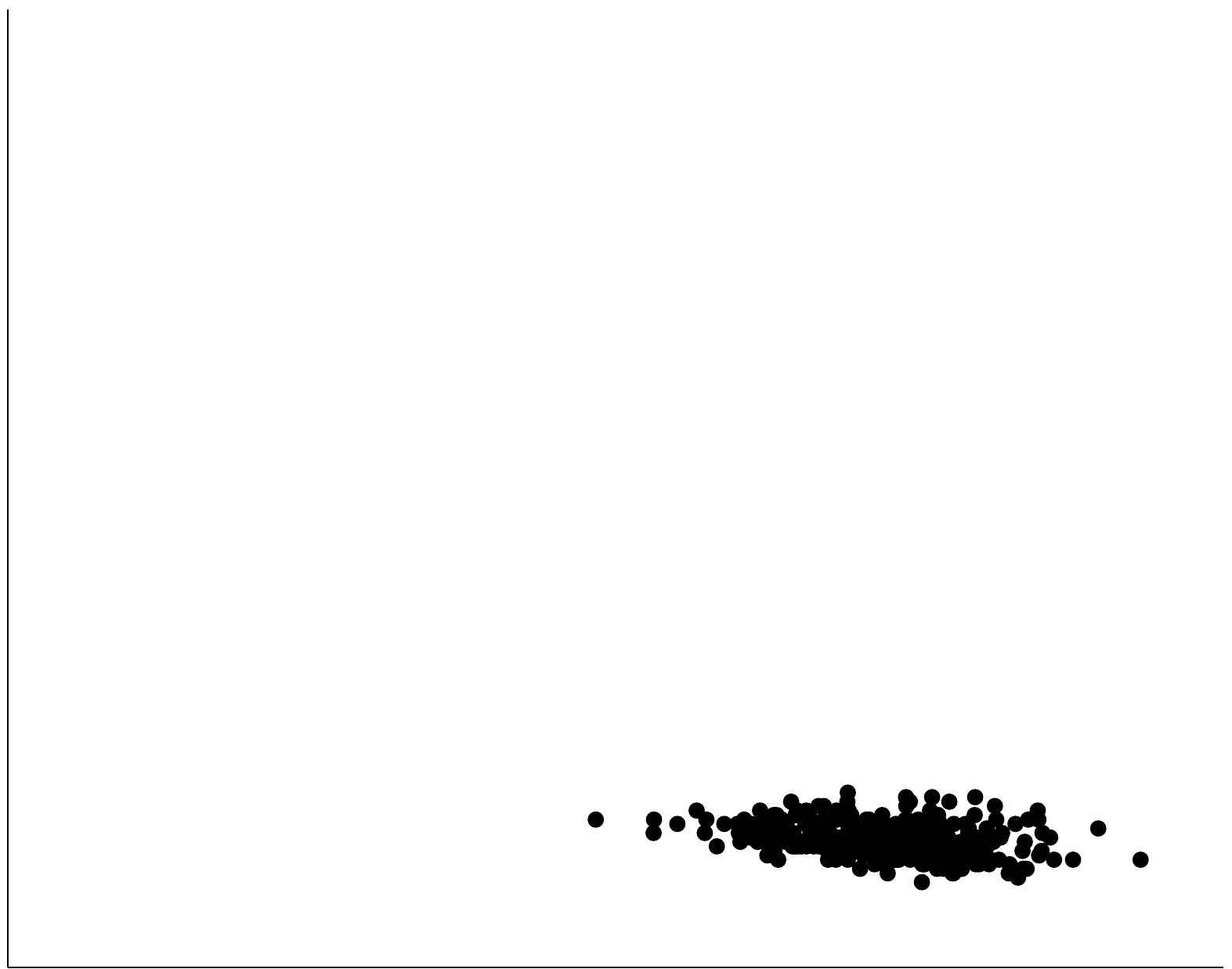}
\end{subfigure}&
\begin{subfigure}{0.1\textwidth}
    \includegraphics[height=10.5mm]{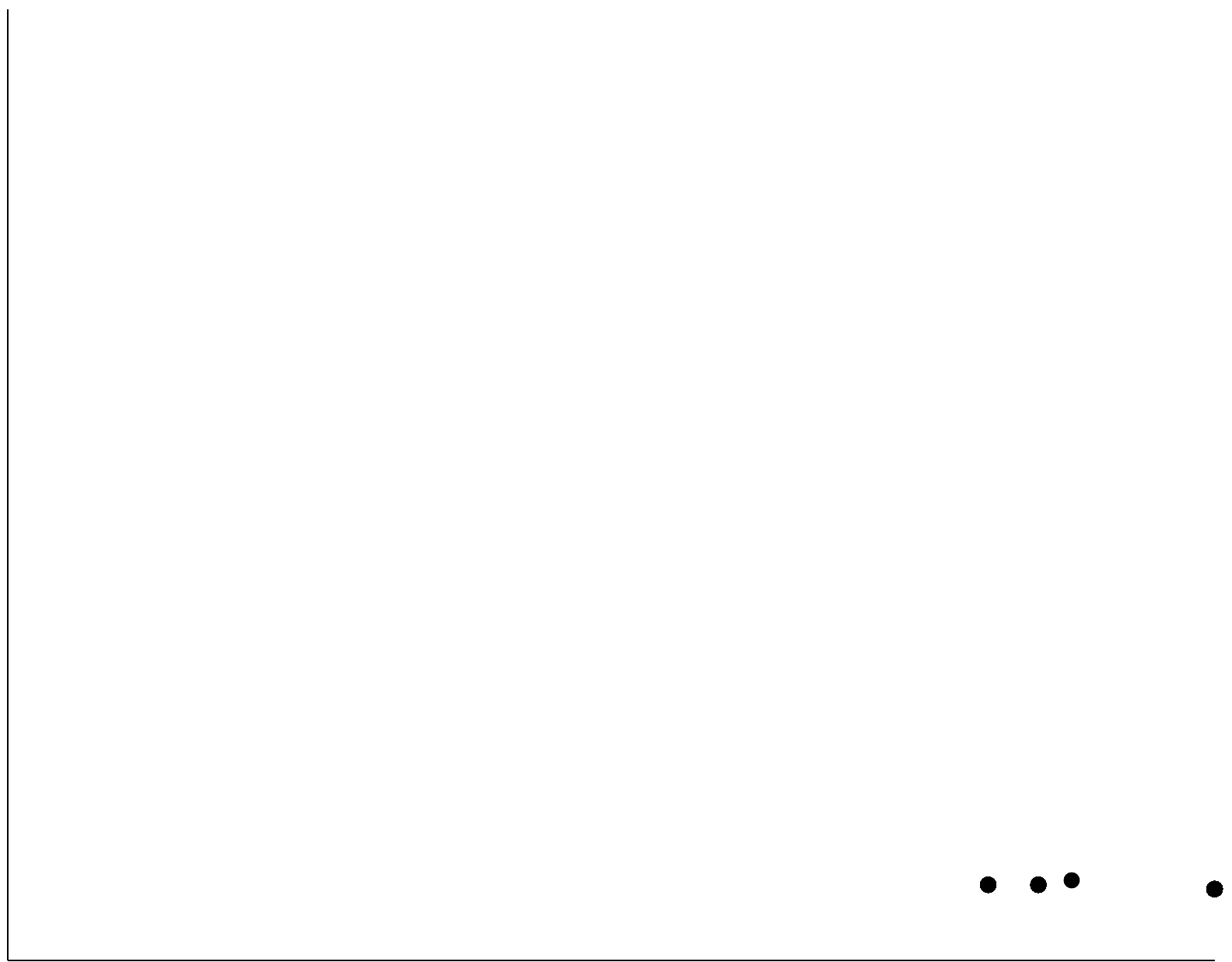}
\end{subfigure}\\ \\
Column &
\begin{subfigure}{0.1\textwidth}
    \includegraphics[height=10.5mm]{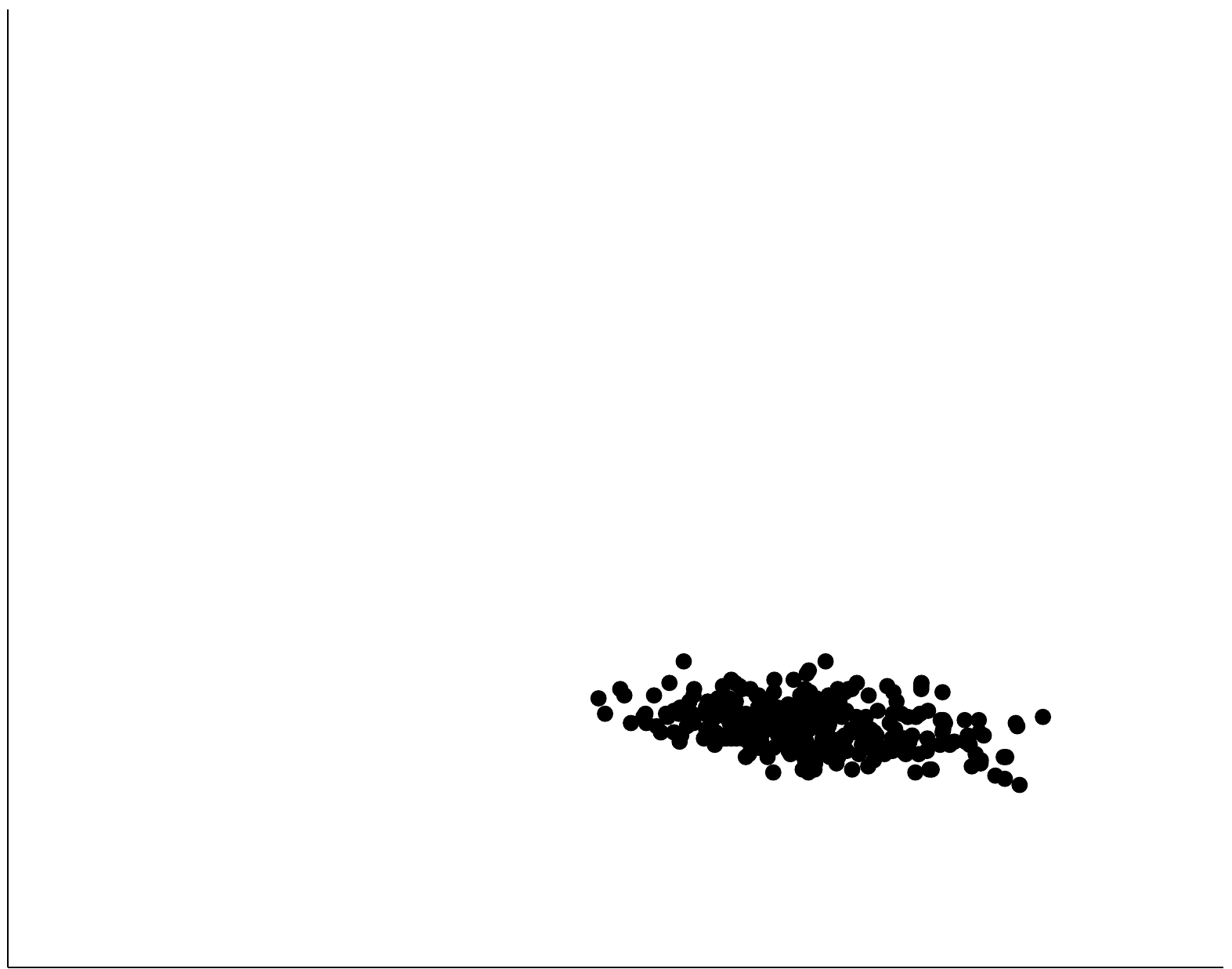}
\end{subfigure}&
\begin{subfigure}{0.1\textwidth}
    \includegraphics[height=10.5mm]{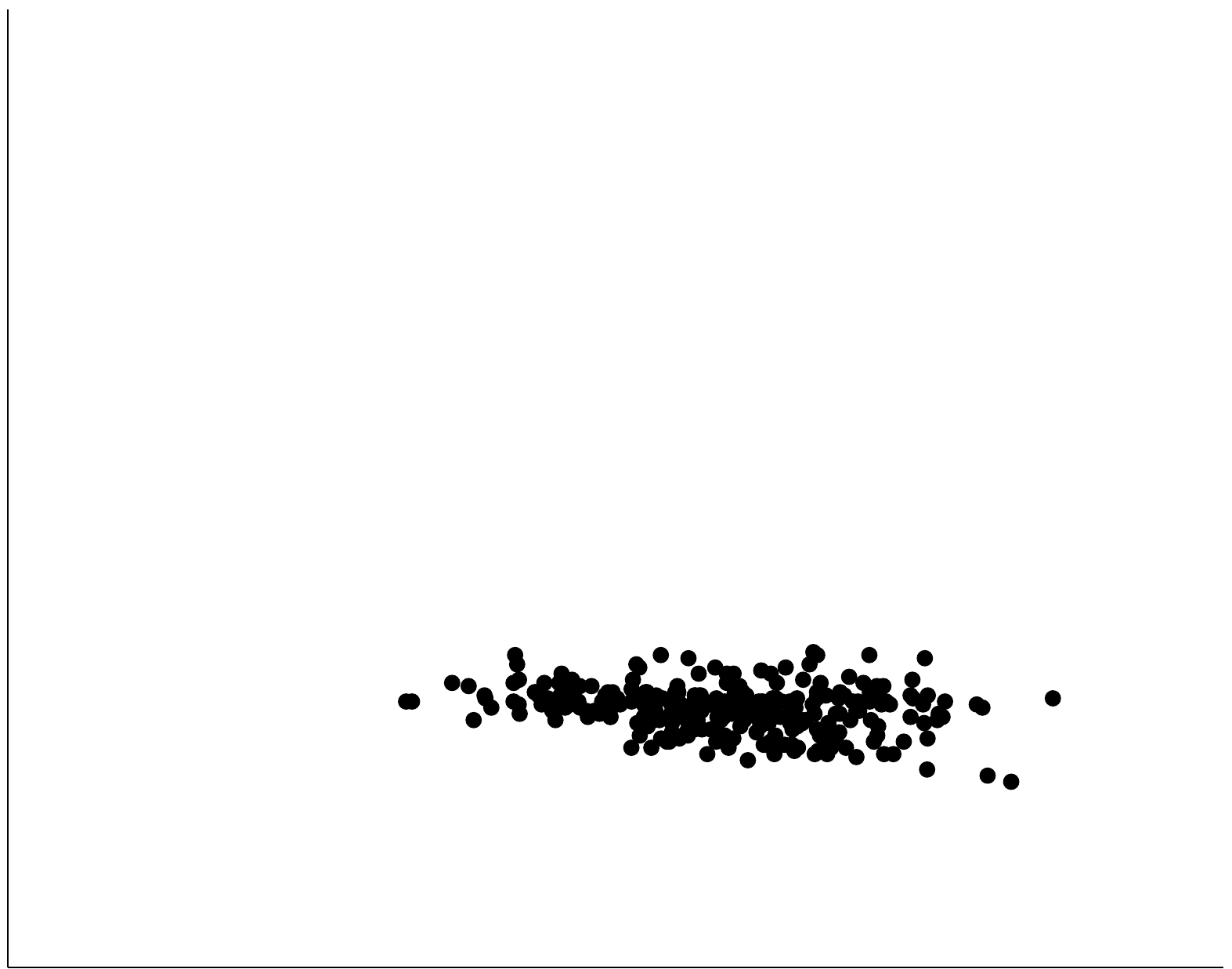}
\end{subfigure}&
\begin{subfigure}{0.1\textwidth}
    \includegraphics[height=10.5mm]{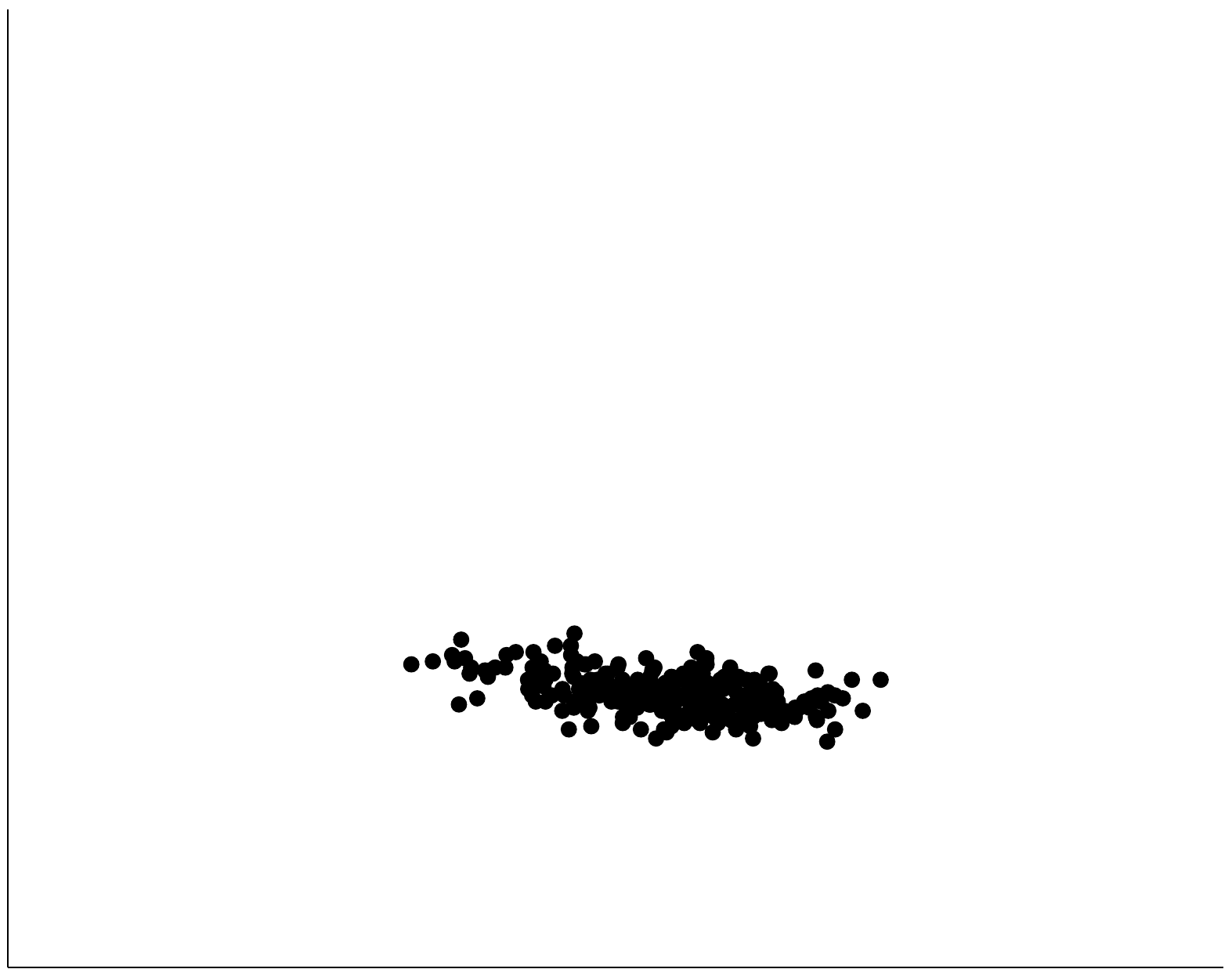}
\end{subfigure}&
\begin{subfigure}{0.1\textwidth}
    \includegraphics[height=10.5mm]{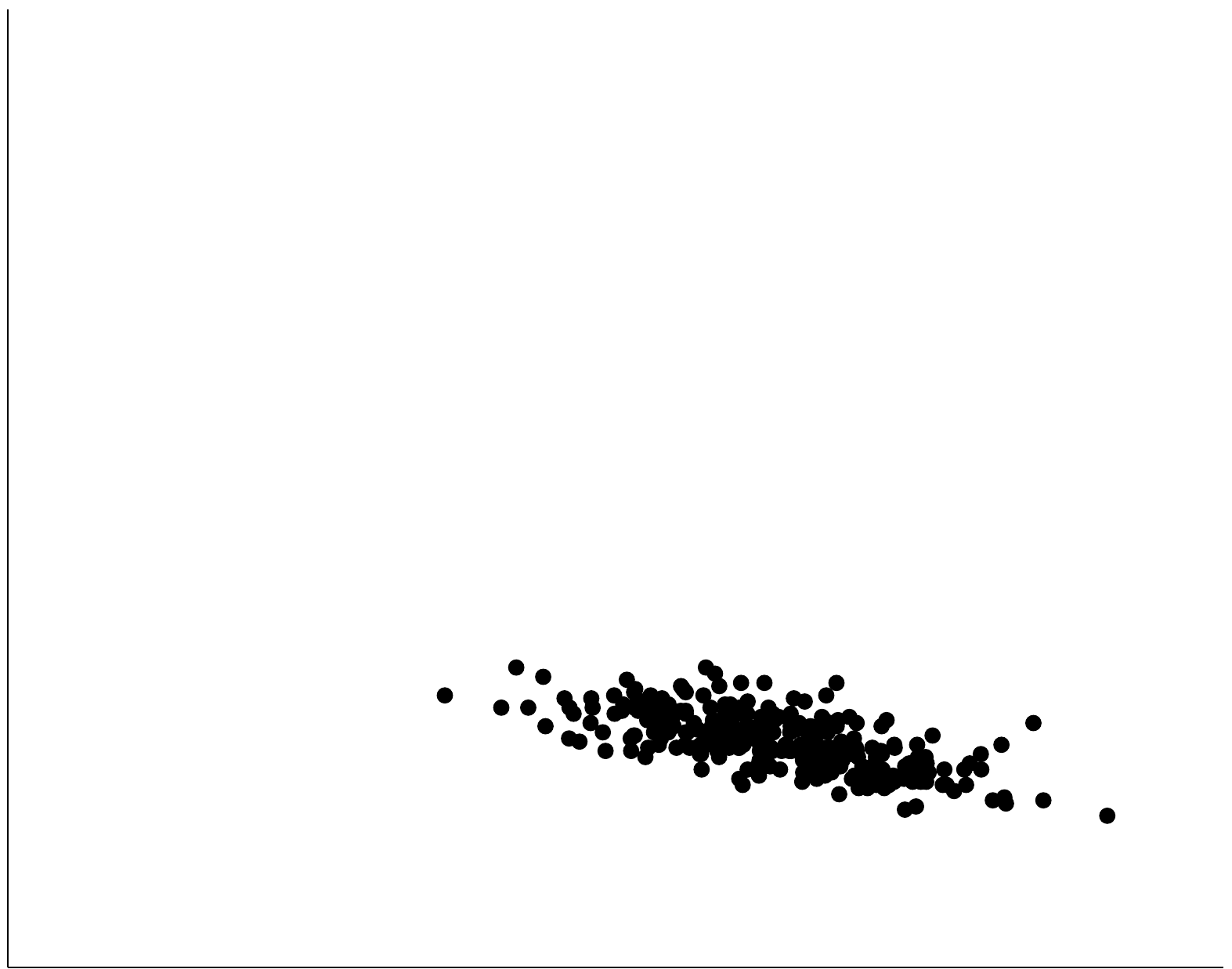}
\end{subfigure}&
\begin{subfigure}{0.1\textwidth}
    \includegraphics[height=10.5mm]{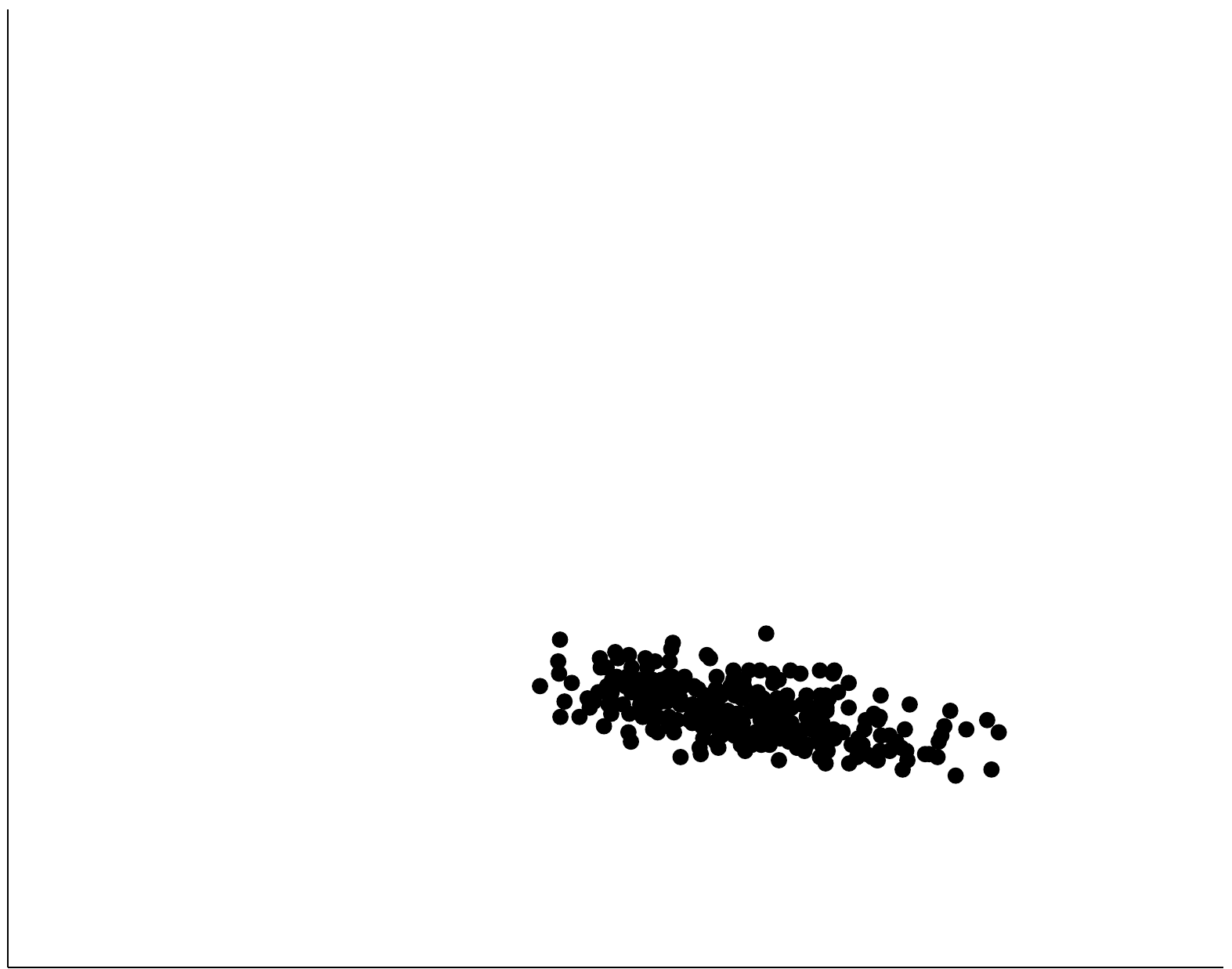}
\end{subfigure}&
\begin{subfigure}{0.1\textwidth}
    \includegraphics[height=10.5mm]{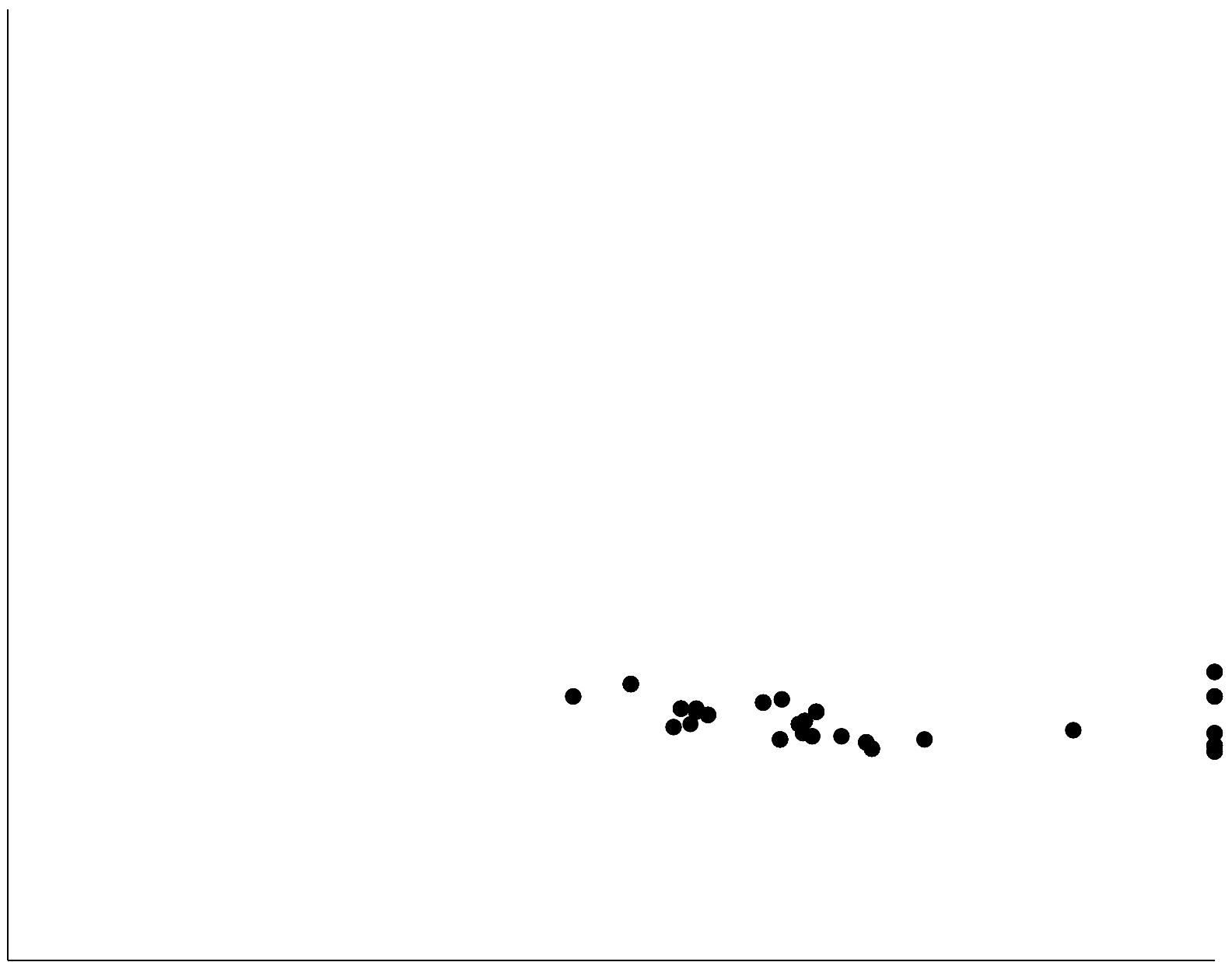}
\end{subfigure}&
\begin{subfigure}{0.1\textwidth}
    \includegraphics[height=10.5mm]{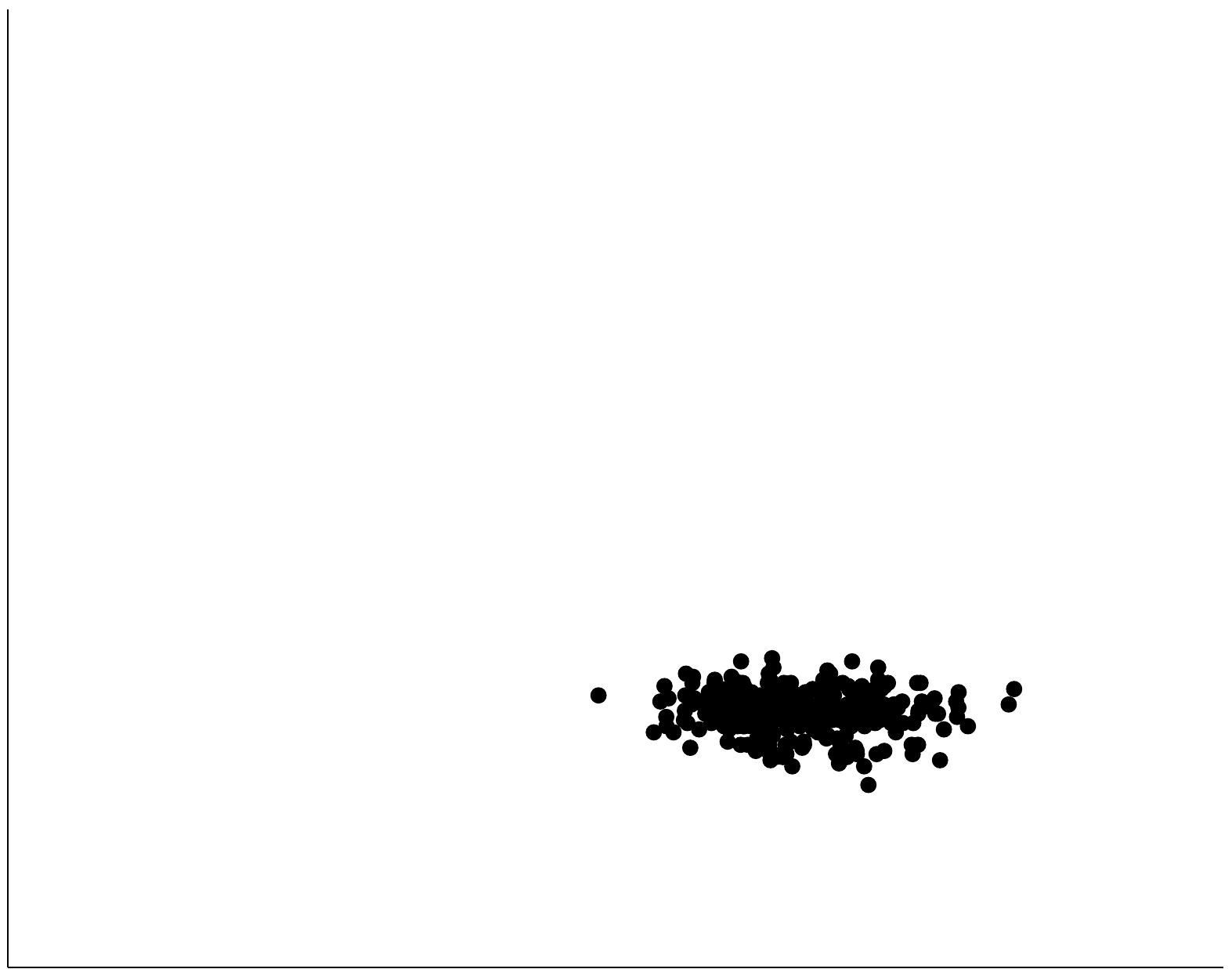}
\end{subfigure}&
\begin{subfigure}{0.1\textwidth}
    \includegraphics[height=10.5mm]{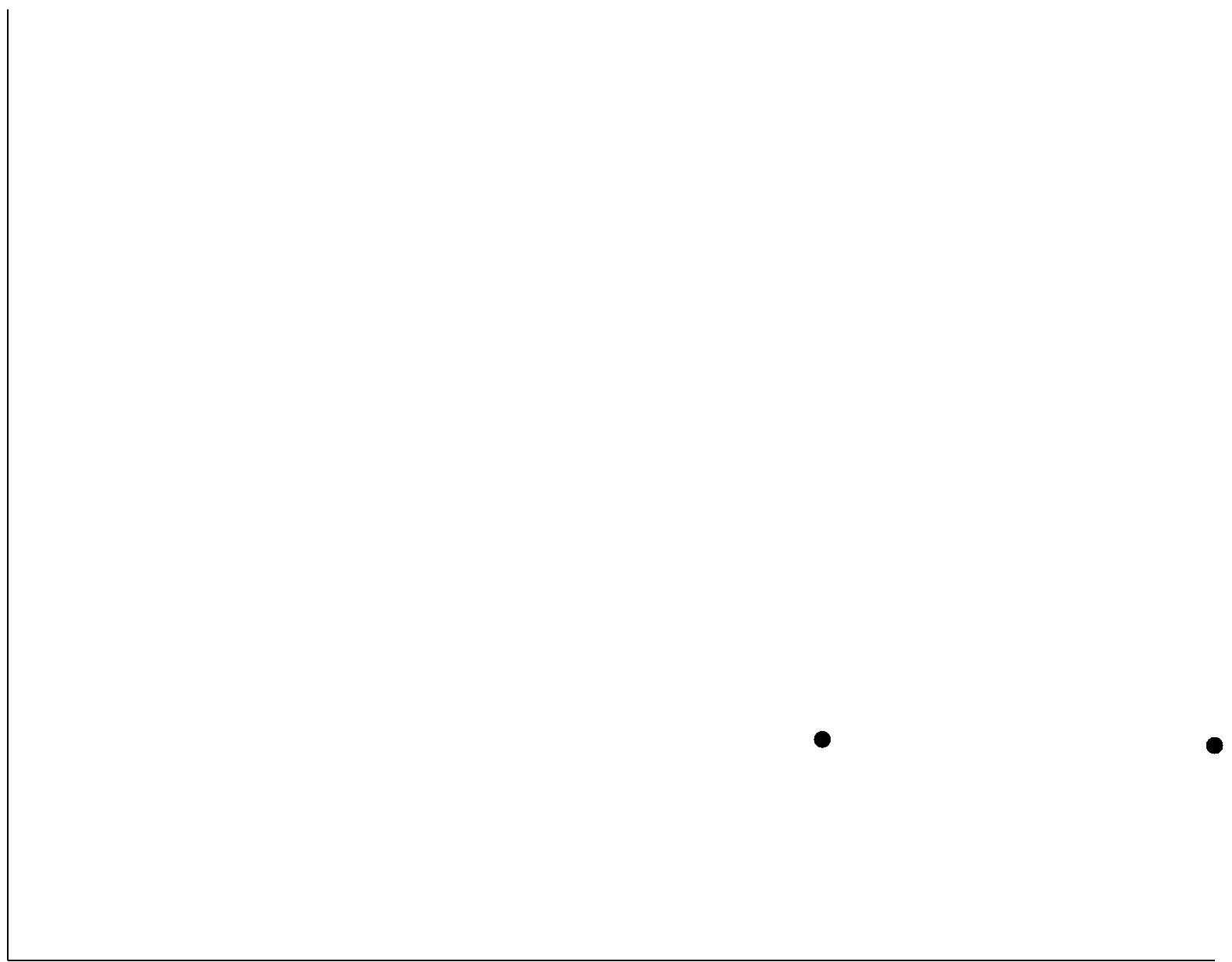}
\end{subfigure}\\ \\
Seeds &
\begin{subfigure}{0.1\textwidth}
    \includegraphics[height=10.5mm]{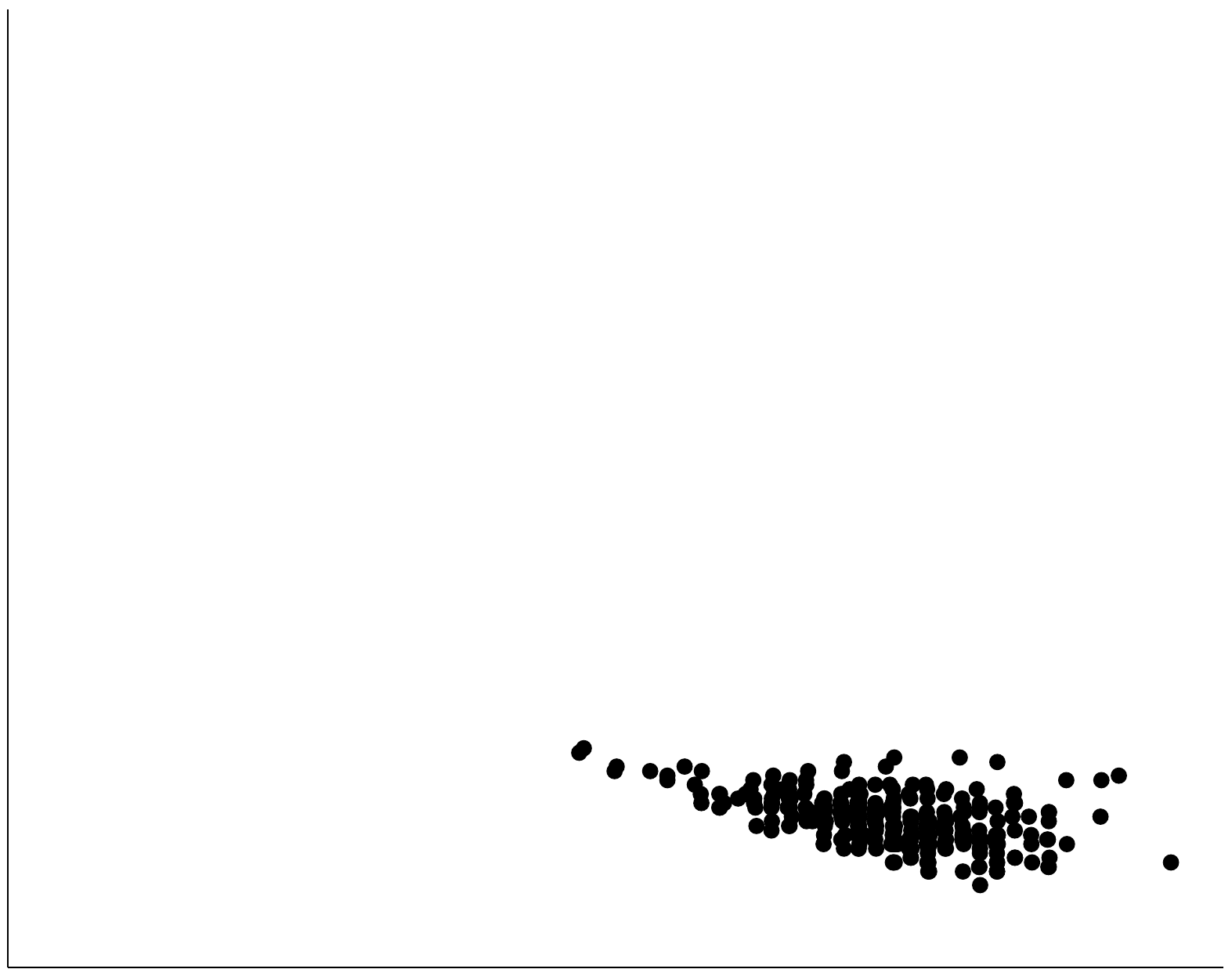}
\end{subfigure}&
\begin{subfigure}{0.1\textwidth}
    \includegraphics[height=10.5mm]{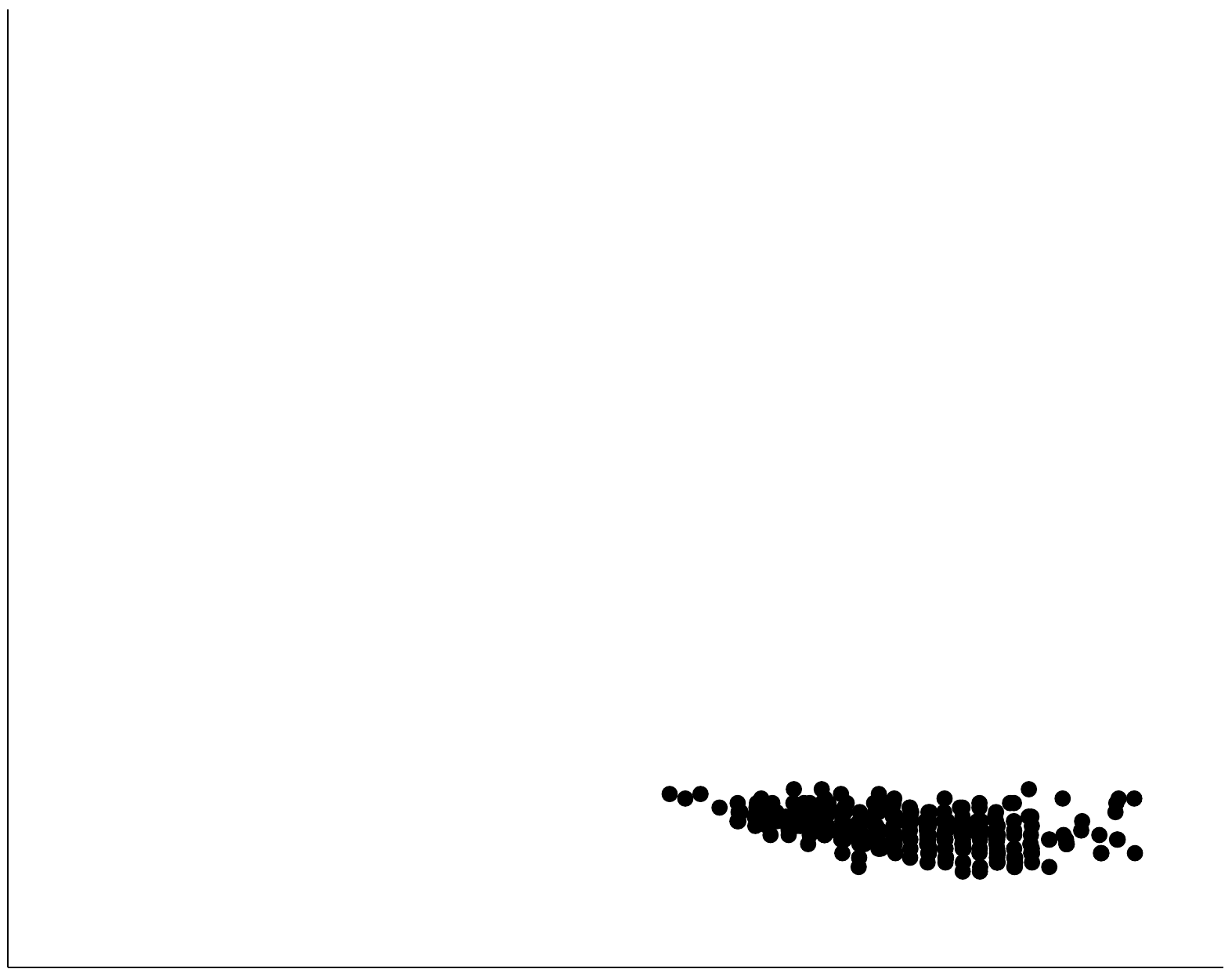}
\end{subfigure}&
\begin{subfigure}{0.1\textwidth}
    \includegraphics[height=10.5mm]{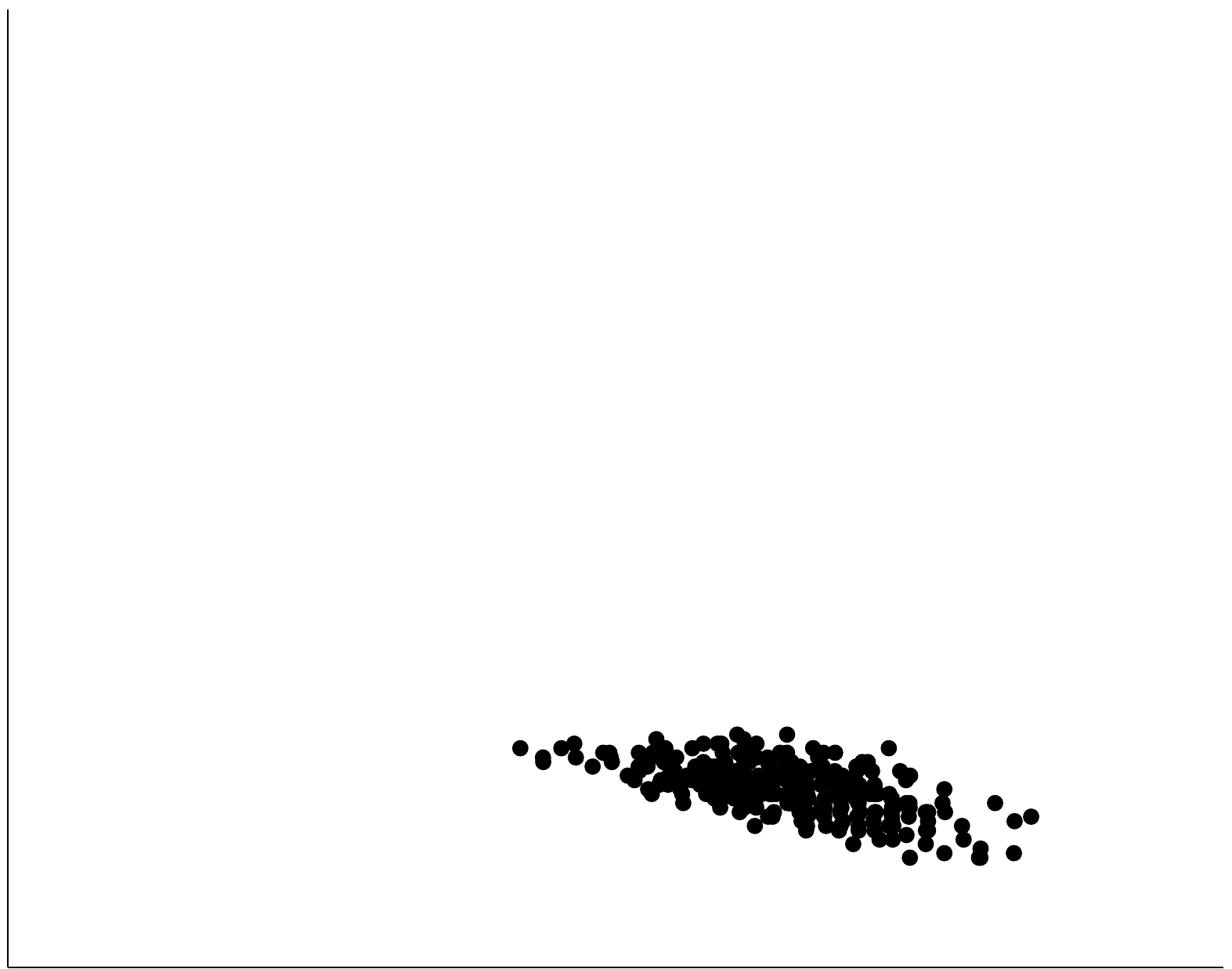}
\end{subfigure}&
\begin{subfigure}{0.1\textwidth}
    \includegraphics[height=10.5mm]{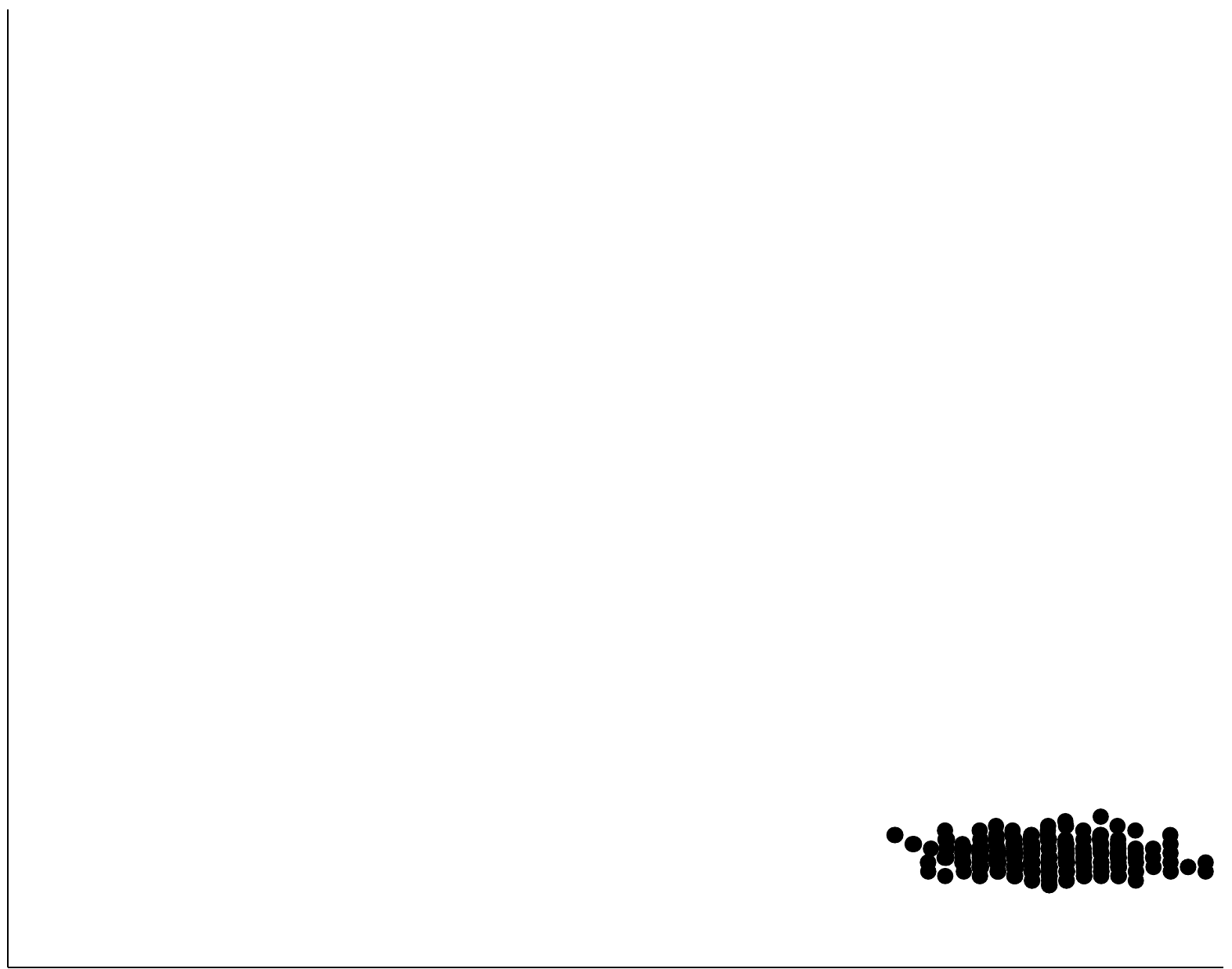}
\end{subfigure}&
\begin{subfigure}{0.1\textwidth}
    \includegraphics[height=10.5mm]{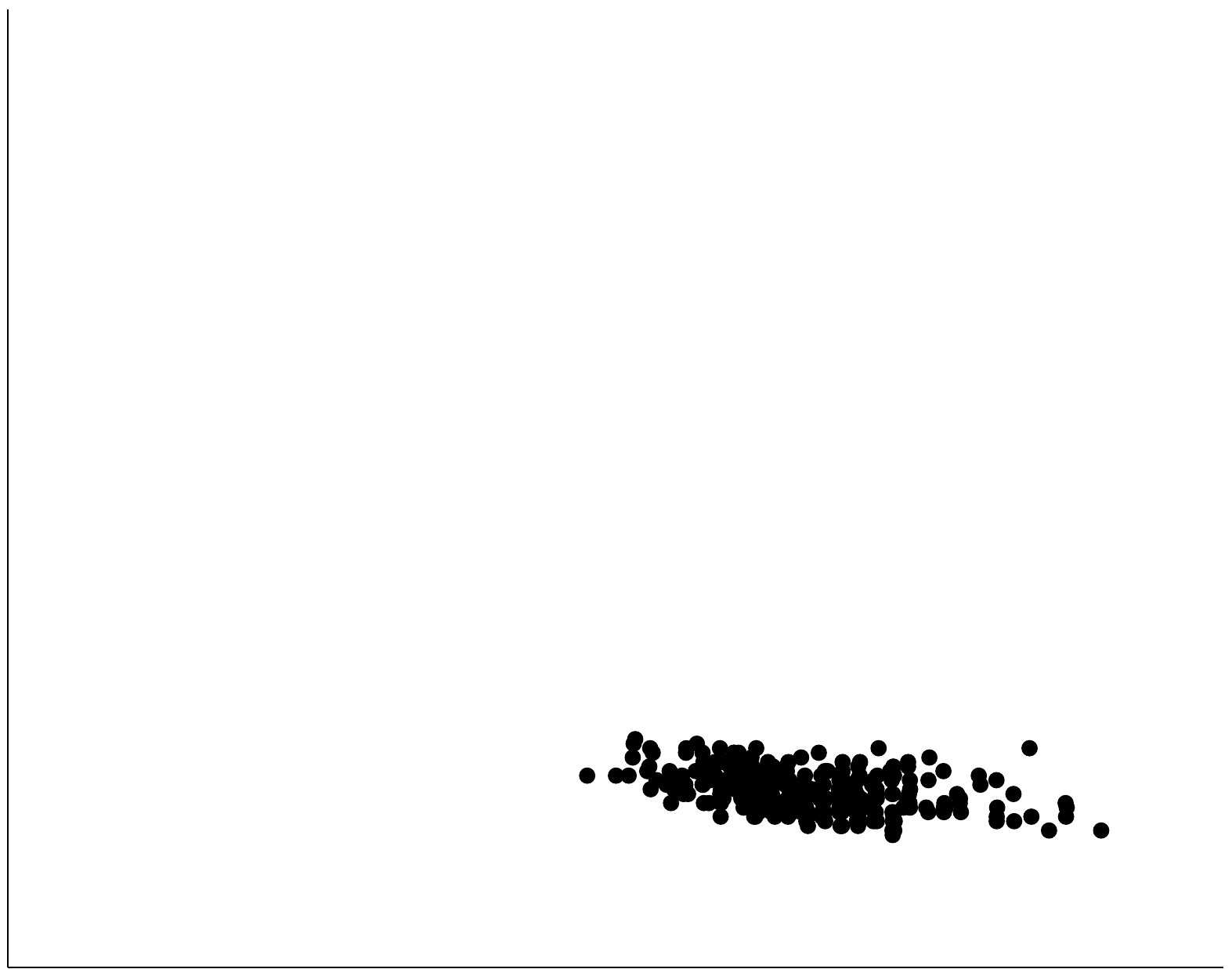}
\end{subfigure}&
\begin{subfigure}{0.1\textwidth}
    \includegraphics[height=10.5mm]{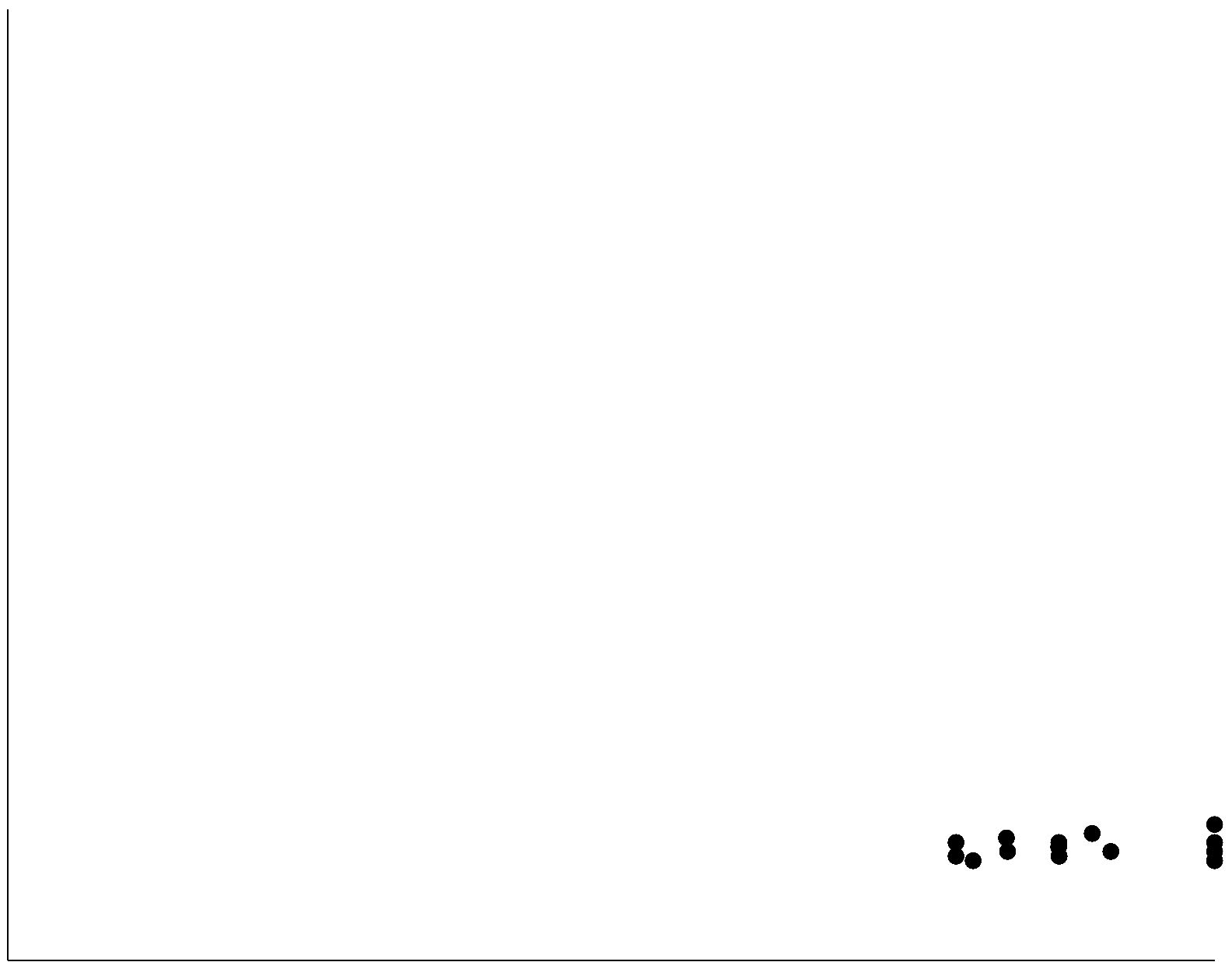}
\end{subfigure}&
\begin{subfigure}{0.1\textwidth}
    \includegraphics[height=10.5mm]{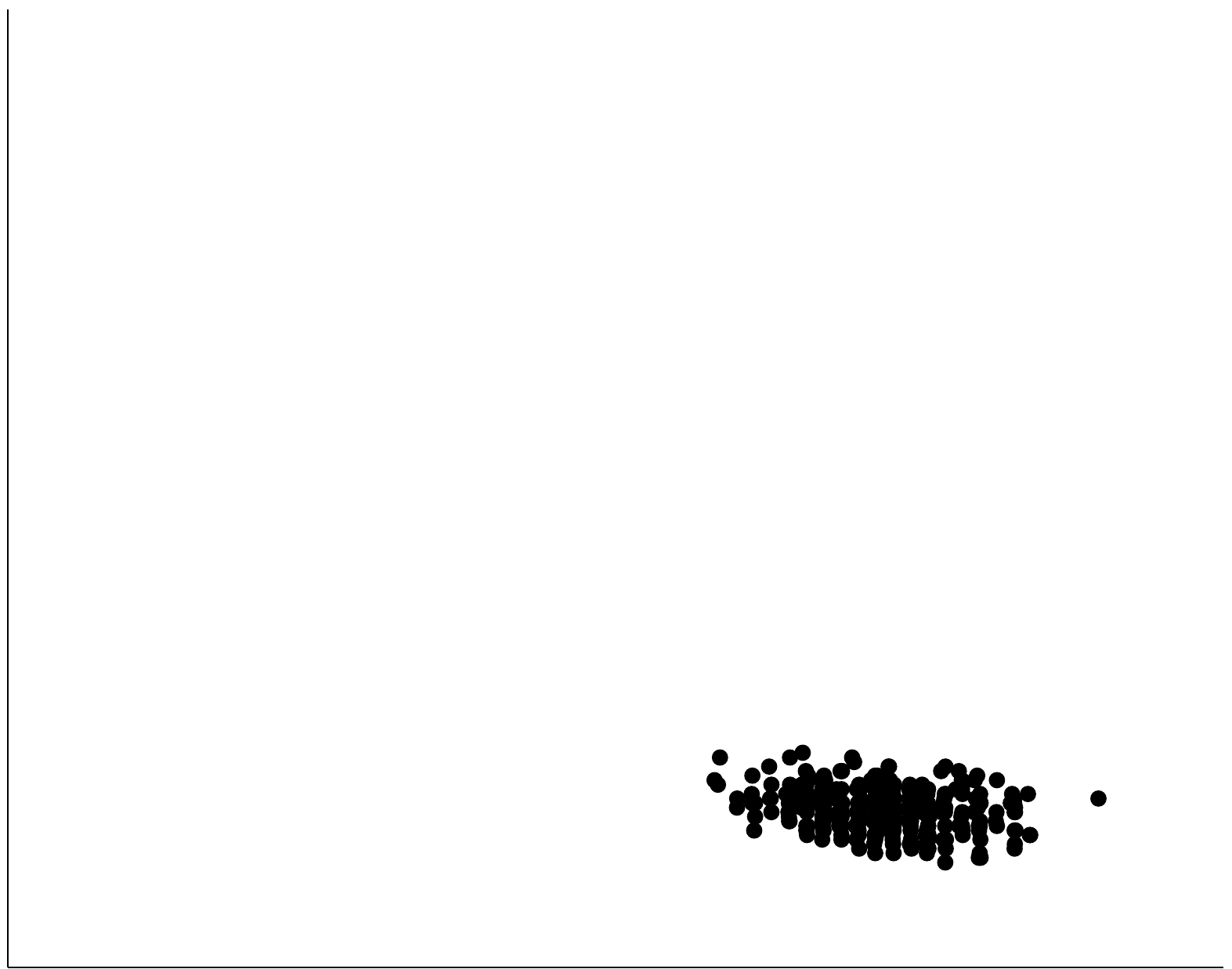}
\end{subfigure}&
\begin{subfigure}{0.1\textwidth}
    \includegraphics[height=10.5mm]{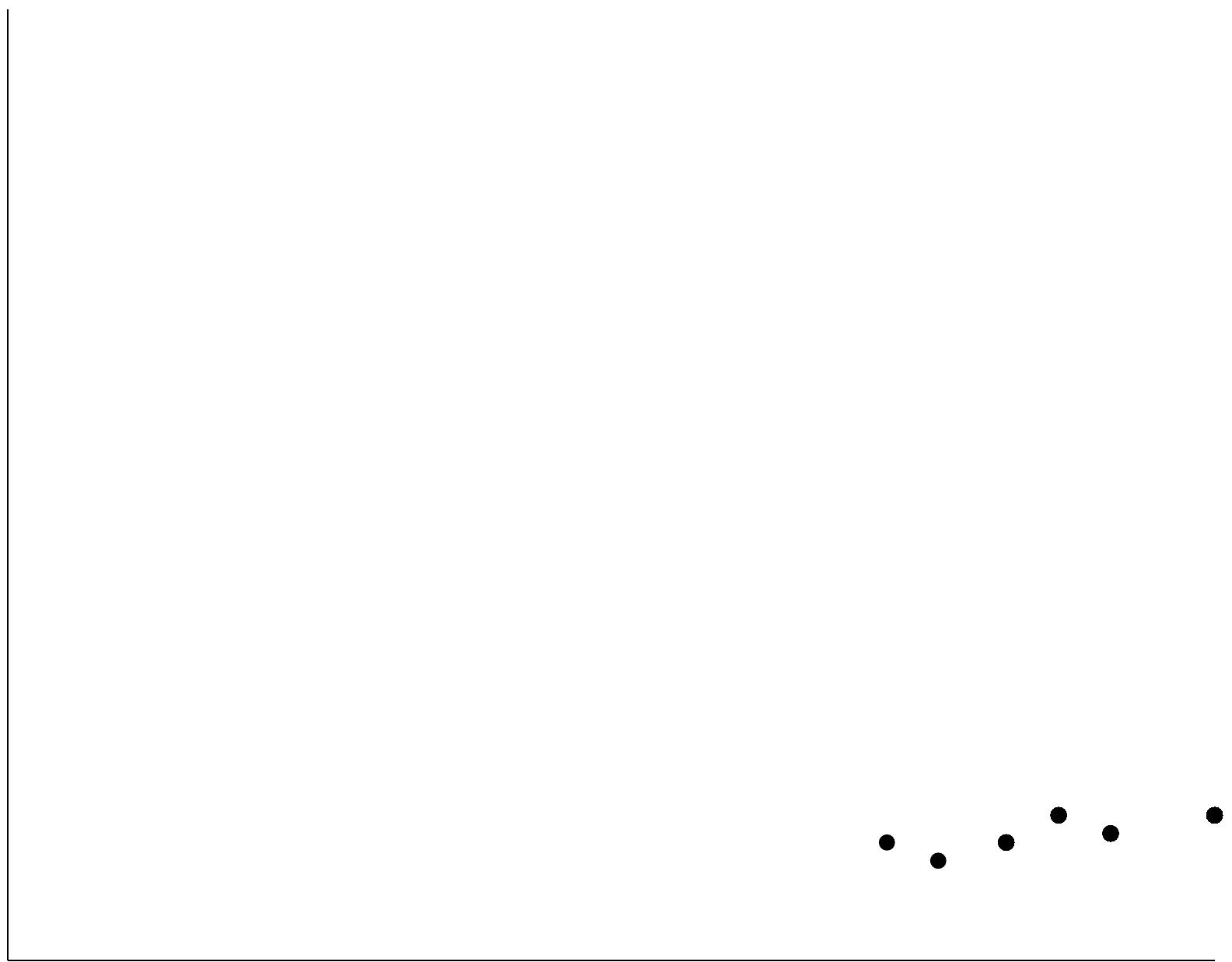}
\end{subfigure}\\
\end{tabular}
}
\label{kappaerror10}
\end{sidewaystable}

\begin{sidewaystable}[htb]
\caption{Kappa-error plots for 30\% missingness for different ensemble methods. The  most  desirable  pairs  of classifiers will lie at the bottom left corner (high diversity and low average error).}
\scalebox{0.9}{
\begin{tabular}{p{2cm}cccccccc}
Dataset & BagNoImp & BagMEI & BagGRandI & BagEM & BagMIGRandI & BagMIEMI & MIGRandI & MIEMI \\ \\ \hline
Breast Tissue &
\begin{subfigure}{0.1\textwidth}
    \includegraphics[height=10.5mm]{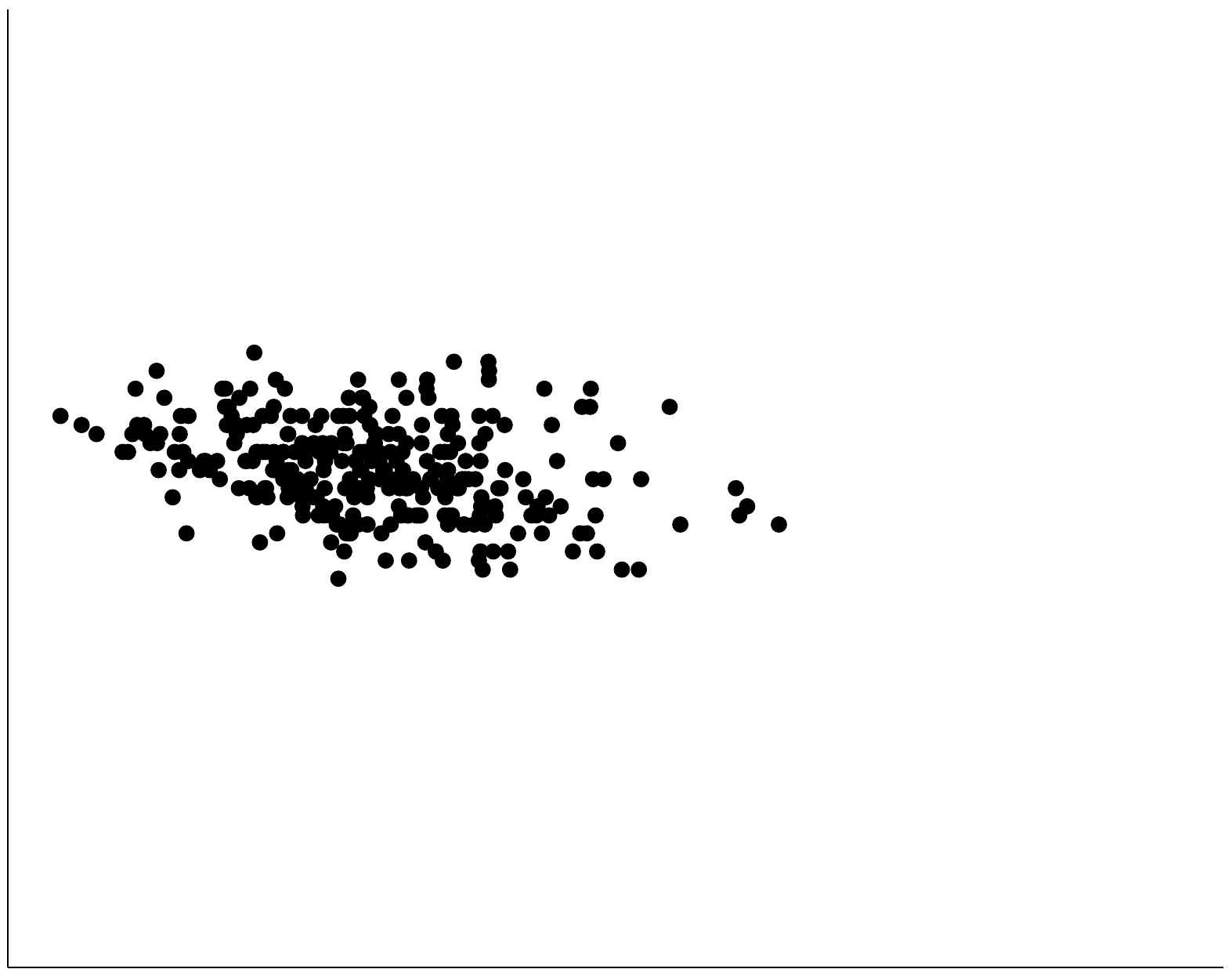}
\end{subfigure}&
\begin{subfigure}{0.1\textwidth}
    \includegraphics[height=10.5mm]{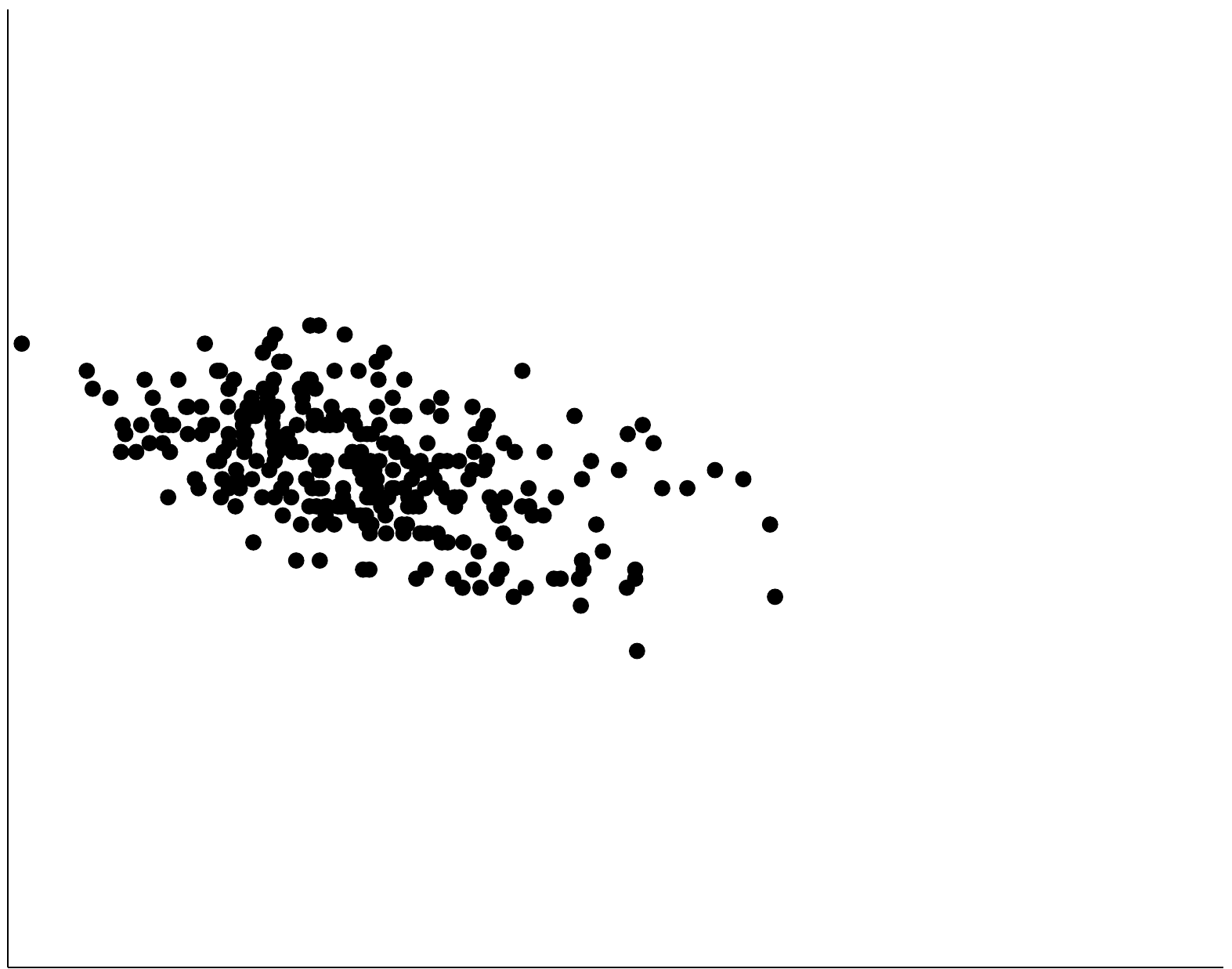}
\end{subfigure}&
\begin{subfigure}{0.1\textwidth}
    \includegraphics[height=10.5mm]{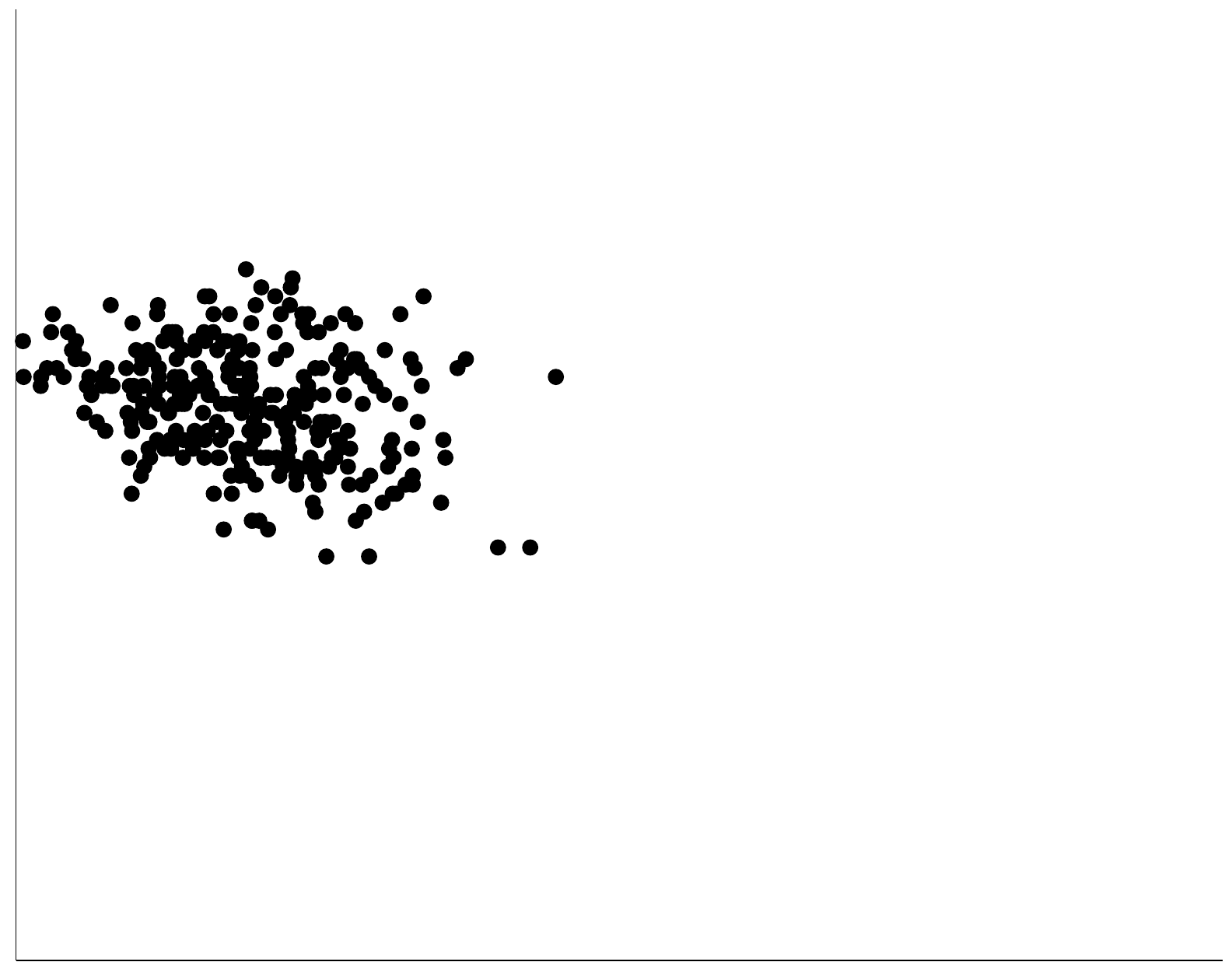}
\end{subfigure}&
\begin{subfigure}{0.1\textwidth}
    \includegraphics[height=10.5mm]{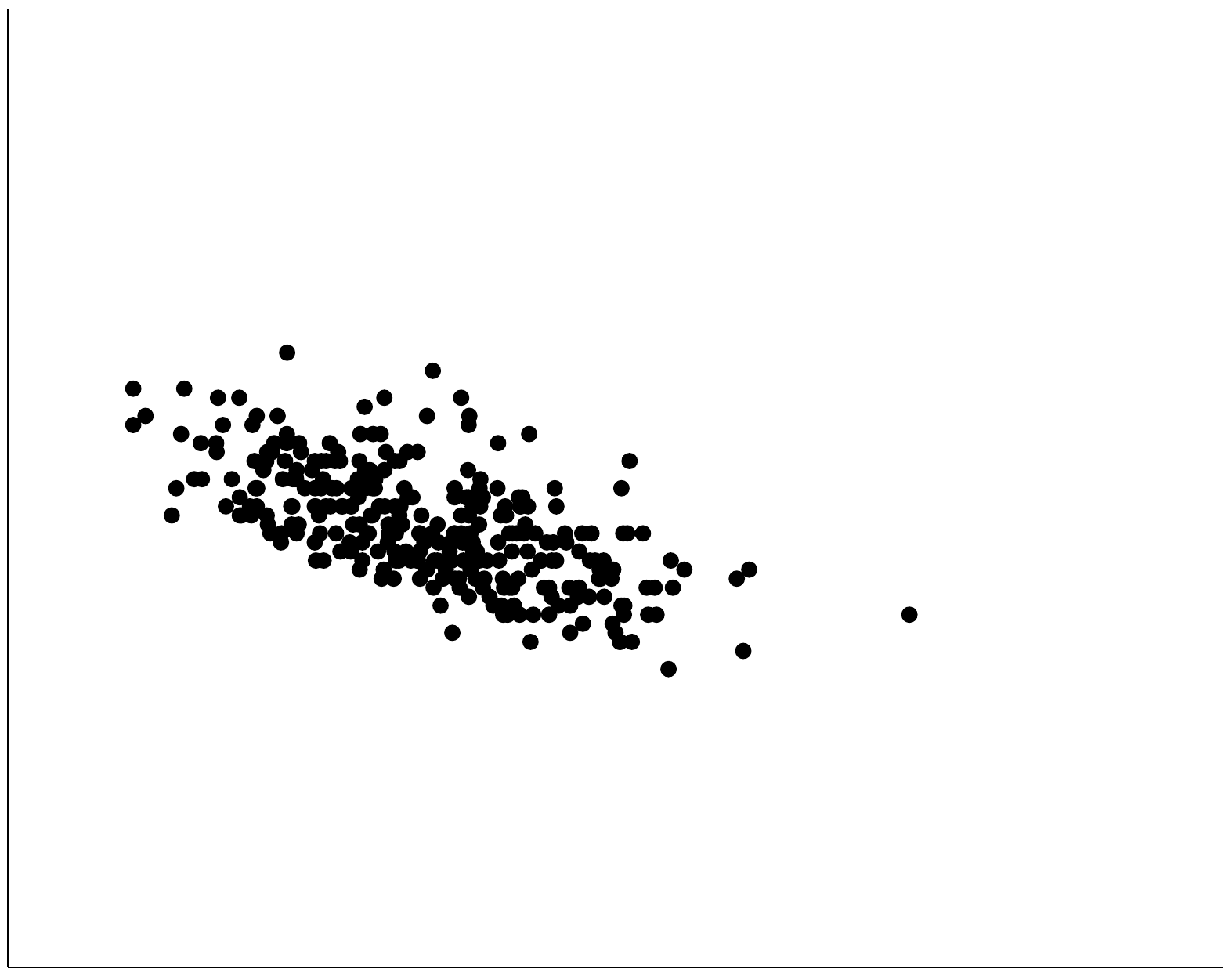}
\end{subfigure}&
\begin{subfigure}{0.1\textwidth}
    \includegraphics[height=10.5mm]{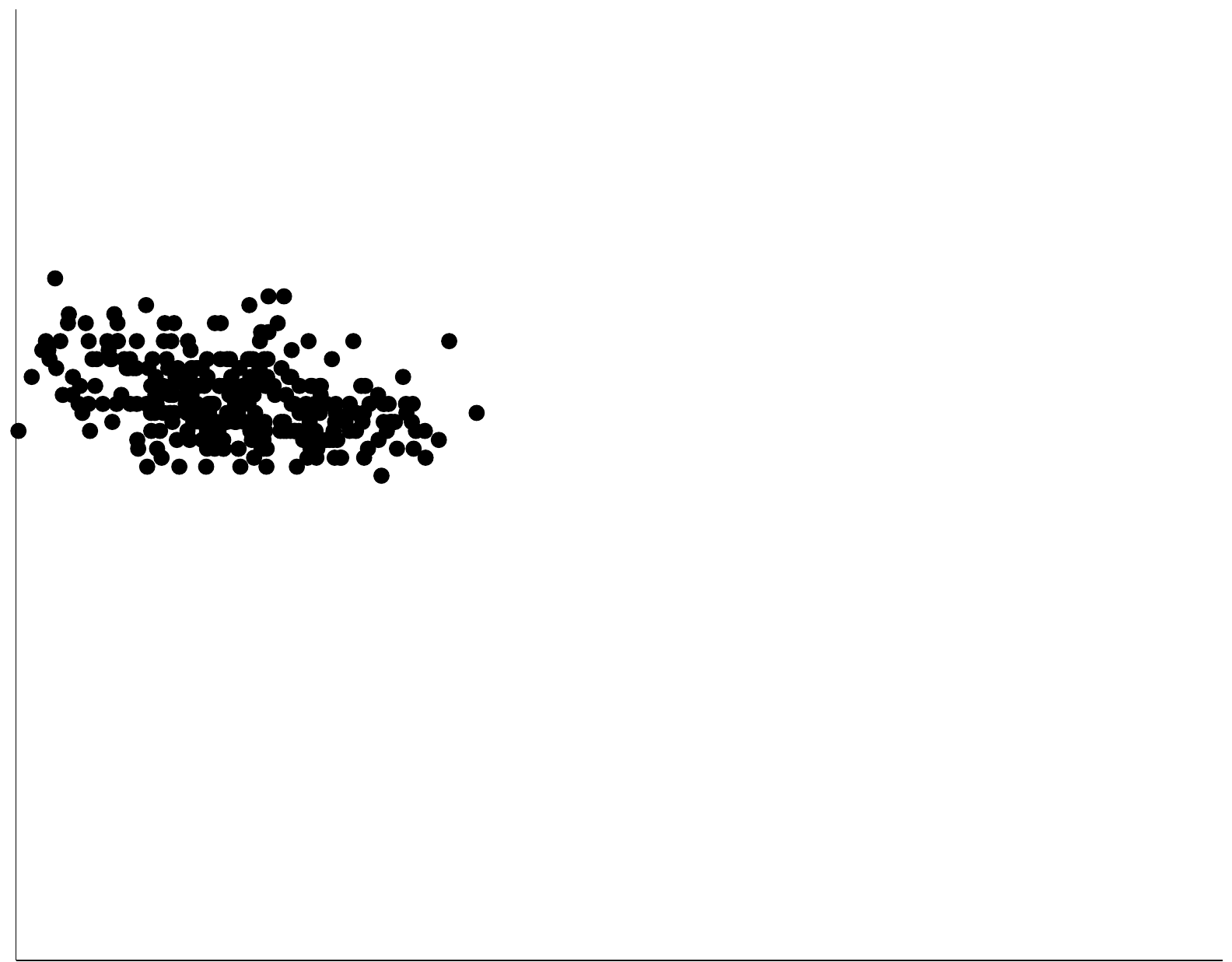}
\end{subfigure}&
\begin{subfigure}{0.1\textwidth}
    \includegraphics[height=10.5mm]{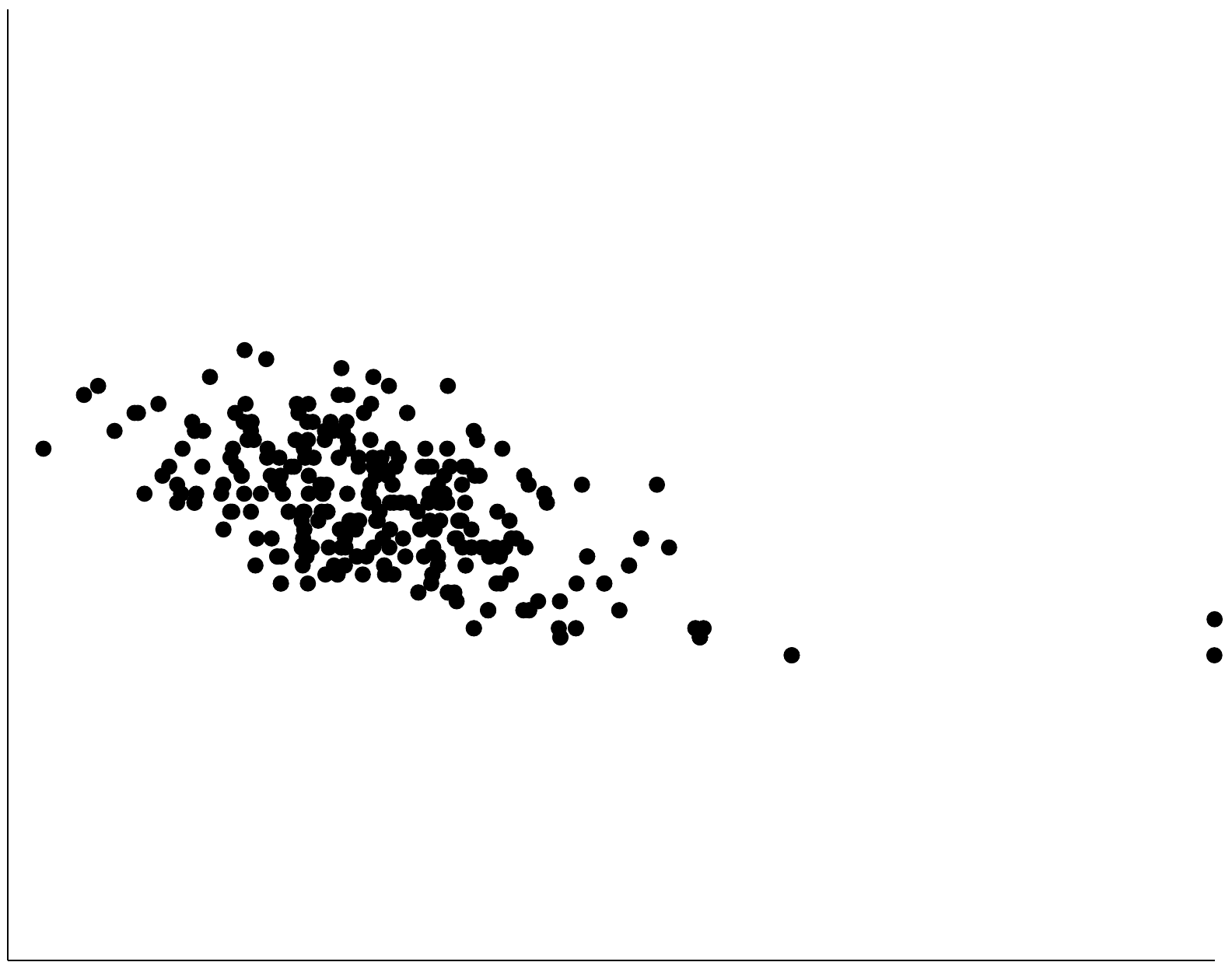}
\end{subfigure}&
\begin{subfigure}{0.1\textwidth}
    \includegraphics[height=10.5mm]{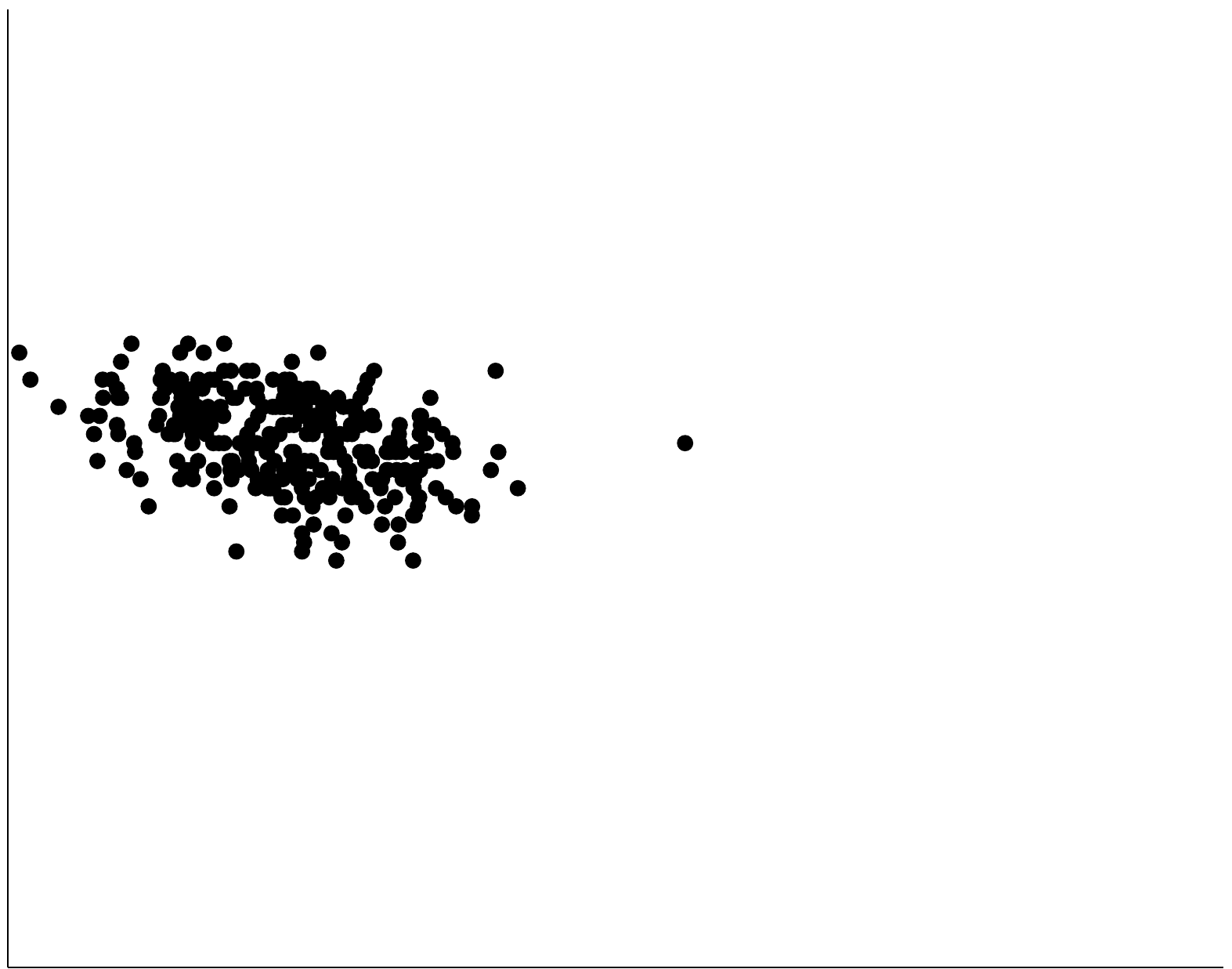}
\end{subfigure}&
\begin{subfigure}{0.1\textwidth}
    \includegraphics[height=10.5mm]{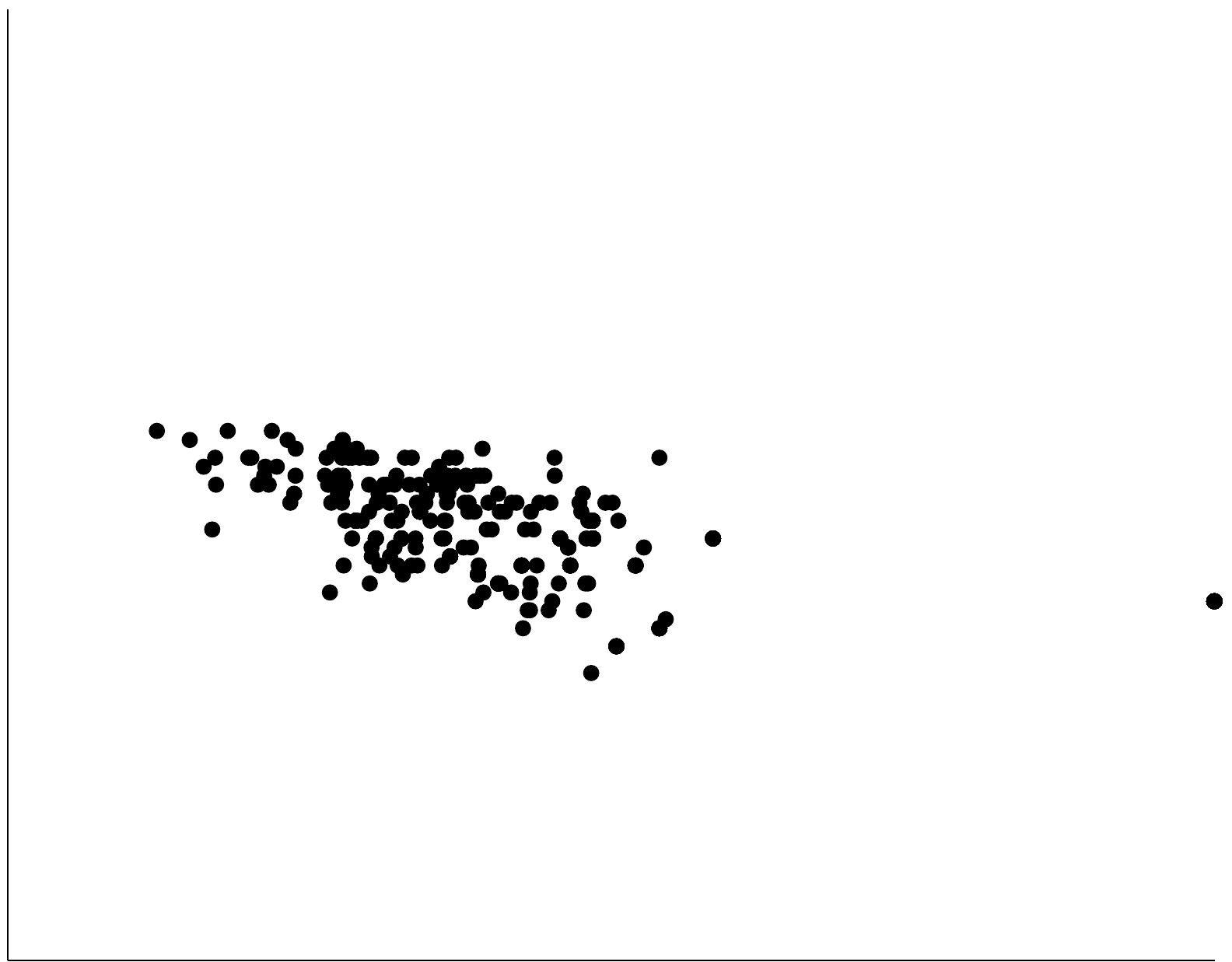}
\end{subfigure}\\ \\
New-Thyroid &
\begin{subfigure}{0.1\textwidth}
    \includegraphics[height=10.5mm]{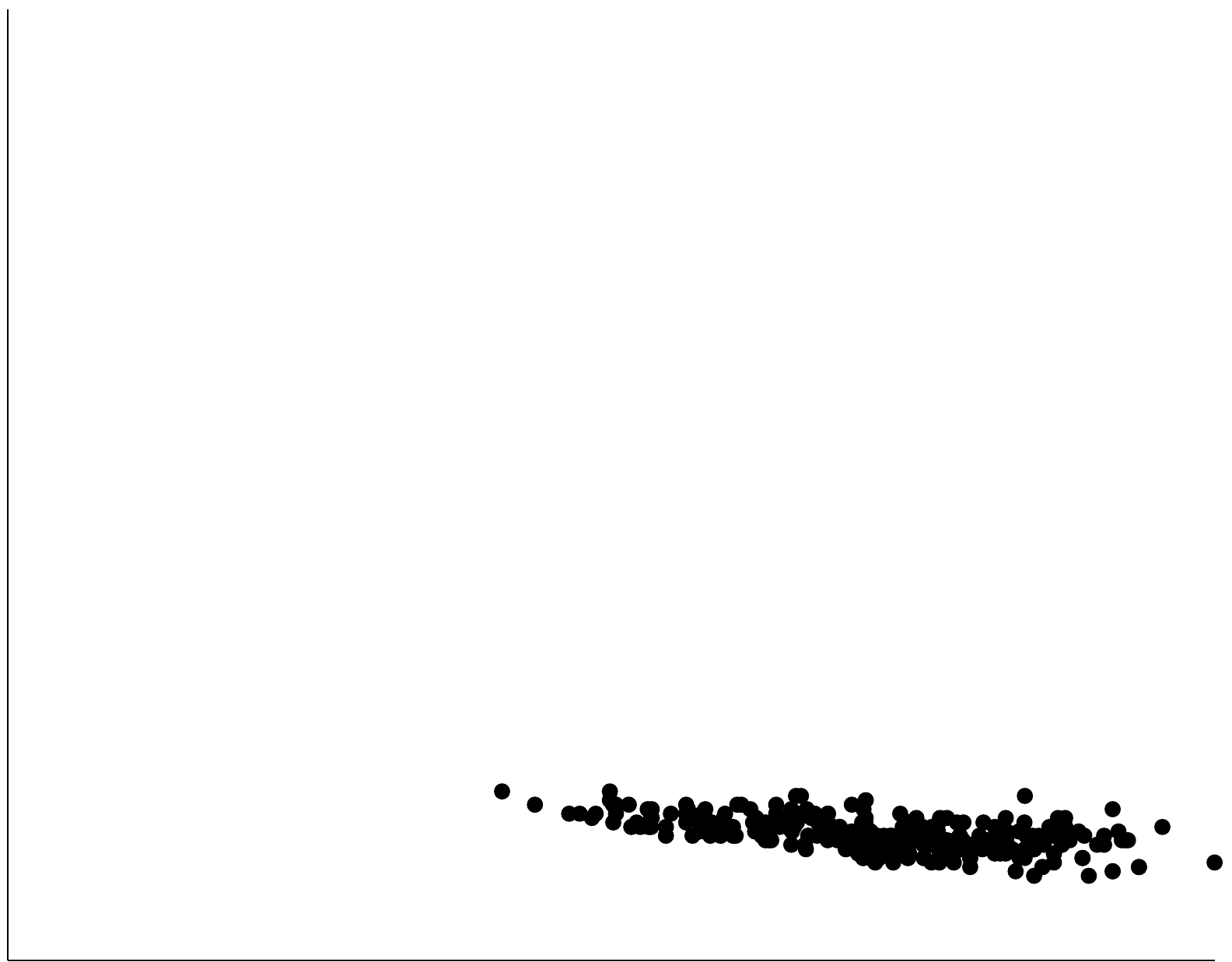}
\end{subfigure}&
\begin{subfigure}{0.1\textwidth}
    \includegraphics[height=10.5mm]{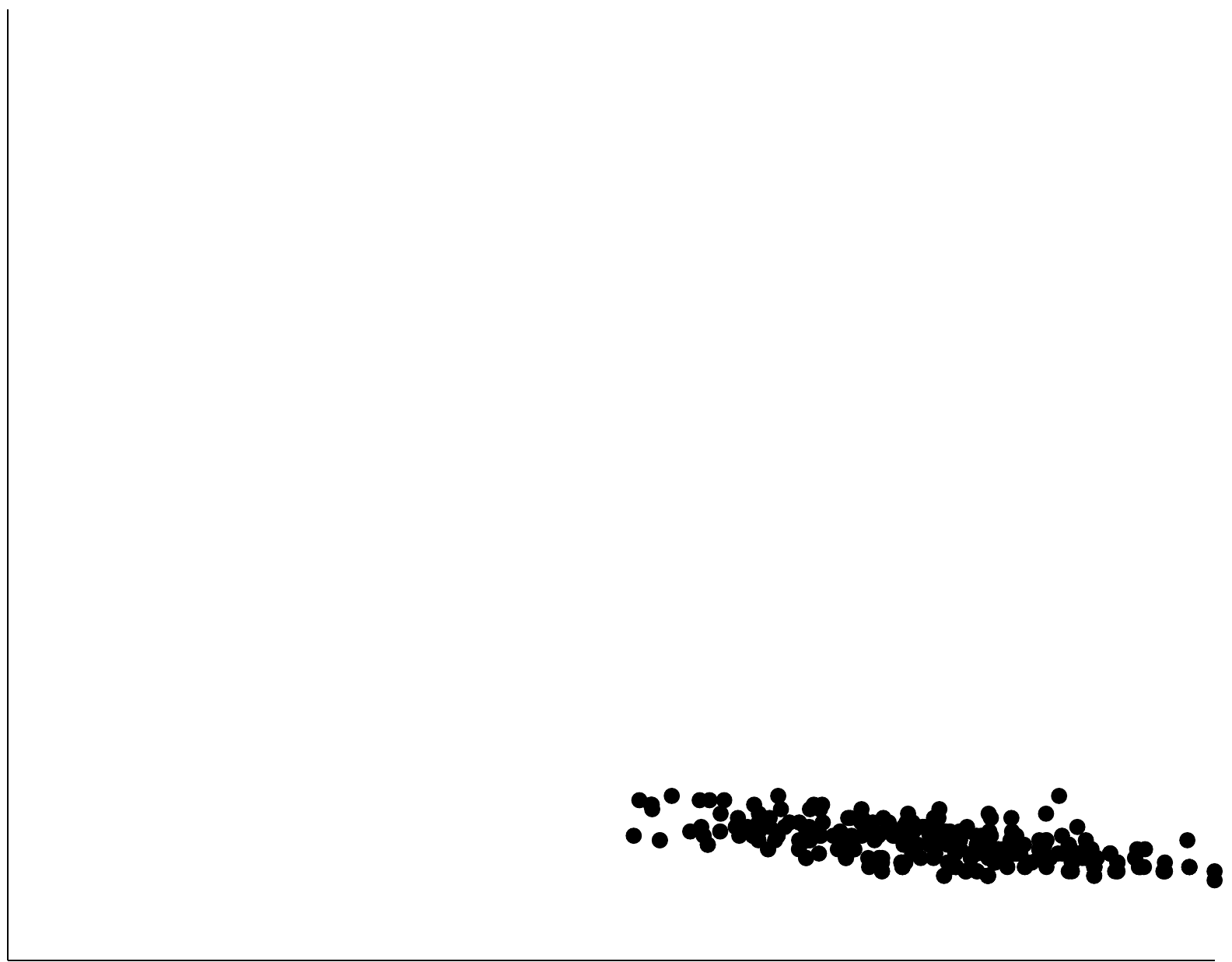}
\end{subfigure}&
\begin{subfigure}{0.1\textwidth}
    \includegraphics[height=10.5mm]{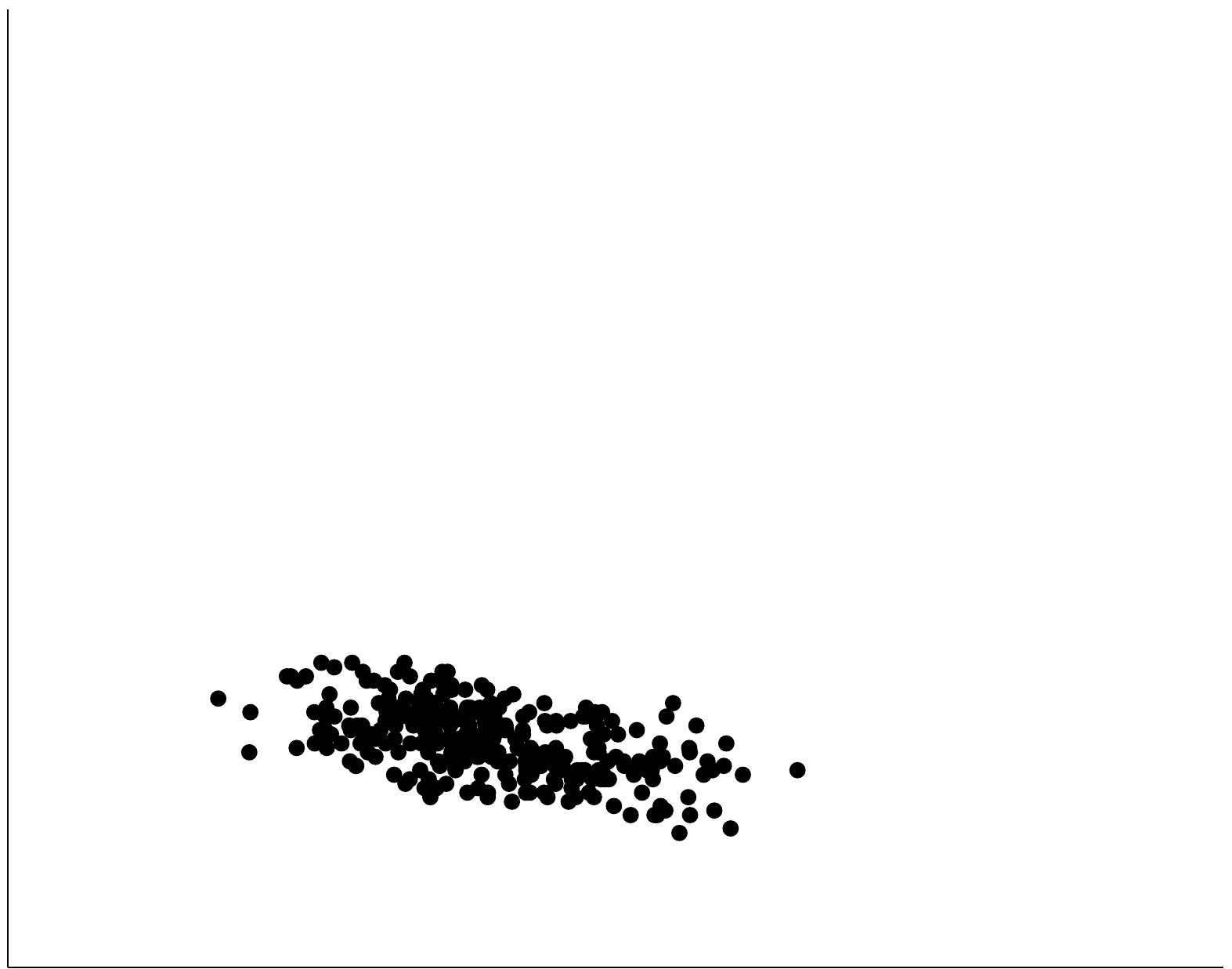}
\end{subfigure}&
\begin{subfigure}{0.1\textwidth}
    \includegraphics[height=10.5mm]{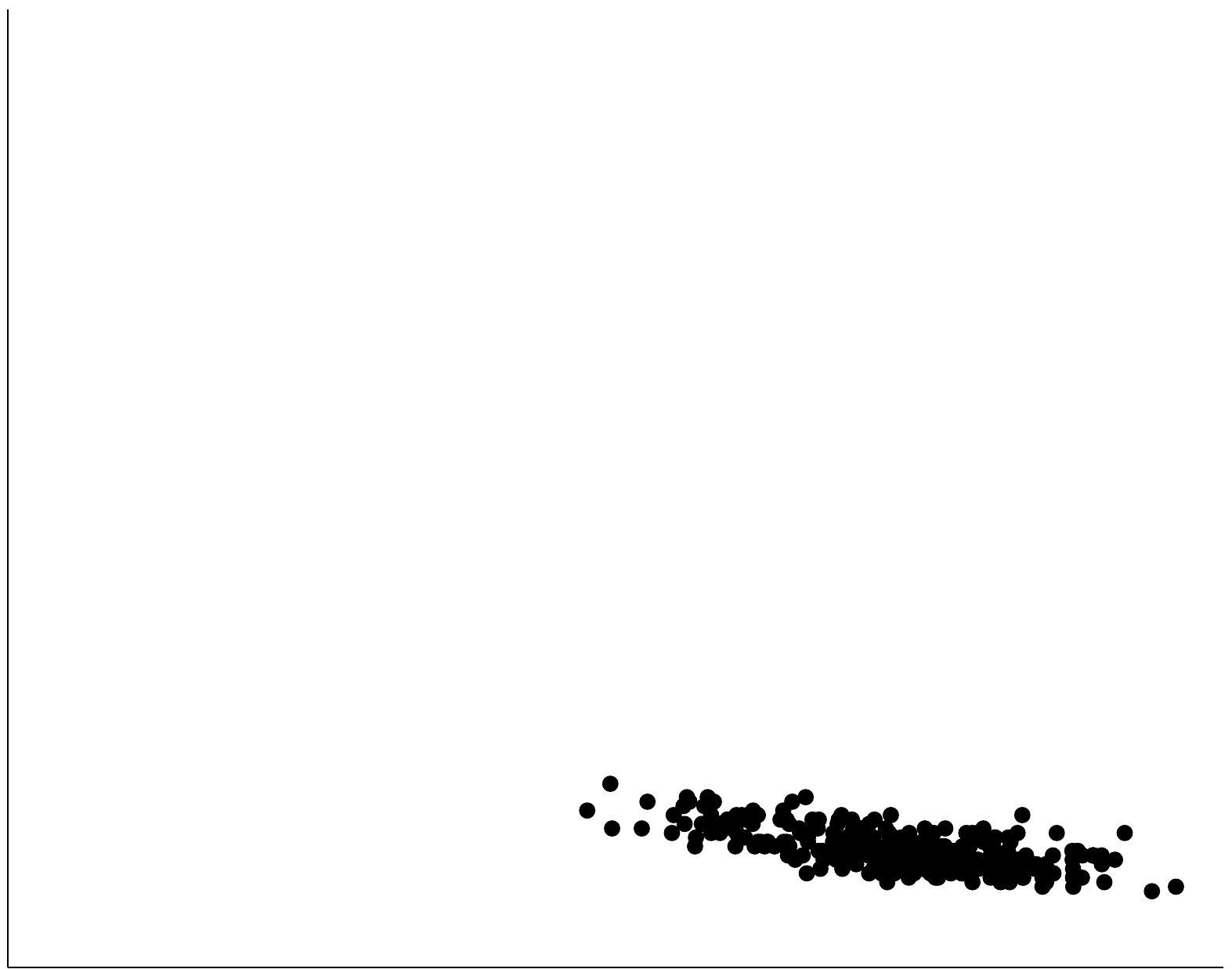}
\end{subfigure}&
\begin{subfigure}{0.1\textwidth}
    \includegraphics[height=10.5mm]{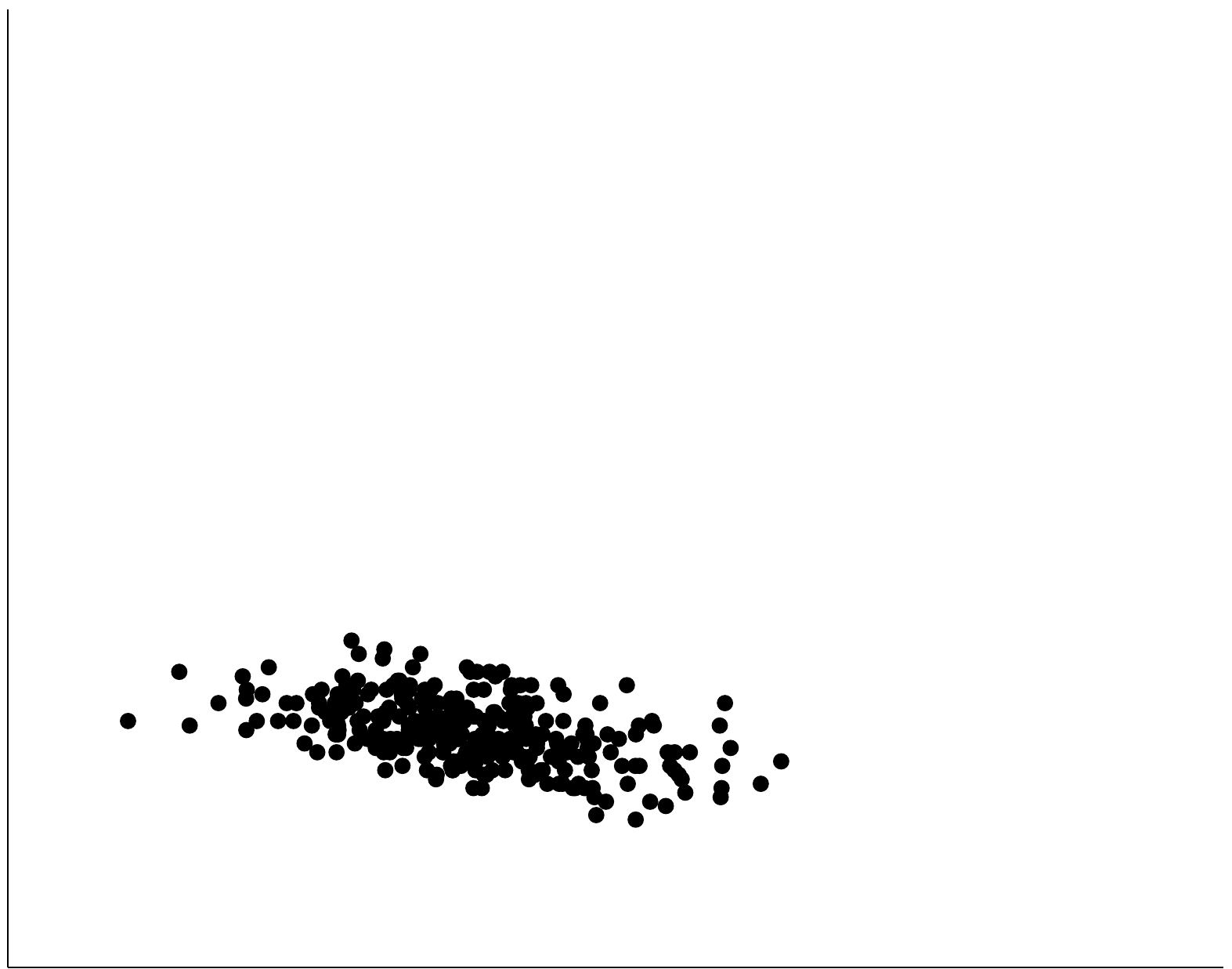}
\end{subfigure}&
\begin{subfigure}{0.1\textwidth}
    \includegraphics[height=10.5mm]{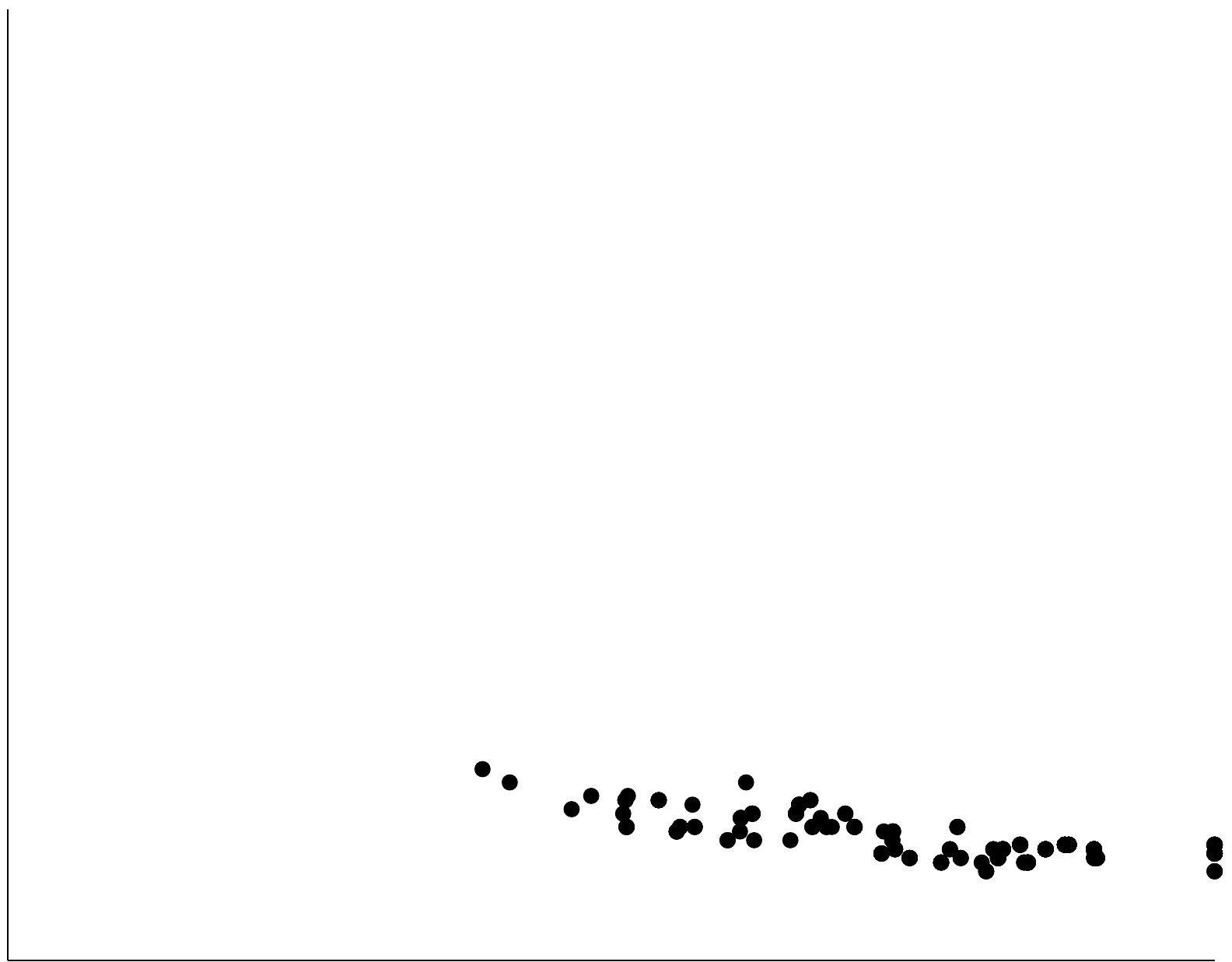}
\end{subfigure}&
\begin{subfigure}{0.1\textwidth}
    \includegraphics[height=10.5mm]{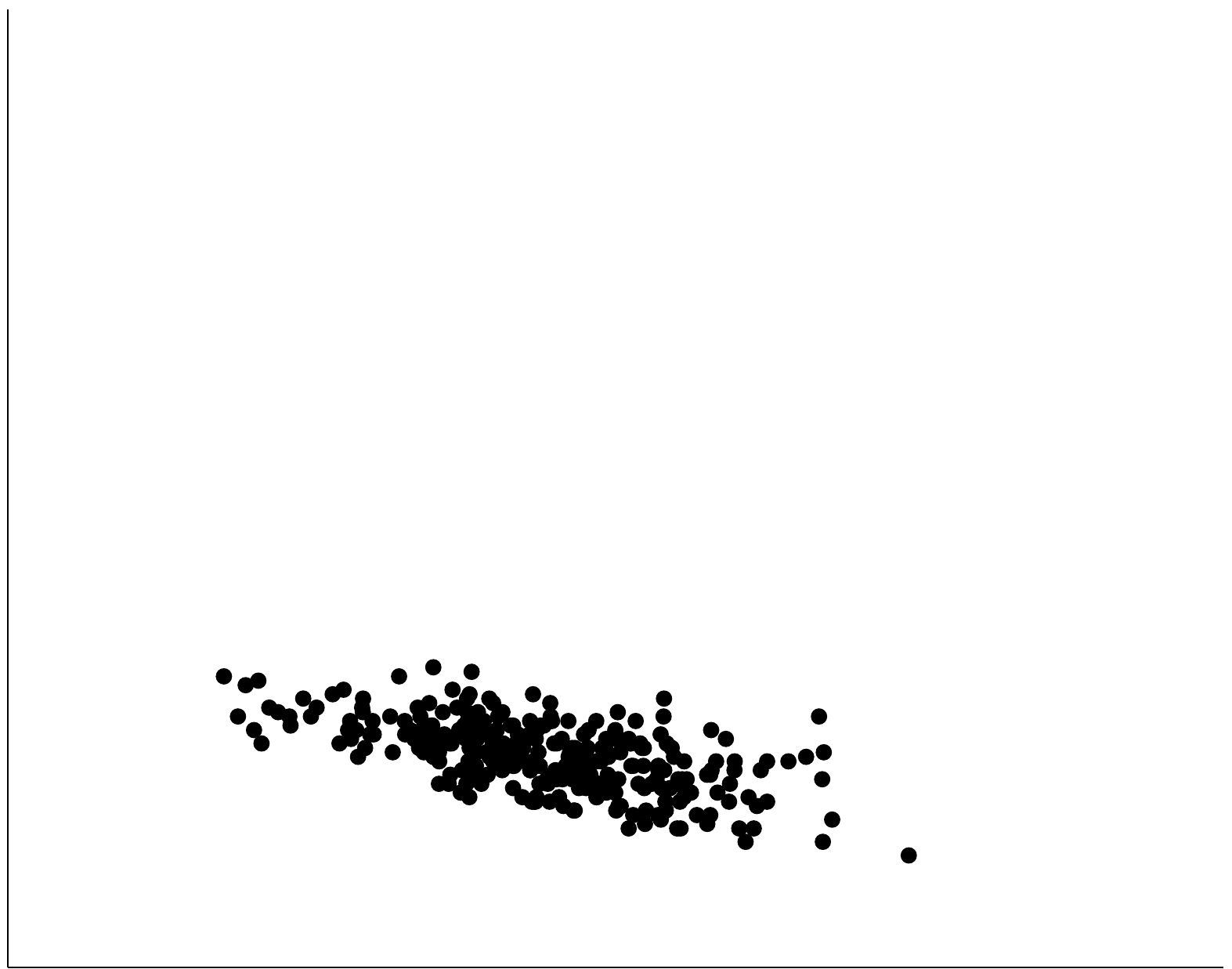}
\end{subfigure}&
\begin{subfigure}{0.1\textwidth}
    \includegraphics[height=10.5mm]{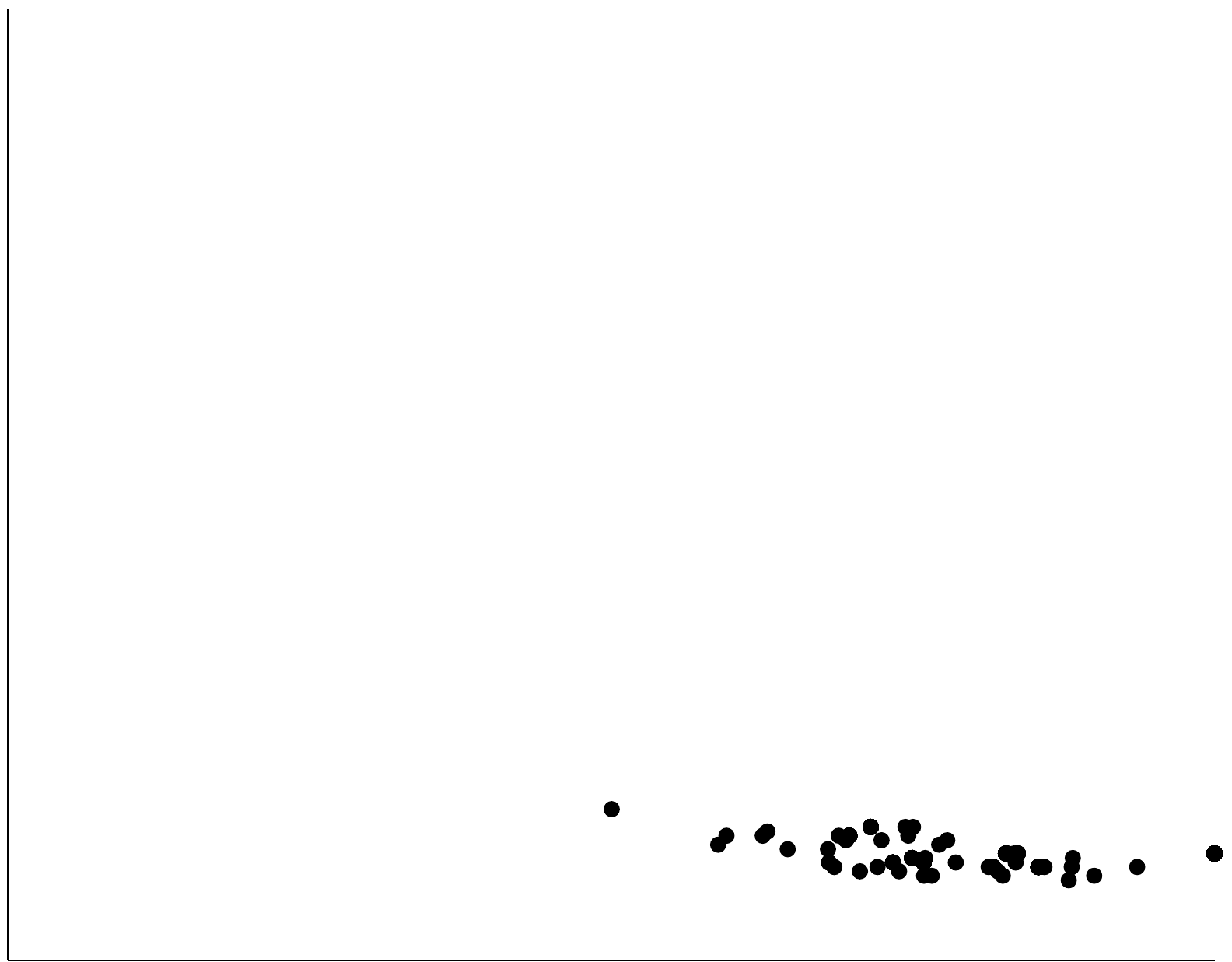}
\end{subfigure}\\ \\
Column &
\begin{subfigure}{0.1\textwidth}
    \includegraphics[height=10.5mm]{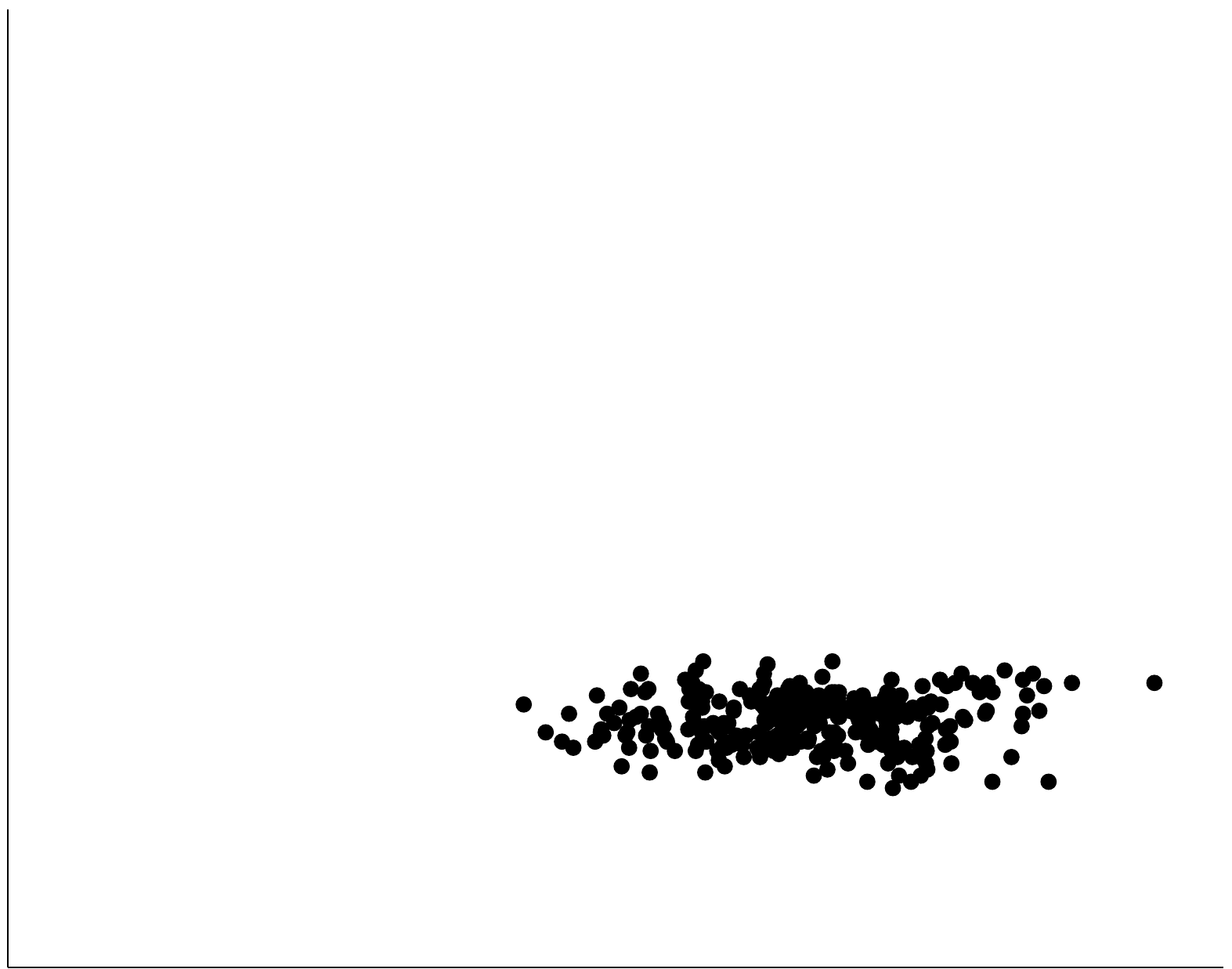}
\end{subfigure}&
\begin{subfigure}{0.1\textwidth}
    \includegraphics[height=10.5mm]{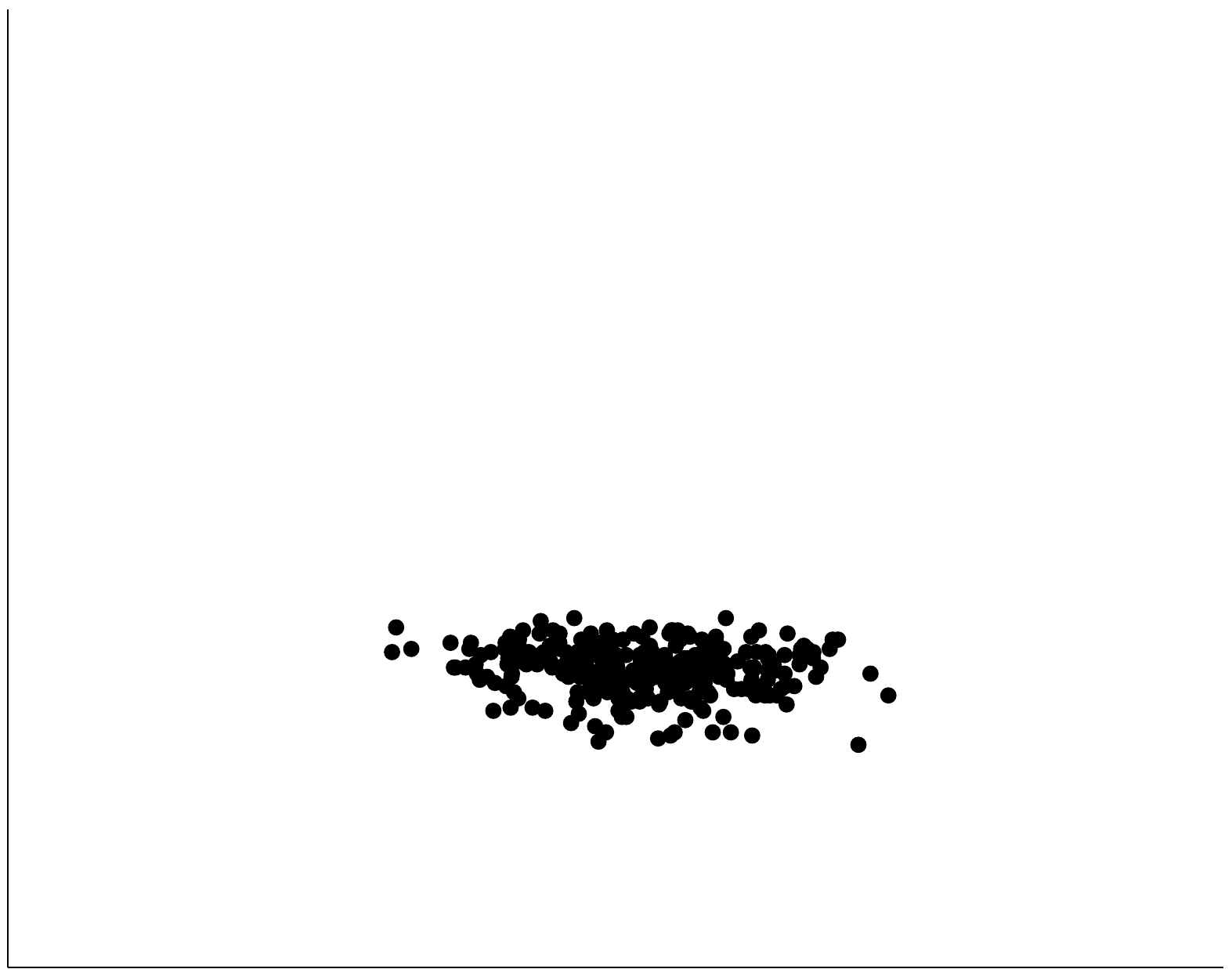}
\end{subfigure}&
\begin{subfigure}{0.1\textwidth}
    \includegraphics[height=10.5mm]{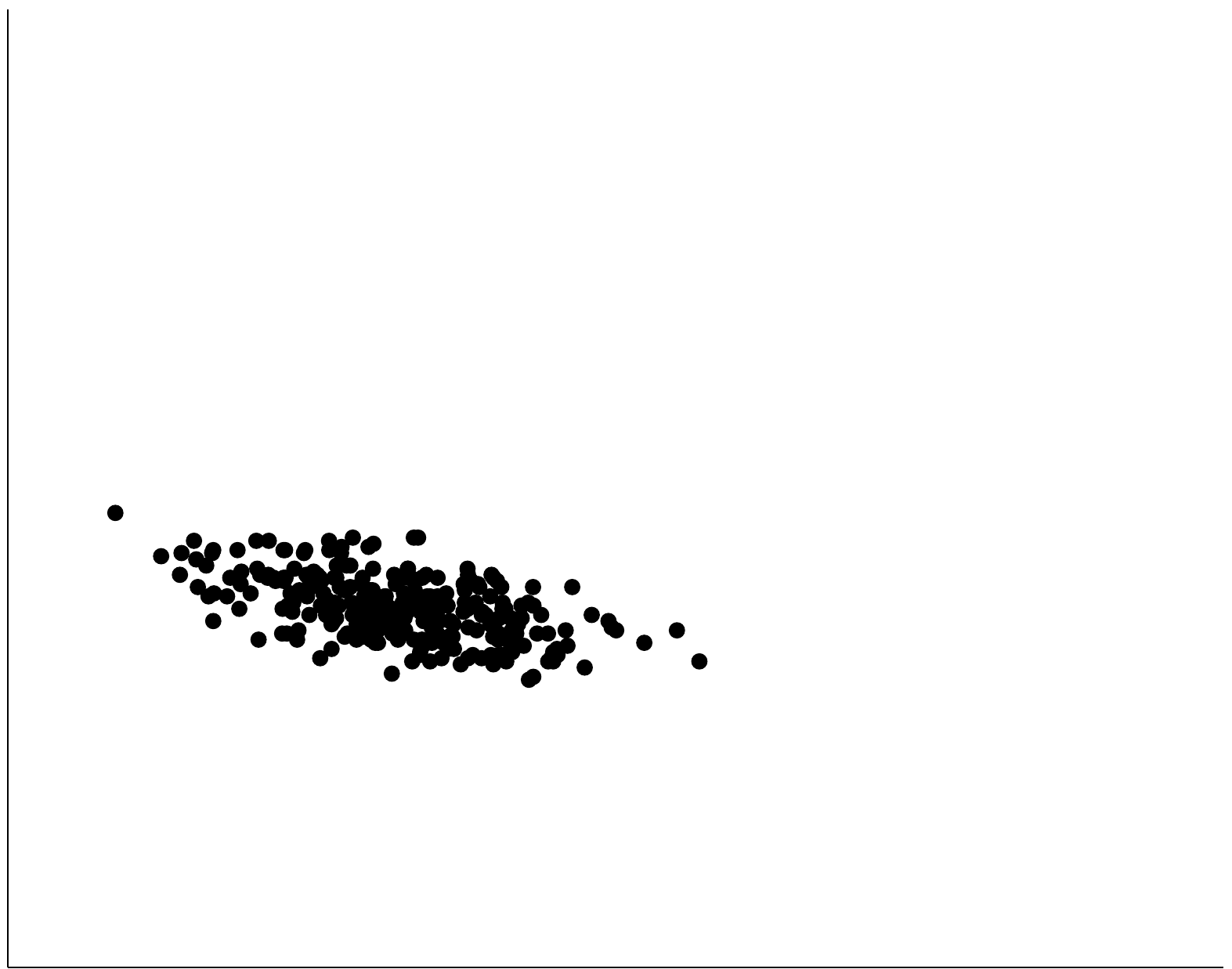}
\end{subfigure}&
\begin{subfigure}{0.1\textwidth}
    \includegraphics[height=10.5mm]{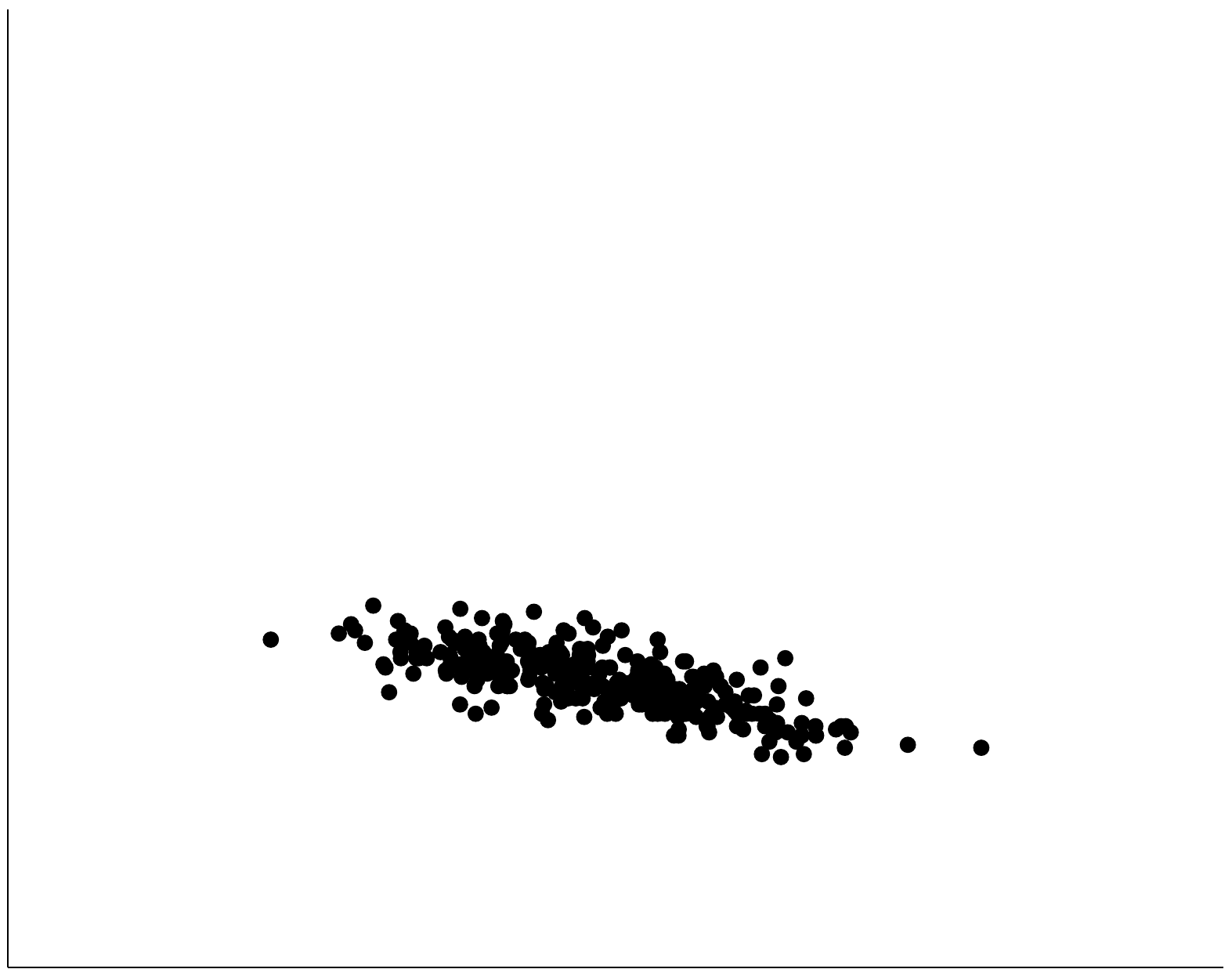}
\end{subfigure}&
\begin{subfigure}{0.1\textwidth}
    \includegraphics[height=10.5mm]{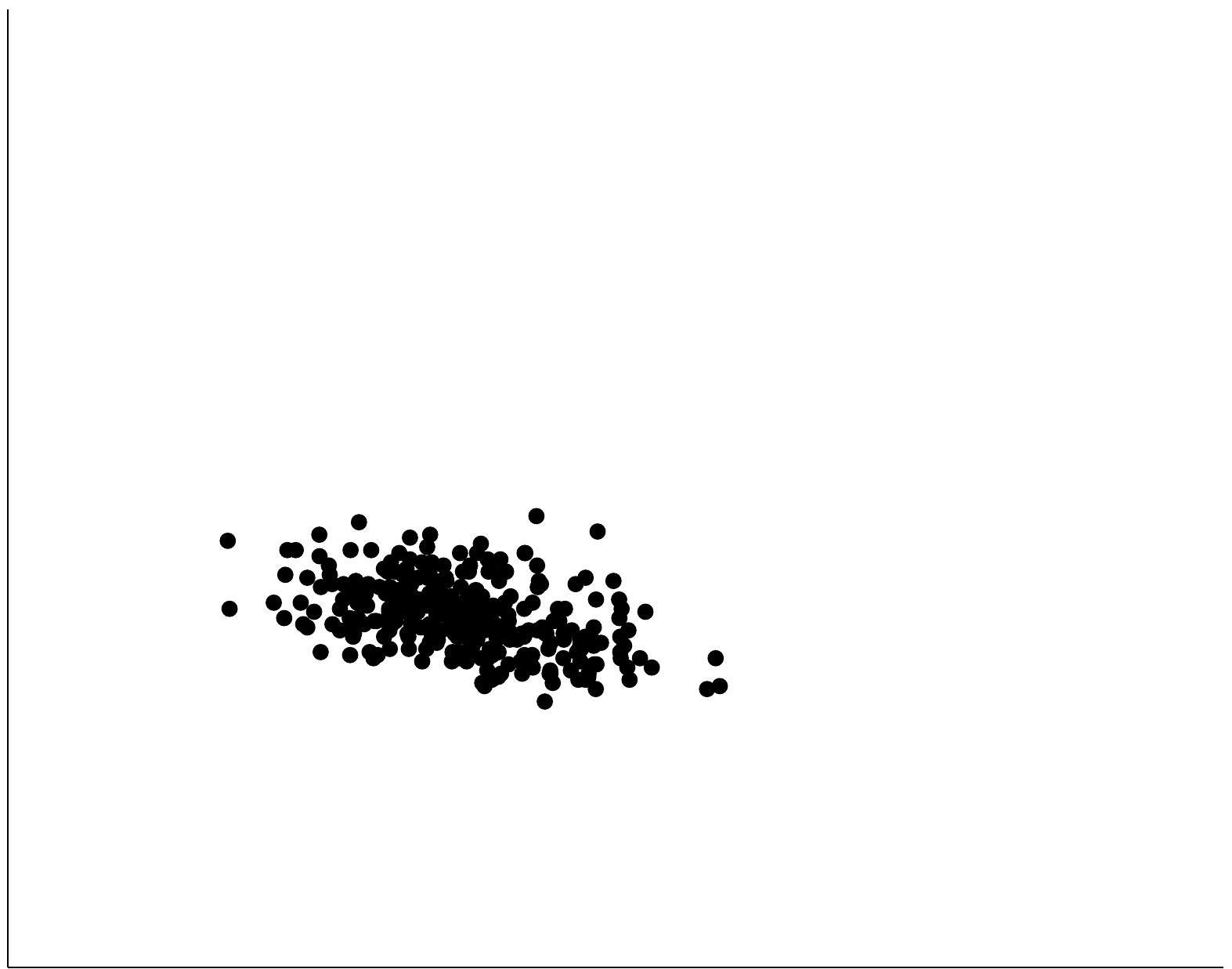}
\end{subfigure}&
\begin{subfigure}{0.1\textwidth}
    \includegraphics[height=10.5mm]{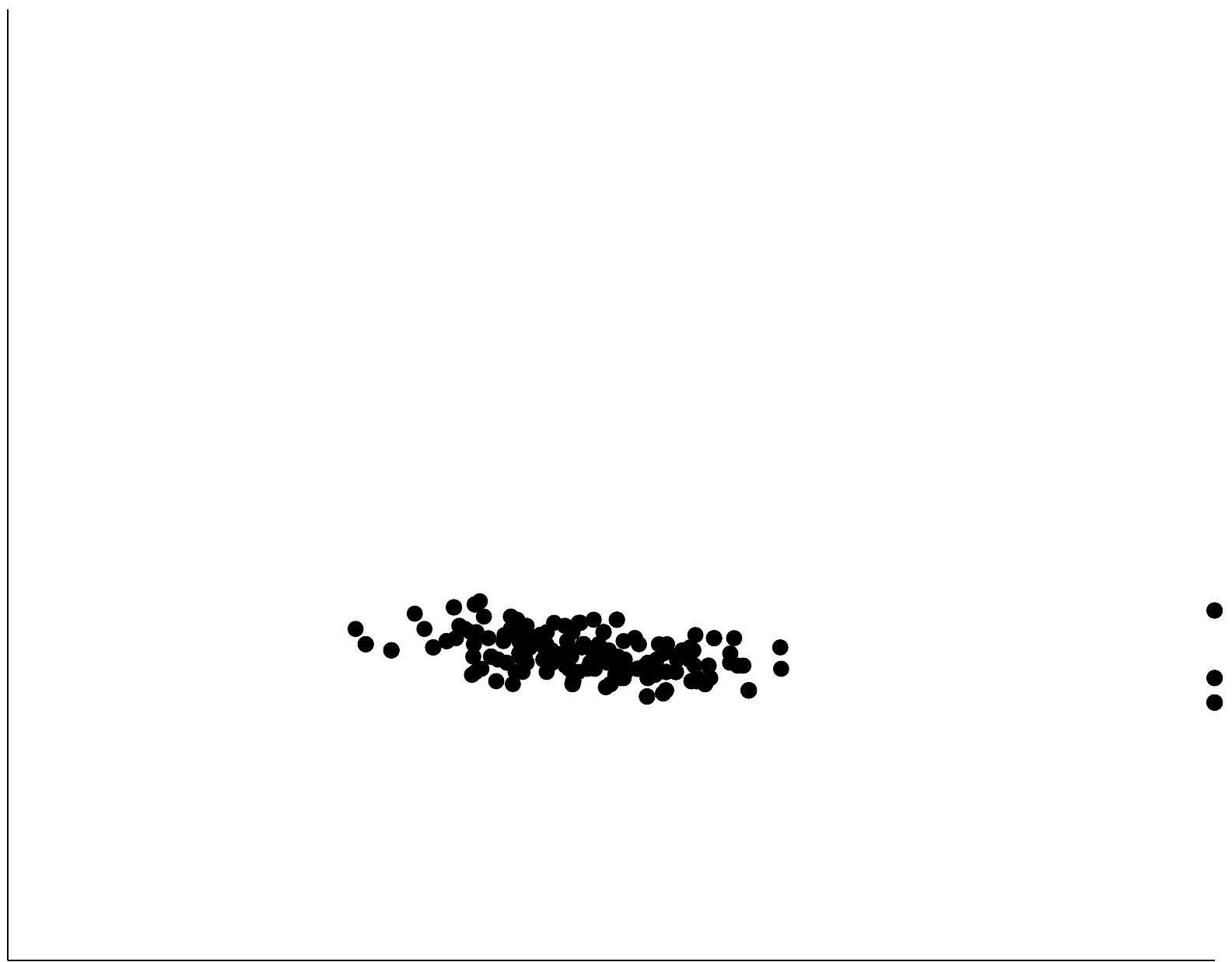}
\end{subfigure}&
\begin{subfigure}{0.1\textwidth}
    \includegraphics[height=10.5mm]{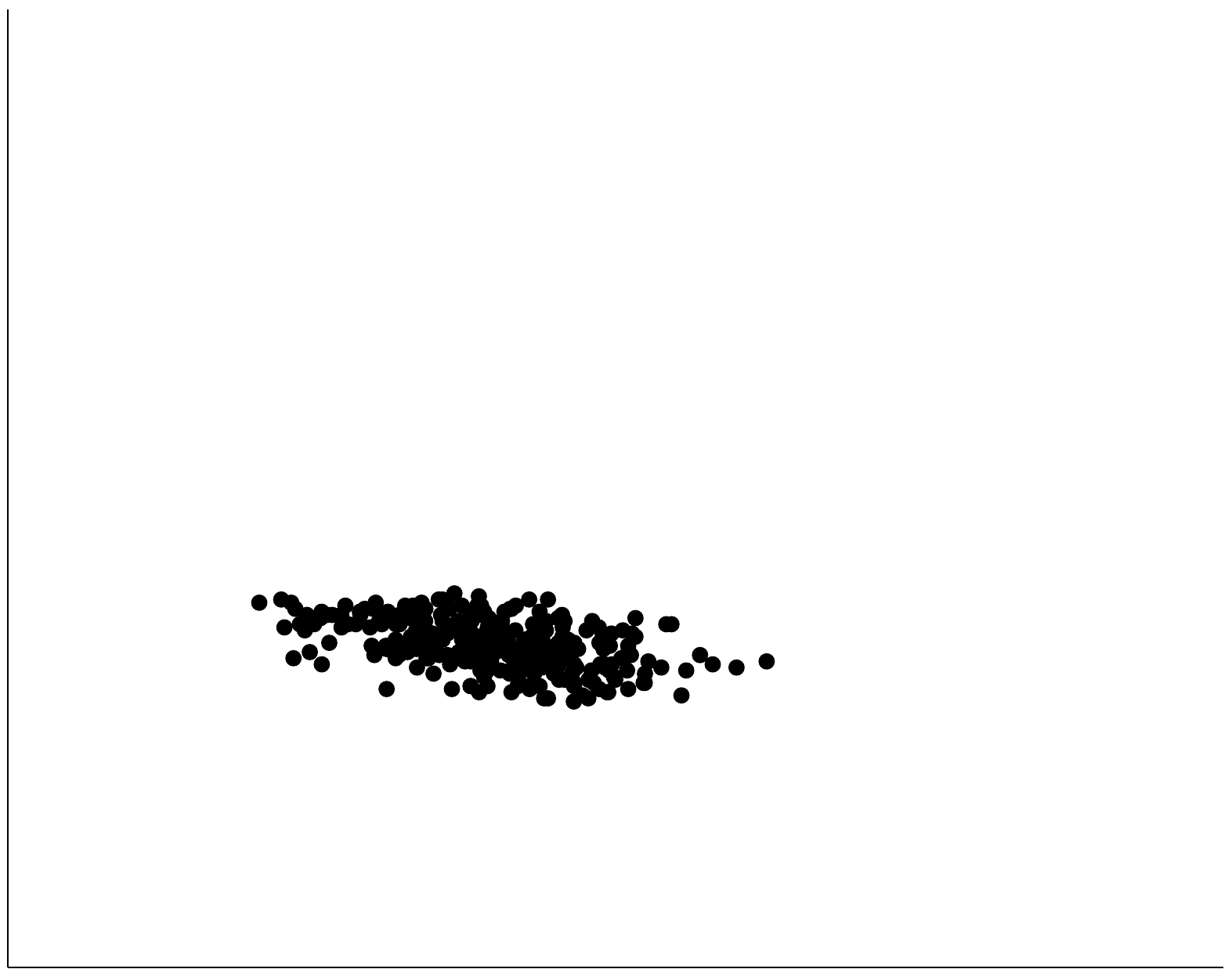}
\end{subfigure}&
\begin{subfigure}{0.1\textwidth}
    \includegraphics[height=10.5mm]{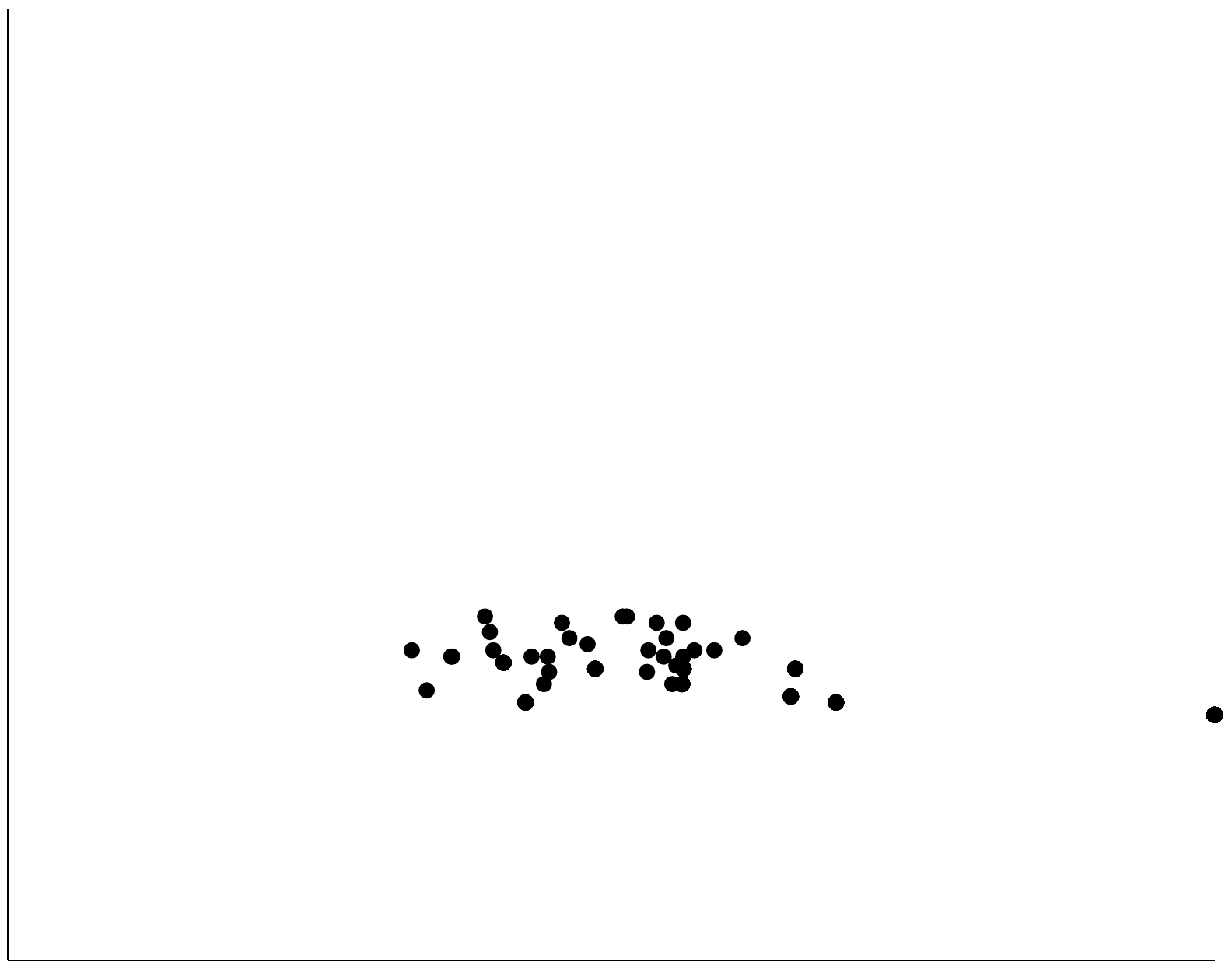}
\end{subfigure}\\ \\
Seeds &
\begin{subfigure}{0.1\textwidth}
    \includegraphics[height=10.5mm]{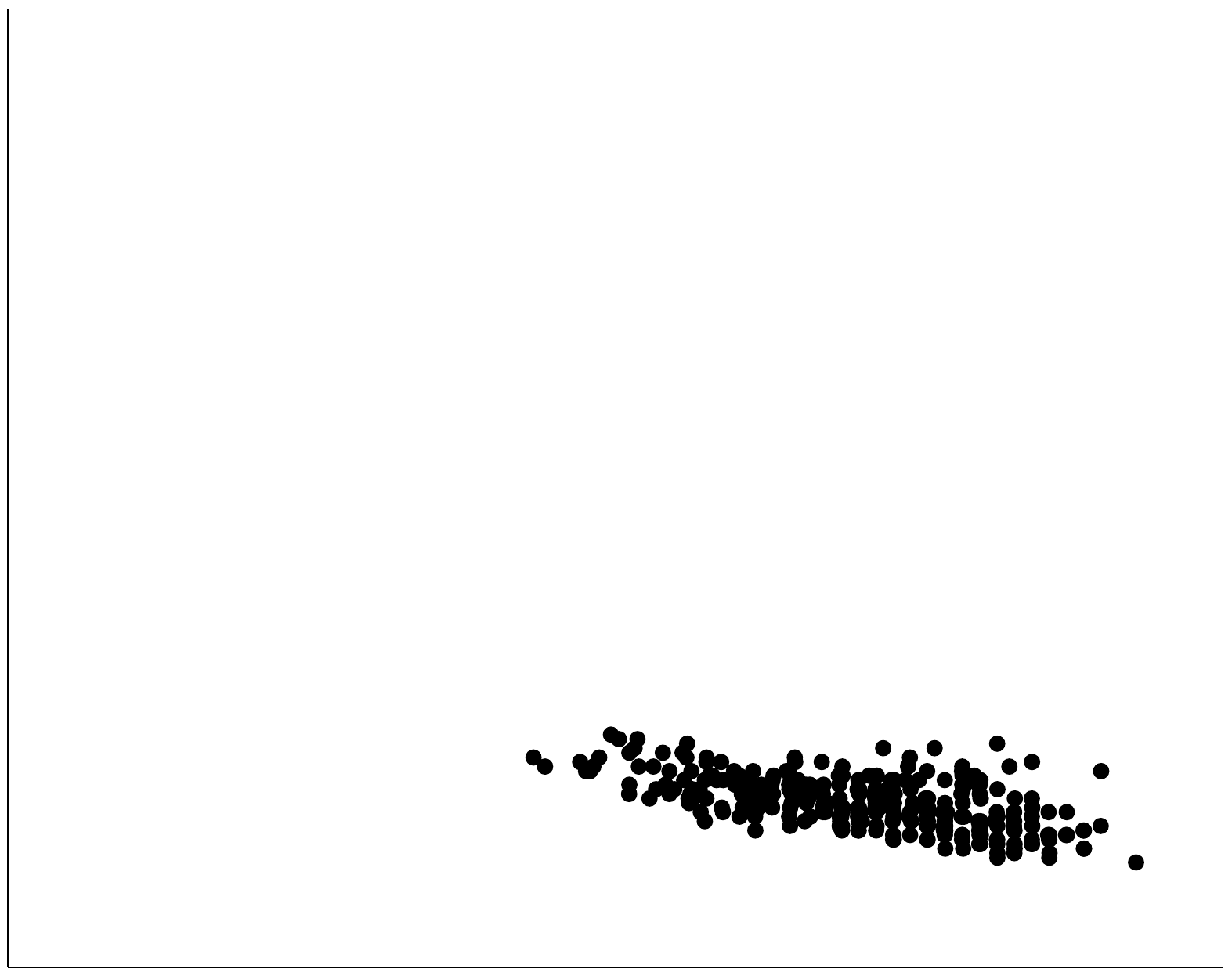}
\end{subfigure}&
\begin{subfigure}{0.1\textwidth}
    \includegraphics[height=10.5mm]{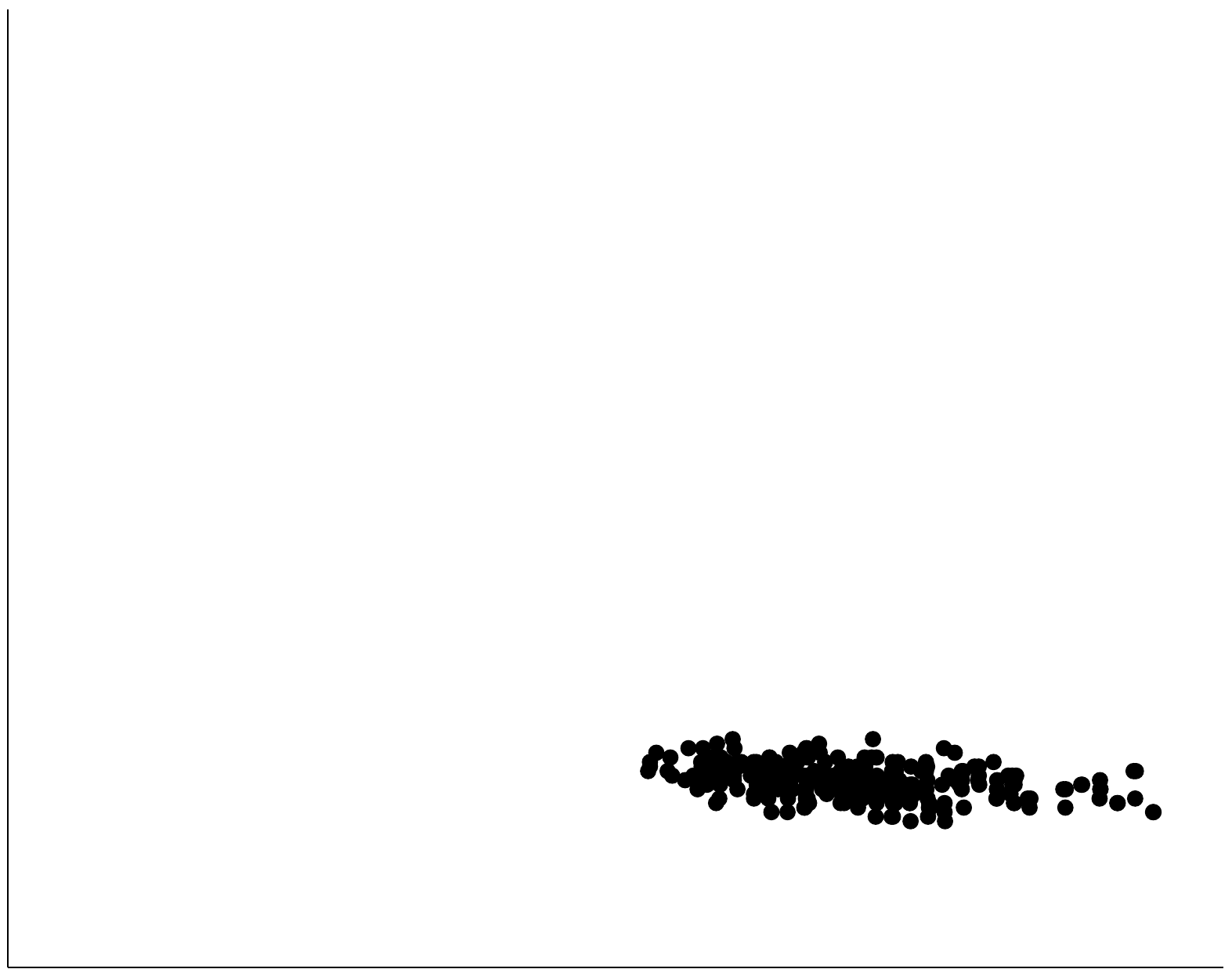}
\end{subfigure}&
\begin{subfigure}{0.1\textwidth}
    \includegraphics[height=10.5mm]{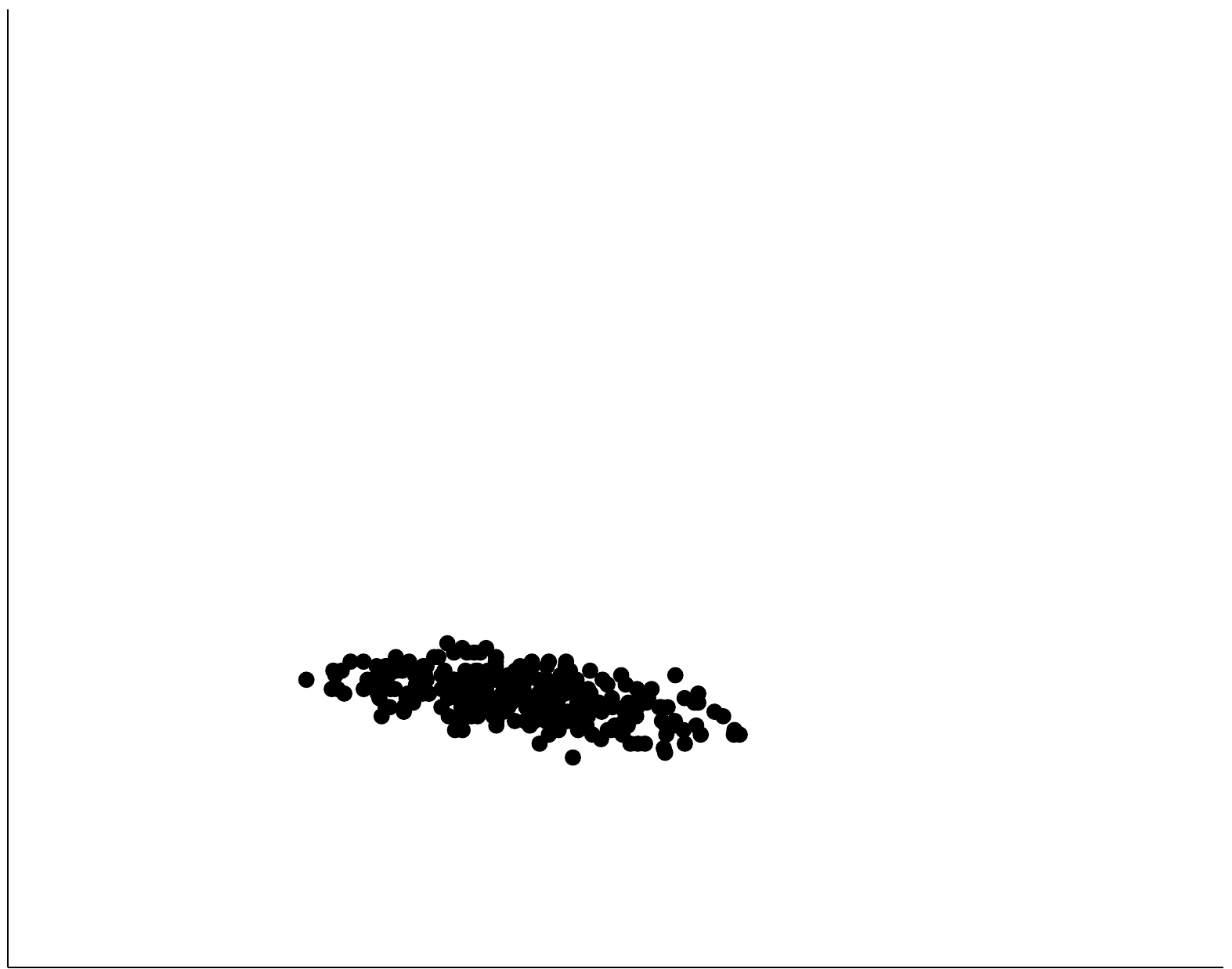}
\end{subfigure}&
\begin{subfigure}{0.1\textwidth}
    \includegraphics[height=10.5mm]{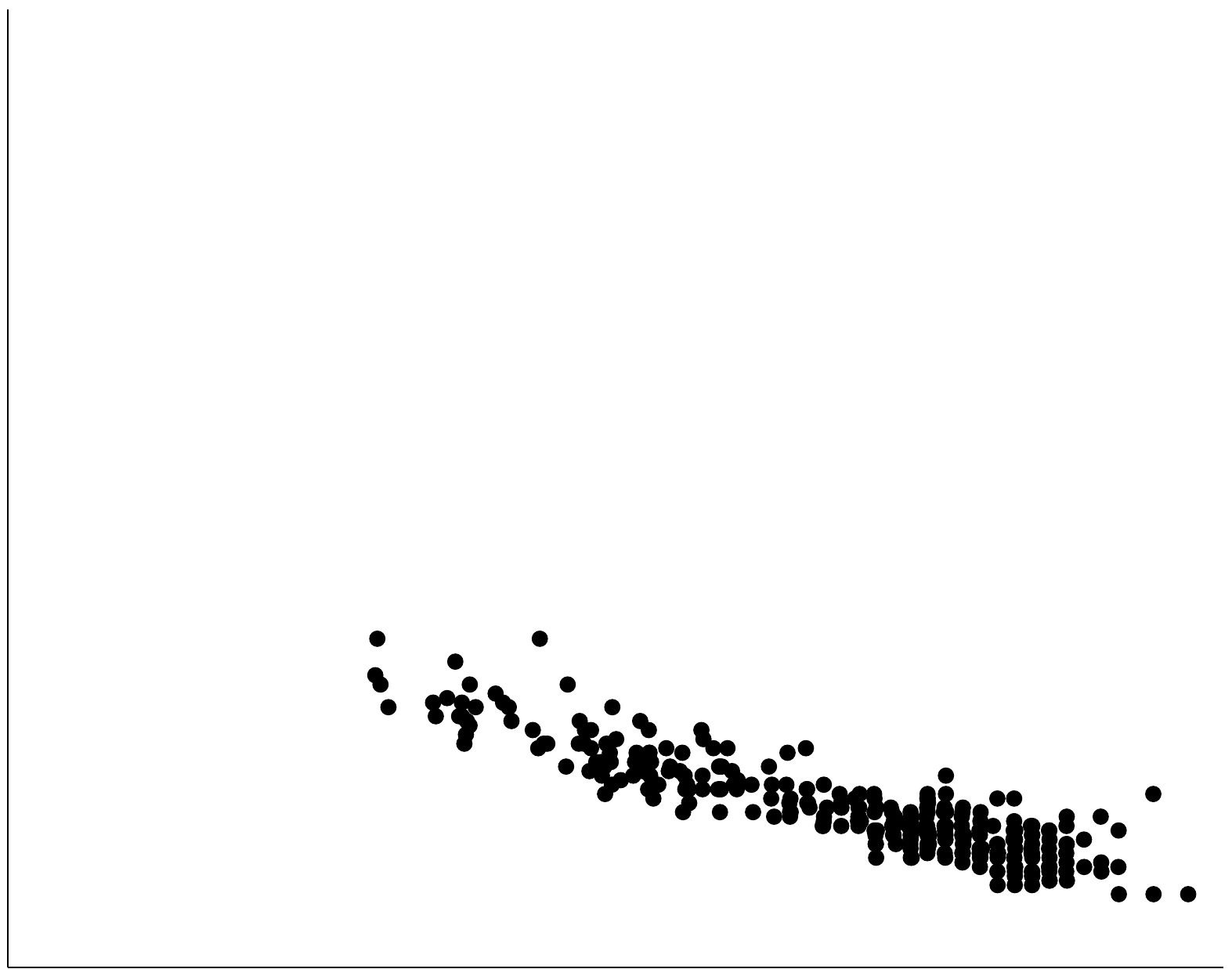}
\end{subfigure}&
\begin{subfigure}{0.1\textwidth}
    \includegraphics[height=10.5mm]{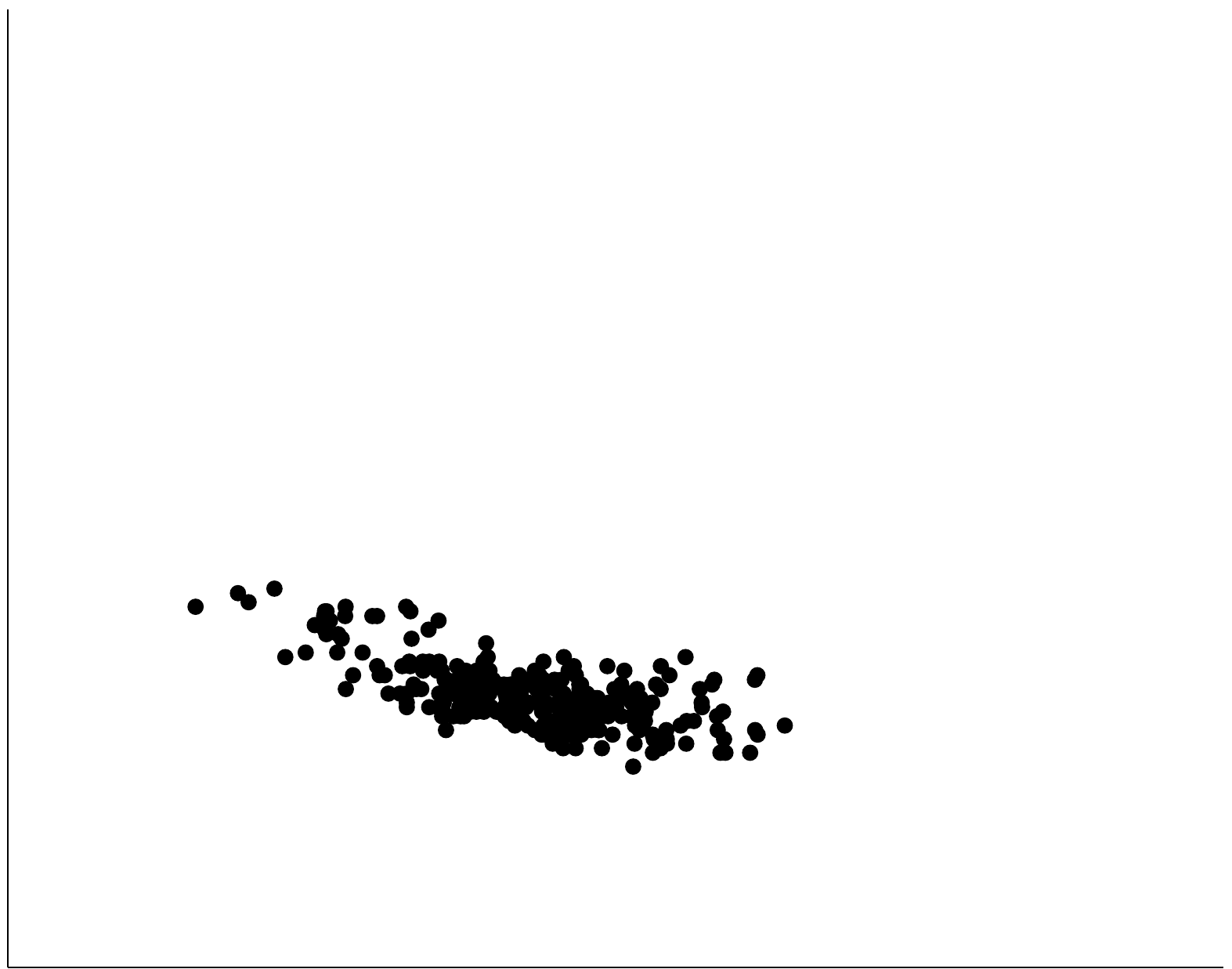}
\end{subfigure}&
\begin{subfigure}{0.1\textwidth}
    \includegraphics[height=10.5mm]{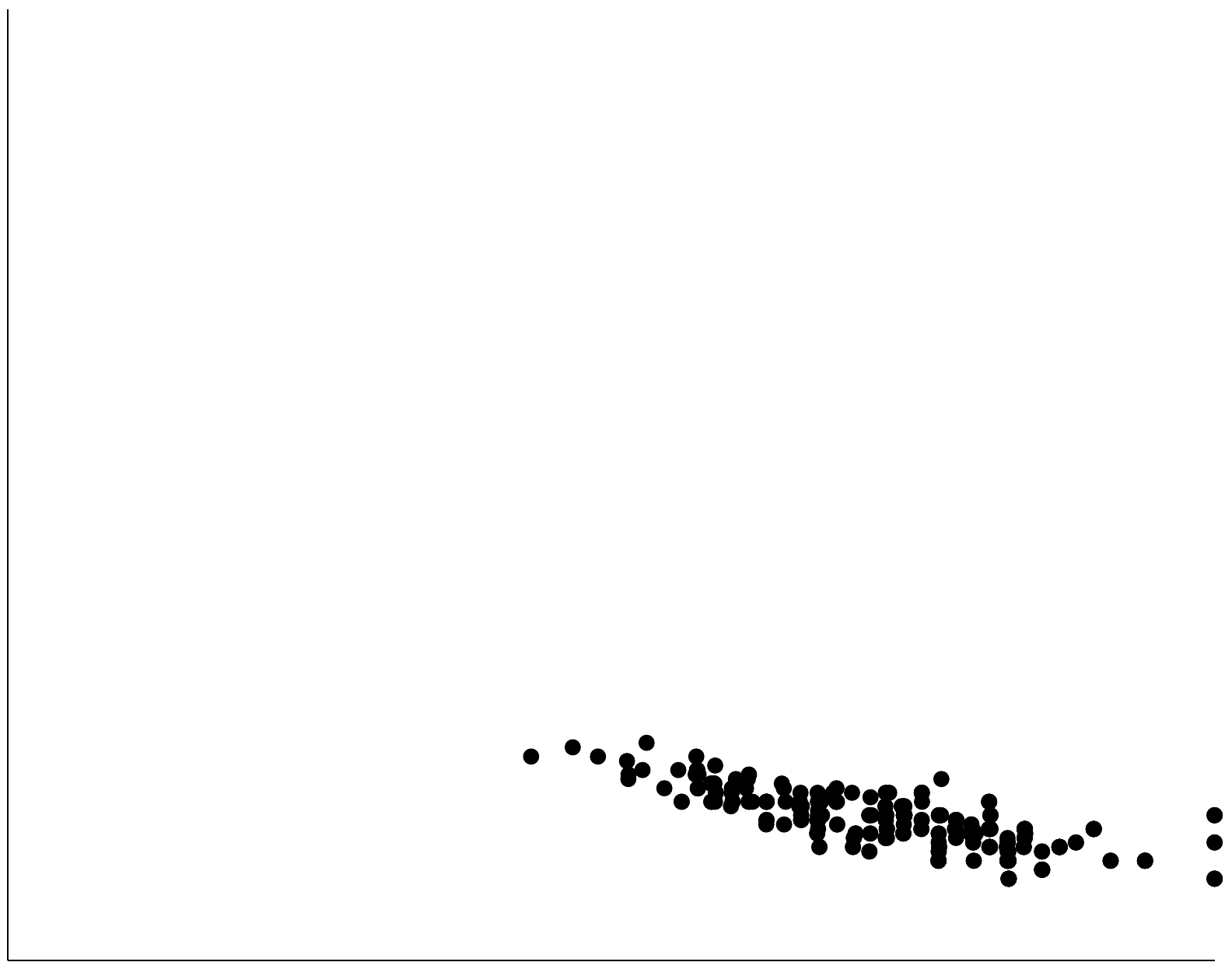}
\end{subfigure}&
\begin{subfigure}{0.1\textwidth}
    \includegraphics[height=10.5mm]{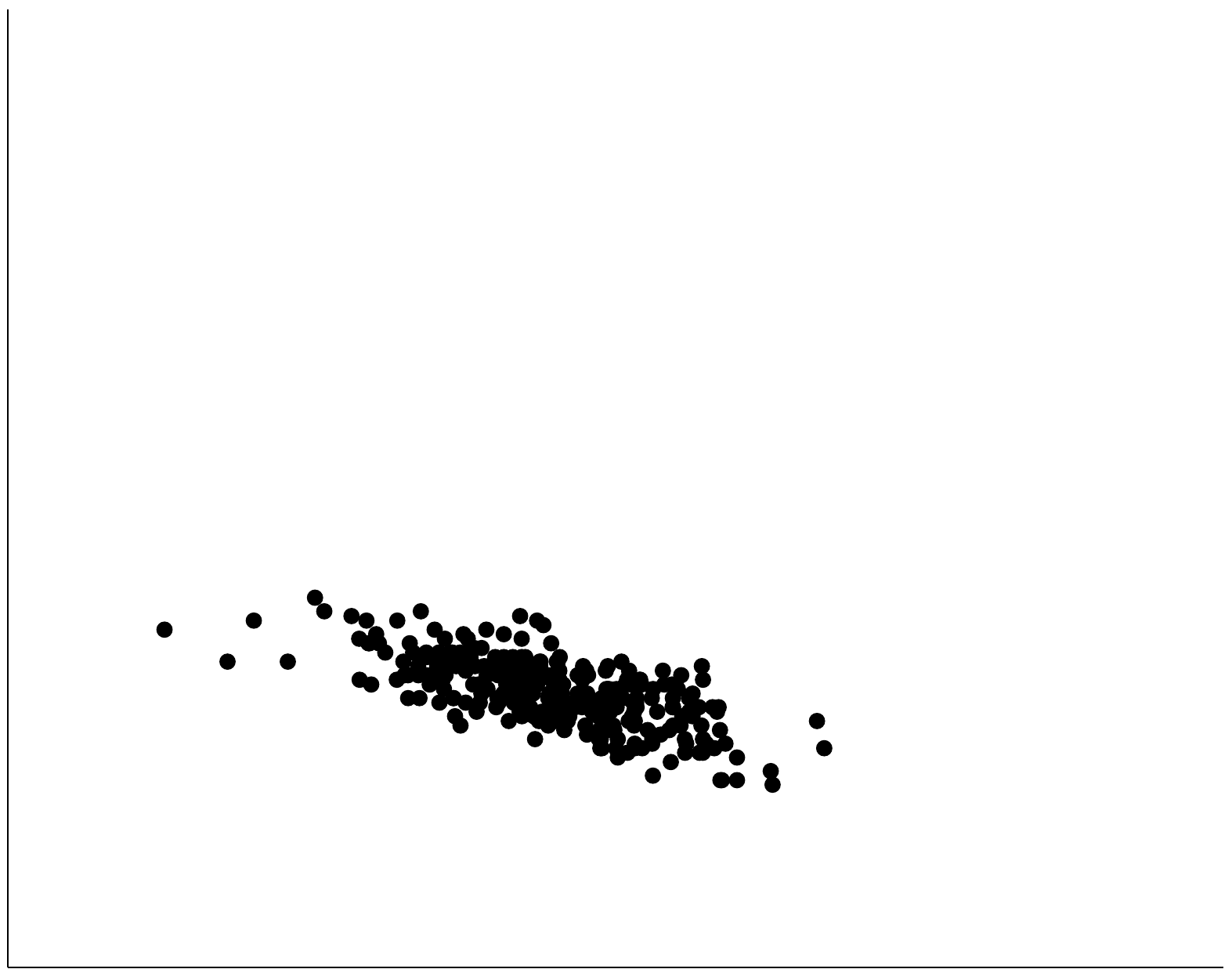}
\end{subfigure}&
\begin{subfigure}{0.1\textwidth}
    \includegraphics[height=10.5mm]{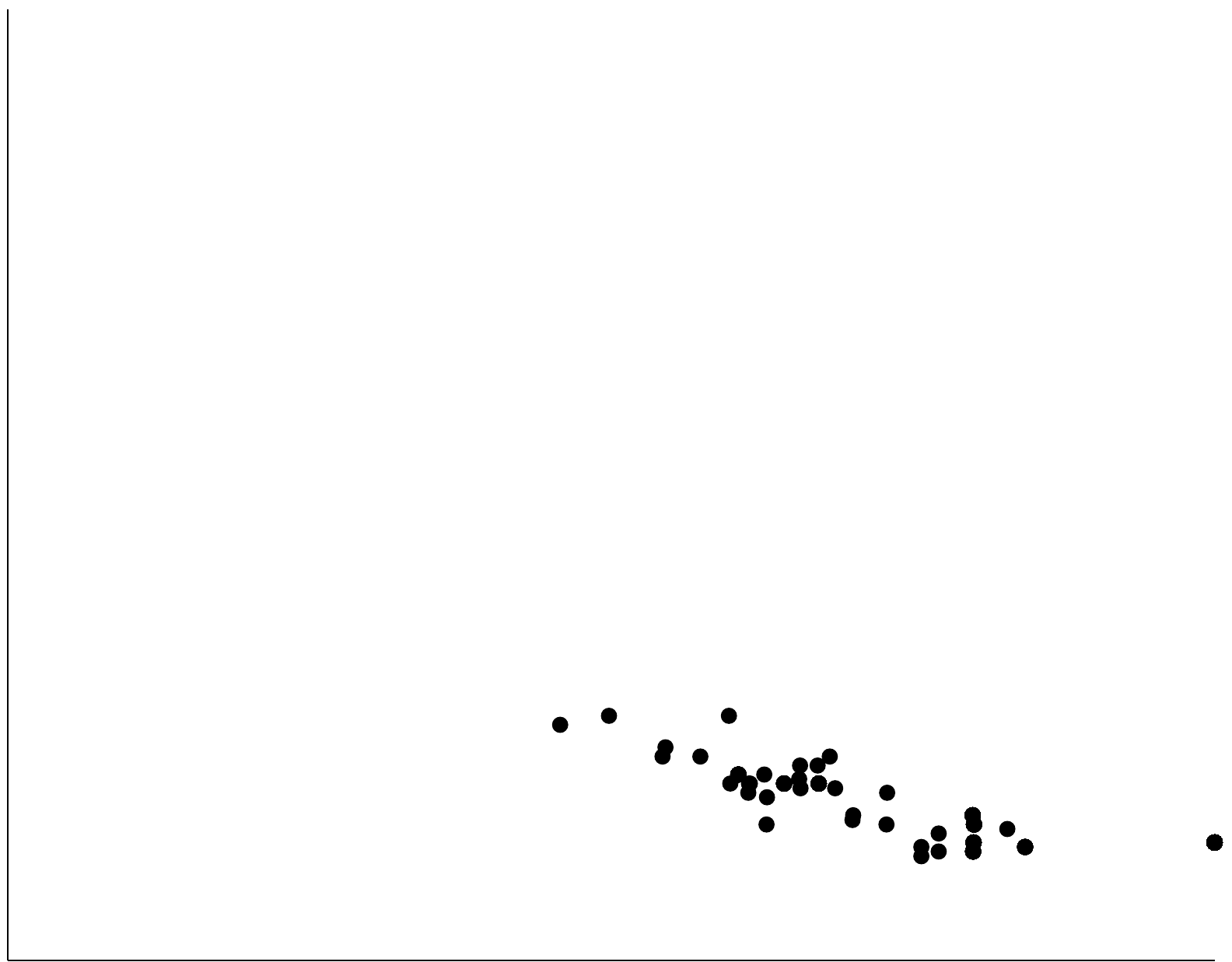}
\end{subfigure}\\ \\
\end{tabular}
}
\label{kappaerror30}
\end{sidewaystable}

\subsection{Kappa-error Plots}
Kappa-error plots \cite{Kappa} is a method to understand the diversity-error behaviour of an ensemble. 
These plots represent a point for each pair of classifiers in the ensemble. For each classifier pair ($D_i$ and $D_j$), the {\it x} coordinate is a measure of diversity of the two classifiers $D_i$ and $D_j$ known as the {\it kappa} ($\kappa $) measure, where low values suggest high diversity. The {\it y} coordinate is the average error $E_{i,j}$ of the two classifiers $D_i$ and $D_j$. When the agreement of the two classifiers equals than expected by chance, $\kappa = 0$; when they agree on every instance, $\kappa = 1$ \cite{Dietterich2000}. Negative values of $\kappa$ mean a systematic disagreement between the two classifiers. The   most   desirable  pairs  of classifiers (high diversity and low average error) will lie at the bottom left corner.

An ensemble of size $L$ create $L$($L$-1)/2 pairs of  classifiers. In our case, an ensemble has 25 classifiers, therefore there are 400 dots in each plot.
We draw kappa-error plots for four datasets, i.e., Breast Tissue, New-Thyroid, Column, and Seeds, for different ensemble methods.
The scales of $\kappa$ and $E_{i,j}$ are same for each given dataset, so we can easily compare different ensemble methods. Tables \ref{kappaerror10} and \ref{kappaerror30} show the kappa-error plots of the testing phase of first run of the first cross-validation fold for each of the data at missing ratio of $10\%$ and $30\%$. The rows show each of the datasets and the columns show the kappa-error plots for the different ensemble imputation methods. 

Some of the plots show only few points, which means that only few of the classifiers have distinct results. Generally, these kinds of graphs are for MIEM method. This suggests that this method is not creating diverse classifiers. 
These plots suggest that most of the ensemble methods have similar diversity pattern. However, BagEM classifiers have better accuracy as groups of points are lower as compared to other ensemble methods. Accurate classifiers with reasonable  diversity is  the reason for the robust performance of BagEM at high missingness ratio.
\section{Conclusion and Future Work}
\label{sec:conclusion}
Handling missing data is a challenging task in data mining application. MI methods are commonly employed because they can model the uncertainty due to missingness. Bootstrapping is another method through which diversity may be incorporated in the incomplete data. Combining both the ideas together can lead to more accurate and diverse classifier that can lead to robust ensemble with respect to high missingness ratio. In this paper, we presented different variations of combining ideas from MI and bootstrapping for data imputation and compare their performances. Our results show that ensemble based imputations perform better than their single imputation counterparts for smaller missingness ratio of $10\%$ or more. The performance of MI over bootstrap samples with EM as the base imputation method does not degrade much for up to $30\%$ missingness ratio. It is consistently observed that no imputation on incomplete data with bootstrapping performs better than single imputation and is equivalent to other ensemble imputation methods for missingness ratio of up to $10\%$. The kappa-error plots further verify that bagging and MI lead to diverse and accurate classifiers. Thus, their ensemble are more robust to missingness, in comparison to MI ensemble or single imputation methods. These findings in this paper are important from data scientists' perspective because based on the missing ratio in the data, they can choose the right type of classification strategies without performing hit and trial methods. As we carried out the experiments  by using MCAR missingness, our findings are valid only for this type of  missingness.  In future, we will extend this study by including other imputation methods, such as KNN and Bayesian methods over multiple datasets. We will also use non-decision tree based supervised classifiers with these ensemble methods. We will also study the performance of various ensemble methods for cases when the decision boundary is created only by missing values. 
\section{Acknowledgement} This work was supported by a UAE university Start-up grant (grant number G00002668; fund number 31T101).

\bibliography{references}
\bibliographystyle{plain}

\end{document}